\documentclass[twocolumn]{svjour3}          
\smartqed  
\usepackage{graphicx}
\usepackage{amsmath,amssymb} 
\usepackage{color}
\usepackage{subfigure}
\usepackage{romannum}
\usepackage{booktabs}
\usepackage{multirow}
\usepackage{caption}
\usepackage{bbding}
\usepackage{float}
\restylefloat{table}
\usepackage{appendix} 
\usepackage{flushend,cuted}
\begin{document} 	

	\title{CNN-Based Automatic Urinary Particles Recognition
	}
	
	
	\author{Rui Kang \and Yixiong Liang \Envelope \and Chunyan Lian \and Yuan Mao
	}
	
	
	\institute{R. Kang \and Y. Liang \Envelope \and C. Lian \and Y. Mao \at
		School of Information Science and Engineering, Central South University, Changsha 410083, China \\
		\email{yxliang@csu.edu.cn}           
	}
	
	\date{Received: date / Accepted: date}

	\maketitle

\begin{abstract}

The urine sediment analysis of particles in microscopic images can assist physicians in evaluating patients with renal and urinary tract diseases. Manual urine sediment examination is labor-intensive, subjective and time-consuming, and the traditional automatic algorithms often extract the hand-crafted features for recognition. Instead of using the hand-crafted features, in this paper, we exploit CNN to learn features in an end-to-end manner to recognize the urine particles. We treat the urine particles recognition as object detection and exploit two state-of-the-art CNN-based object detection methods, Faster R-CNN and SSD, as well as their variants for urine particles recognition. We further investigate different factors involving these CNN-based object detection methods for urine particles recognition. We comprehensively evaluate these methods on a dataset consisting of 5,376 annotated images corresponding to 7 categories of urine particles, i.e., erythrocyte, leukocyte, epithelial cell, crystal, cast, mycete, epithelial nuclei, and obtain a best mAP (mean average precision) of 84.1\% while taking only 72 ms per image on a NVIDIA Titan X GPU. 
\end{abstract}

\section{Introduction}
The urine sediment examination of biological particles in microscopic images is one of the most commonly performed vitro diagnostic screening tests in clinical laboratories and it plays an important role in evaluating the kidney and genitourinary system and monitoring body state. General indications for urinalysis include: the possibility of urinary tract infection or urinary stone formation; non-infectious renal or post-renal diseases; in pregnant women and patients with diabetes mellitus or metabolic states who may have proteinuria, glycosuria, ketosis or acidosis/alkalosis \cite{kouri2000eclm,ince2016comparison}. 

Traditionally, the trained technicians count the number of each kind of particles of urinary sediment by visual inspection. The manual urine sediment examination works but is labor-intensive, time-consuming, subjective, and operator-dependent in high-volume laboratories.

\begin{figure*}[t]
      	\centering
         	\subfigure[The traditional multi-stage pipeline]{\label{fig:multi-stage}
      		\includegraphics[width=0.6\linewidth]{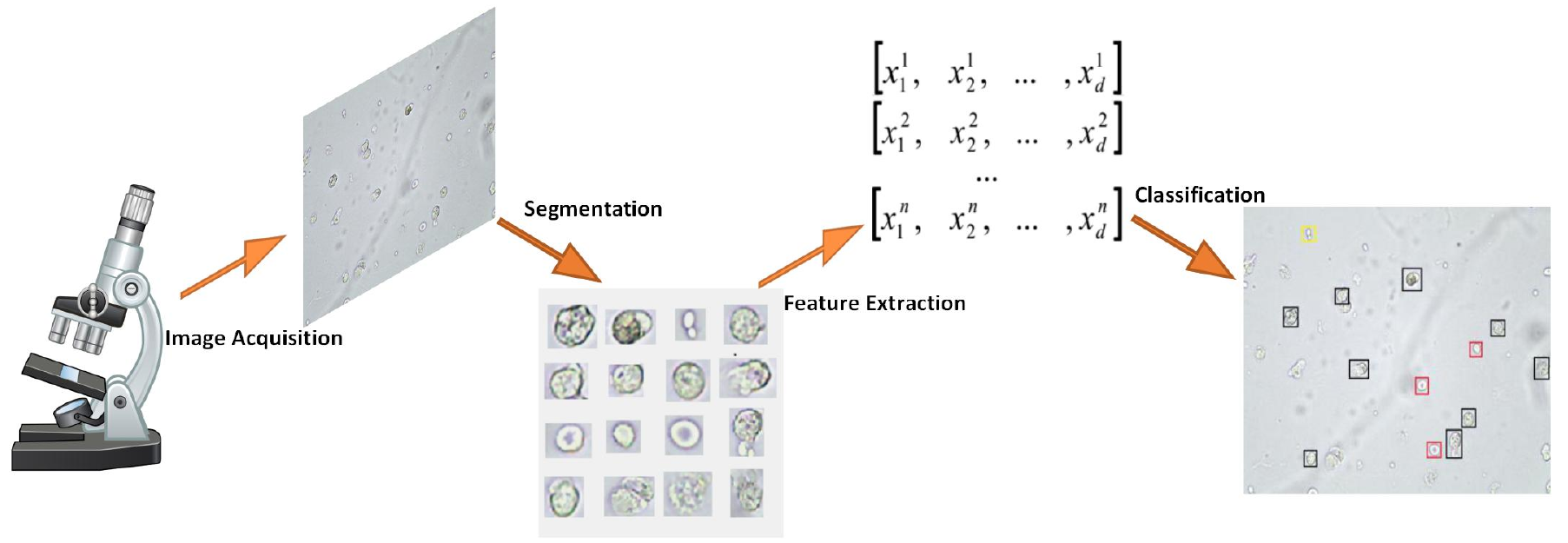}}   	
       \centering
            \subfigure[The CNN-based end-to-end pipeline]{\label{fig:end2end}
      		\includegraphics[width=0.6\linewidth]{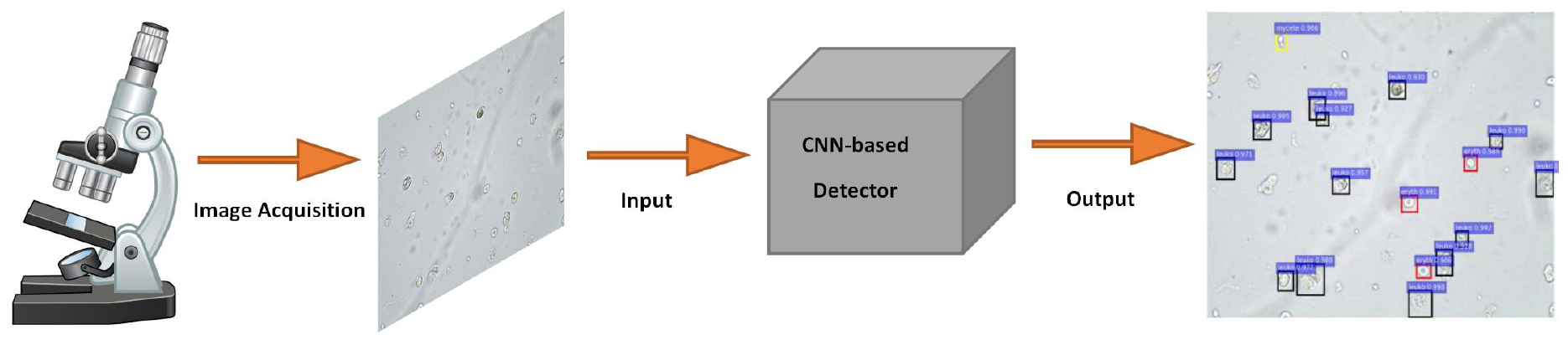}}   	
          \caption{The pipelines for urinary particles recognition.}
      \end{figure*}

The issues involved in the manual analysis have motivated lots of automated methods for the analysis of urine microscope images (e.g. \cite{ranzato2007automatic,liang2009false,shen2009urine,almadhoun2014automated,avci2014new,li2015join}). As shown in figure \ref{fig:multi-stage}, almost all of them follow the multi-stage pipeline, i.e., first generating candidate regions based on segmentation and then extracting hand-crafted features over regions for classification. Therefore, the performance of these methods heavily depends on the accuracy of the segmentation and the effectiveness of the hand-crafted features. However, due to the complicated characteristics of urinary images, the precise segmentation of the interested particles is quite difficult, or even impossible, and the resulting hand-crafted region features are often less discriminatory.

To avoid the precise segmentation stage and improve the discriminability of features, as shown in figure \ref{fig:end2end}, in this paper we exploit CNN to automatically learn task-specific features and perform the urine particles recognition in an end-to-end manner. We treat the urine particles recognition as object detection and exploit two well-known CNN-based object detection methods, Faster R-CNN \cite{ren2015faster} and SSD \cite{liu2016ssd}, along with their variants including Multiple Scale Faster R-CNN (MS-FRCNN) \cite{hoang2016multiple}, Faster R-CNN with online hard example mining \cite{shrivastava2016training} (OHEM-FRCNN) and Trimmed SSD to accomplish it. We investigate different factors such as training strategies, network structures, fine-tuning tricks, data augmentation etc, to make these methods more appropriate for urine particles recognition .

In this study, we exploringly apply both Faster R-CNN and SSD approaches to the recognition of urinary sediment particles. The end-to-end methods integrate feature extraction, location and classification to an unified convolutional network. Avoiding segmentation and hand-crafted features extraction, they can automatically learn urine-specific recognition task from annotated micro-images.

We summarize our contributions as follows:
\begin{itemize}
  \item We exploit Faster R-CNN \cite{ren2015faster} and SSD \cite{liu2016ssd} for urine sediment recognition. It is segmentation free and can learn task-specific features in an end-to-end manner.
  \item We investigate various factors to improve the performance of Faster R-CNN \cite{ren2015faster} and its variants \cite{hoang2016multiple,shrivastava2016training} for urine particles recognition. 
  \item We propose a scheme, Trimmed SSD, to prune the network structure adopted in SSD \cite{liu2016ssd} to achieve better performance for urine sediment recognition.
  \item We obtain a best mAP of 84.1\% while taking only 72 ms per image for 7 categories recognition of urine sediment particles. Importantly, we also get a best AP of 77.2\% for cast particles, the most valuable but most difficult to detect ingredients \cite{chien2007urine,budak2011comparison,zhou2012automatic}.
\end{itemize}

The remainder of this paper is organized as follows: Section 2 reviews the research status of urine particles recognition and the development of CNN-based methods for generic object detection. Section 3 describes the detection architectures applied to urine particles recognition, i.e., Faster R-CNN, SSD and their variants. Section 4 details the urinalysis database organization and provides deep analysis of extensive experiments for urine particle recognition. Section 5 shows more experimental comparisons intuitively. Section 6 presents our final conclusions.

\section{Related work}
\subsection{Urine particles recognition}
The recognition of urinary sediment particles has been extensively studied following the traditional multi-stage pipeline (Fig \ref{fig:multi-stage}) and a variety of approaches can be adopted in each stage. 

In \cite{ranzato2007automatic}, Rabznto et al. first obtained patches of interest by a detection algorithm, and then extracted invariant features based on ``local jets'' \cite{schmid1997local}. Classification performed with a mixture of Gaussians classifier \cite{bishop1995neural}. Although the system presented reliable recognition results on a pollen dataset, more accurate location for interest patches needed to be improved. Liang et al. \cite{liang2009false} adopted a two-step process (the first location step and the second tuning step) to segment particles' contour. After features extraction by a novel local jet context feature scheme, they proposed a two-tier classification strategy to better reduce the false positive rate caused by impurity and poor focused regions. Shen et al. \cite{shen2009urine} used AdaBoost to select a little part typical Harr features for SVM classification, and improved system speed via cascade accelerating algorithm. Zhou et al. \cite{zhou2012automatic} demonstrated an easy-implemented automatic urinalysis system employing a SVM classifier to distinguish casts from other particles. In paper \cite{avci2014new}, a new technique based on the Adaptive Discrete Wavelet Entropy Energy for adaptive feature extraction was proposed, which follows the image preprocessing stage including noise reduction, contrast enhancement and segmentation. In classification, the Artificial Neural Network (ANN) classifier was selected for the best performance. Li et al. \cite{li2015join} mainly focused on the texture feature extraction of the segmented urinary particles. After the adhesive particles separation by watershed algorithm, the authors combined the Gabor filter with the scattering transform for robust feature description, which not only keeps invariant of scaling, rotation and translation but also shows good performance in the course of SVM classification.

The conventional recognition model works for automated urinalysis, but importantly, segmentation, feature extraction and classification all need to be carefully designed. In addition, the complicated characteristics of urine micro-images also bring more challenges to this task. Therefore, there is an increasing demand for better solutions relying more on automatic learning and less on hand-designed heuristics.

      \begin{figure*}[t]
      	\centering
      	\subfigure[Faster R-CNN]{\label{fig:FRCNN}
      		\includegraphics[height=4.5cm,width=4.8cm]{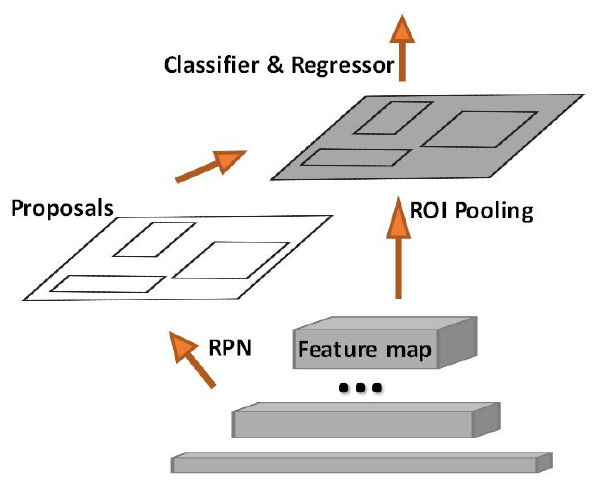}}   	
      	\hspace{-2ex}
      	\subfigure[MS-FRCNN]{\label{fig:MS-FRCNN}
      		\includegraphics[height=4.5cm,width=5.6cm]{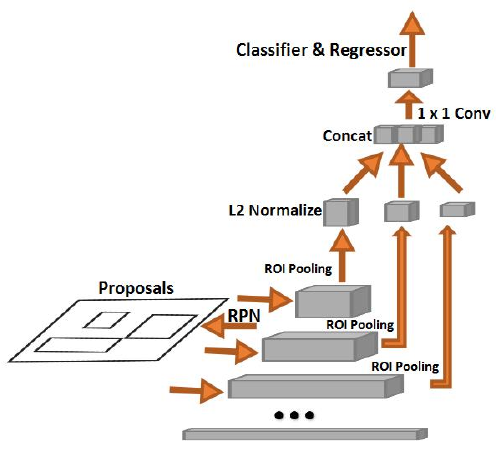}}
      	\subfigure[OHEM-FRCNN]{\label{fig:OHEM-FRCNN}
      		\includegraphics[height=5cm,width=6cm]{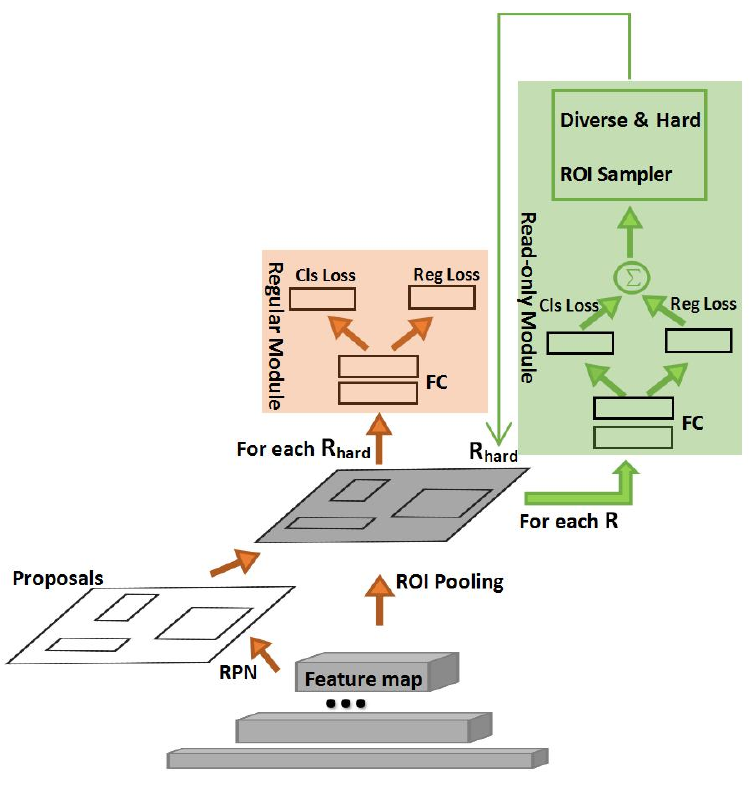}}   	
      	\caption{The architectures of Faster R-CNN, MS-FRCNN and OHEM-FRCNN.}
      	\label{fig:methods}
      \end{figure*}
           
\subsection{CNN-based object detection}
Since the revival of deep Convolutional Neural Networks (CNNs) \cite{lecun1998gradient,krizhevsky2012imagenet} with prominent performance, the state-of-the-art methods in image classification and object detection have all used deep learning techniques. Specially in generic object detection, there are two established series as representatives of deep learning methods: the Overfeat \cite{sermanet2013overfeat} series based on sliding windows and the R-CNN \cite{girshick2014rich} series based on region proposals classification.

On the one hand, since Girshick et al. proposed R-CNN \cite{girshick2014rich} combing region proposals with CNNs, the method has attracted wide attentions and has been improved in a variety of ways. First, in order to mitigate the time-consuming process of features computation, SPP-net \cite{he2014spatial} introduces a spatial pyramid pooling layer that can flexibly handle variable-size inputs. Avoiding repeatedly computing the convolutional features (compute only once per image), it accelerates R-CNN significantly. Instead of a spatial pyramid pooling layer, then Girshick extended SPP-net by a ROI pooling layer, and introduced a multi-task loss, namely, the joint classification loss and bounding box regression loss. With the two improvements, the framework can fine-tune all layers in an end-to-end manner, which comparatively speeds up the stages of training and testing, so called as Fast R-CNN \cite{girshick2015fast}. 

Until now, compelling speed and accuracy have been achieved. However, when considering the computation time spent on region proposals (e.g., Selective Search \cite{uijlings2013selective}), this process immediately becomes the bottleneck of object detection systems. Discarding the engineered low-level features used in most popular methods, several papers (like MultiBox \cite{erhan2014scalable,szegedy2014scalable}) generate region proposals directly from a auxiliary deep neural network. Further, in Faster R-CNN \cite{ren2015faster}, Ren et al. merged a region proposal network (RPN) and Fast R-CNN into a single network by sharing their full-image convolutional features, thus producing small marginal cost for region proposals generation. Combining different networks, Faster R-CNN can detect general objects very accurately at near real-time rate.

Faster R-CNN is a powerful baseline system and is flexible to many applications. Although Region-based Fully Convolutional Network (R-FCN) \cite{li2016r} has been proposed recently, Faster R-CNN is still very prevalent in the region-based family. In addition, there are several variants of Faster R-CNN for domain-specific detection. For example, aiming at the detections of driver's cell-phone usage and hands on steering wheel, Le et al. presented a multiple scale Faster R-CNN (MS-FRCNN) \cite{hoang2016multiple}, which mainly addresses low precision problem to small object detection. Also Zhang et al. in \cite{zhang2016faster} improved unsatisfactory accuracy to small instances by pooling finer features from shallower layers and increasing feature maps size via the ``\`{a} trous'' trick.

On the other hand, skipping the proposal step, the OverFeat \cite{sermanet2013overfeat} series  directly predicts confidences and bounding boxes accross multiple categories through a sliding window mechanism. Initially, Overfeat \cite{sermanet2013overfeat} simultaneously run classifier and regressor networks at each spatial location and scale for confidences and bounding boxes prediction, and adopts a greedy merge strategy to complete final detections. But Overfeat is a disjoint system. So, YOLO \cite{redmon2016you} frames object detection as a single regression problem and predicts multiple bounding boxes and class probabilities directly from full image in one evaluation. Both OverFeat and YOLO approaches only use the topmost feature map to detect all categories, which leads to unsatisfactory detection results. In order to flexibly handle various-size objects, SSD \cite{liu2016ssd} sets a set of default boxes over different aspect ratios and scales at each feature map location, and combines all predictions from multiple feature maps with different resolutions. Therefore, SSD has achieved state-of-the-art performance in the OverFeat series. But SSD has a notable drawback. It foregoes reusing the higher-resolution maps of the feature hierarchy, just as pointed out in Feature Pyramid Network(FPN) \cite{lin2016feature}. In the following experiments section, we also demonstrate the importance of these finer features for small objects detection, such as erythrocyte, leukocyte and epithelial nuclei.

\section{Methods}
In this paper, we employ two well-known CNN-based object detection methods,  Faster RCNN \cite{ren2015faster} and SSD \cite{liu2016ssd}, to urine particles recognition, and further exploit several structural variants, namely, Multiple Scale Faster R-CNN (MS-FRCNN) \cite{hoang2016multiple}, Faster R-CNN with online hard example mining (OHEM-FRCNN) and Trimmed SSD.
  
\subsection{Faster R-CNN and its variants}

\textbf{Faster R-CNN} \cite{ren2015faster} is a single unified network which integrates a fully convolutional region proposal generator (RPN) with a fast region-based object detector (Fast R-CNN) \cite{girshick2015fast}. As shown in figure \ref{fig:FRCNN}, the deep detection framework also can be described as the pipeline of ``shareable CNN feature extraction + region proposal generation + region classification and regression''.  Moreover, to predict objects across multiple scales and aspect ratios, the authors in paper \cite{ren2015faster} designed a pyramid of anchors creatively, which is a key component for sharing features without extra cost. Therefore, Faster R-CNN is a segmentation free method without using hand-crafted features and successful for general object detection.

\textbf{MS-FRCNN} \cite{hoang2016multiple} is a follow-up improvement based on Faster R-CNN to detect whether driver's hands are on a steering wheel or not and if a cell-phone is being used. It keeps Region Proposal Network (RPN) unchanged and builds a more sophisticated network for Fast R-CNN detector by a combination of both global context and local appearance features. As figure \ref{fig:MS-FRCNN} shows, each object proposal receives three feature tensors through ROI pooling from the last three convolutional layers. After L2 normalization to each tensor, outputs are concatenated and compressed to maintain the same size as the original architecture.

\textbf{OHEM-FRCNN} is a combination of online hard example mining (OHEM) \cite{shrivastava2016training} and Faster R-CNN \cite{ren2015faster}. OHEM \cite{shrivastava2016training} is a novel bootstrapping for modern CNN-based object detectors trained purely online with SGD, like Fast R-CNN \cite{girshick2015fast}. Instead of a sampled mini-batch \cite{ren2015faster}, it eliminates several heuristics and hyperparameters in common use and selects automatically hard examples by loss. Although OHEM can be applied to any region-based ConvNet detector, no combination of OHEM and Faster R-CNN has been published yet. As figure \ref{fig:OHEM-FRCNN} shows, in this paper we apply OHEM to Faster R-CNN for urine particles recognition. For each iteration, given the feature map from shareable convolutional network and ROIs from RPN, the read-only ROI network performs a forward pass and computes loss for all input ROIs. Then the regular ROI network computes forward and backward passes only for hard examples selected by hard ROI sampling module according to a distribution that favors diverse, high loss candidates.

      \begin{figure}[h!]
      	\centering
      	\includegraphics[height=3cm,width=6.5cm]{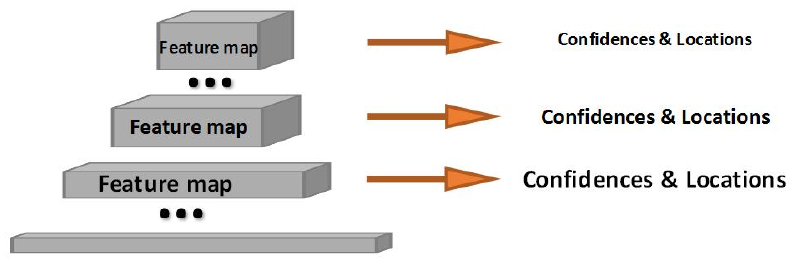}   		
      	\caption{The architecture of SSD.} 
      	\label{fig:SSD}
      \end{figure} 
\subsection{SSD and its variants}
\textbf{SSD} \cite{liu2016ssd}, a single-shot multibox detector for multiple categories, can be decomposed into a truncated base network (usually a VGG-16 net) and several auxiliary convolutional layers used as feature maps and predictors. Unlike Faster R-CNN \cite{ren2015faster}, SSD increases detection speed by removing the region proposal generation and the subsequent pixel or feature resampling stages. Unlike YOLO \cite{redmon2016you}, it improves detection quality by applying a set of small convolutional filters to multiple feature maps to predict confidences and boxes offsets for various-size categories (like figure \ref{fig:SSD} shows).

\textbf{Trimmed SSD} is a simplified version of the original SSD model \cite{liu2016ssd}. As figure \ref{fig:SSD} shows, from bottom to top, original SSD selects conv4\_3, fc7 (convolutional layer), conv6\_2, conv7\_2, conv8\_2, conv9\_2 and pool6 as feature maps to produce confidences and locations. If we directly transfer it to urine particles recognition with only 7 categories, it may produce a large number of redundant prediction results interfering with the final detection performance. And the framework is too complicated to perfectly fit our dataset. For simplification, we attempt to remove several top convolutional layers from the auxiliary network of SSD, which leads to the trimmed SSD.

~\\
When applying above methods to urine particles recognition, we effectively adopt the mechanism of deep transfer learning, and conduct extensive experimental analysis to demonstrate the impact of various factors. Specifically, in the course of Faster R-CNN being used, we explore different training strategies, network structures and anchor scales, and carry out data augmentation to further increase training samples. Also, when using SSD, we adjust several parameters, including the scales of default boxes (similar to the Faster R-CNN anchors) and the size of input images to boost small objects detection.

\section{Experiments}
\subsection{Dataset organization}
\begin{figure*}[t]
          	\begin{minipage}[t]{0.5\linewidth}
          	\centering
          	\includegraphics[height=5cm,width=8cm]{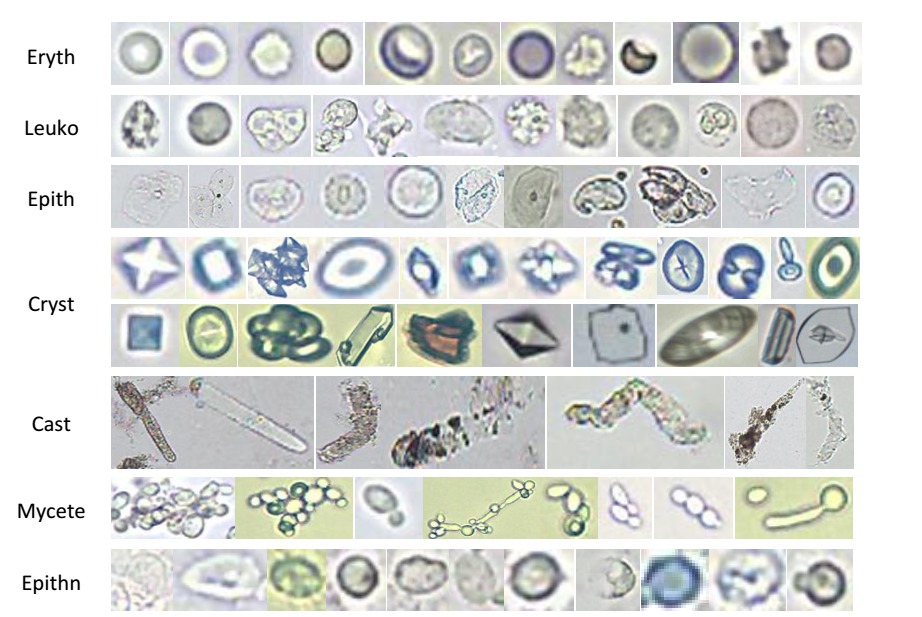}
          	\caption{Selected samples of urinary sediment particle.}
          	\label{fig:urine}
          	\end{minipage}
          	\hspace{0.2cm}   	
          	\begin{minipage}[t]{0.5\linewidth}
          	\centering
          	\includegraphics[height=5cm,width=8cm]{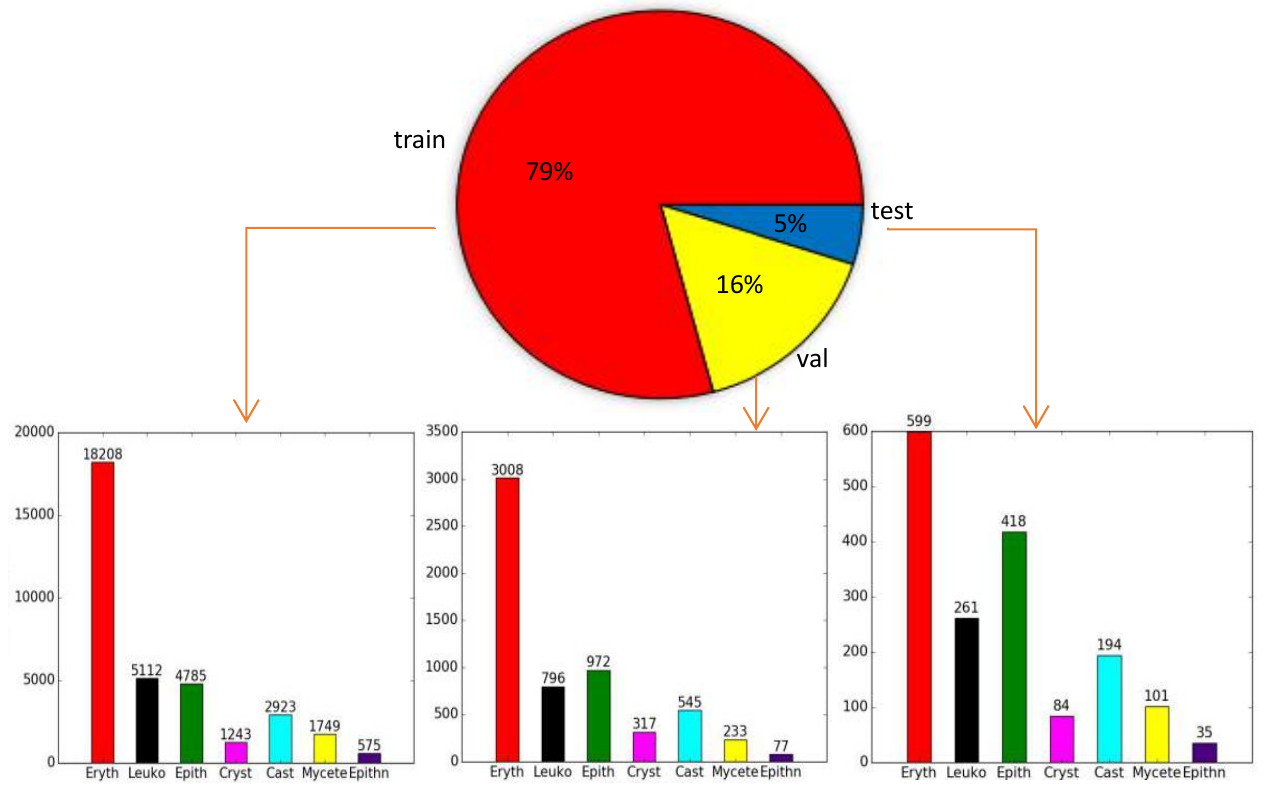}
          	\caption{Dataset organization and categories distribution.}          	
          	\label{fig:distribution}
            \end{minipage}                	
          \end{figure*}

In order to perform our study, we first establish the urinalysis micro-images database that is marked with ground truth boxes by clinical experts. All 6,804 annotated color images have a size of 800 x 600, which include 8 categories of urinary sediment particles, i.e., erythrocyte (eryth), leukocyte (leuko), epithelial cell (epith), crystal (cryst), cast, mycete, epithelial nuclei (epithn) and noise. Specifically, eryth, leuko, crystal, mycete and epithn are only annotated at high-power field, epith and cast only at low-power field. Figure \ref{fig:urine} shows 7 categories of urinary sediment particles from our database, each of which includes many subcategories with various shapes.

In fact, our 6,804 annotated images have a total of 273,718 ground truths, where meaningless noise occupies 230,919 annotations, up to eight-four percent. We remove images only including noise and  finally get 5,376 useful images, more concretely, which contain (ground truth boxes) 21,815 for eryth, 6,169 for leuko, 6,175 for epith, 1,644 for cryst, 3,663 for cast, 2,083 for mycete and 687 for epithn. From the final 5,376 images, we randomly select 268 images making up 1/20 as test set, and the others as trainval set, where train set makes up 5/6. Figure \ref{fig:distribution} demonstrates the details of dataset organization and categories distribution. The top pie chart shows how 5,376 images are organized into train/val/test sets. The bottom bar graphs display detailed objects distribution for the imbalanced database.

\subsection{Experimental analysis}

Our experiments perform on a 64 bits Ubuntu 14.04.5 computer with CPU Intel(R) Core(TM) i7-5960X CPU@ 3.00GHz, NVIDIA Corporation GM200 [GeForce GTX TITAN X] and Python 2.7.6. In training stage, we adopt the transfer learning mechanism: first initialize CNN frameworks with models pre-trained on ImageNet dataset, then fine tune them using different strategies. By default, we still use PASCAL-style Average Precision (AP) at a single IoU threshold of 0.5 and mean Average Precision (mAP) to evaluate our detection results. Lots of significant test results have been obtained via various kinds of training trials. Details are listed below.

\subsubsection{Urine particles recognition based on Faster R-CNN}

When training Faster R-CNN, we fine-tune pre-trained models with SGD for 70k mini-batch iterations (unless specified otherwise), with a mini-batch size of 128 on 1 GPU, a momentum of 0.9 and a weight decay of 0.0005. We start from a learning rate 0.001, and decrease it by 1/10 after 50k iterations. But fine-tuning PVANet \cite{kim2016pvanet} adopts a learning rate policy of plateau: 0.003 base learning rate, 0.3165 gamma and a different weight decay of 0.0002. 

~\\
\textbf{Different training strategies.} As all know, there are two training solutions in the released Faster R-CNN python code, 4-step alternating training and approximate joint training (also called as end2end training). In order to select one more effective and efficient solution for the following networks training, we design this experiment based on small ZF \cite{zeiler2014visualizing} net and medium VGG-16 \cite{simonyan2014very} net. By default, the training parameters of two networks remain the same with the released code. Table \ref{tab:table1} shows that adopting the strategy of approximate joint training takes less time, but yields higher mAP (nearly the same accuracy on VGG-16 net), so the next series of experiments all adopt the end2end training solution.
\begin{table*}[t]
	\centering
	\tiny	
	\setlength{\abovecaptionskip}{0cm}
	\setlength{\belowcaptionskip}{-2cm}
	\caption{Training time and mAP by different training solutions. Experiments perform on the ZF and VGG-16 networks, and keep the default settings from the released python code, in which the iteration parameters of 4-step alternating training is [80,000 40,000 80,000 40,000].}
	\begin{tabular}{c|c|c|c}
		net   & training strategy & training time(h) & mAP \\
		\midrule
		\multirow{2}[4]{*}{ZF} & 4-step alternating training & 5.33  & 0.694 \\
		\cmidrule{2-4}          & end2end training & \textbf{4.6} & \textbf{0.723} \\
		\midrule
		\multirow{2}[3]{*}{VGG-16} & 4-step alternating training & 12.23 & 0.756 \\
		\cmidrule{2-4}          & end2end training & \textbf{11.68} & \textbf{0.757} \\
		
	\end{tabular}%
	\label{tab:table1}%
\end{table*}%
\begin{table*}[t]
	\centering
	\tiny
	\setlength{\abovecaptionskip}{0cm}
	\setlength{\belowcaptionskip}{-2cm}	
	\caption{Comparisons of detection results using different networks and different anchor scales. By default, we keep anchor ratios unchanged: only PVANet sets it to \{0.5, 0.667, 1.0, 1.5, 2.0\}, all the others set \{2:1, 1:1, 1:2\}. The last column is an approximate time of net forward-propagating when test an image. And the last two rows are test results after 60k mini-batch iterations.}
	\begin{tabular}{c|c|c|ccccccc|c}
		\multirow{2}[1]{*}{net} & \multirow{2}[1]{*}{anchor scales} & \multirow{2}[1]{*}{mAP} &  \multirow{2}[1]{*}{eryth} & \multirow{2}[1]{*}{leuko} & \multirow{2}[1]{*}{epith} & \multirow{2}[1]{*}{cryst} & \multirow{2}[1]{*}{cast} & \multirow{2}[1]{*}{mycete} & \multirow{2}[1]{*}{epithn} & test time  \\
		&       &       &       &       &       &       &       &        &       & (sec/img) \\
		\midrule
		\multirow{4}[2]{*}{ZF} & $\{ 128^{2}, ~256^{2}, ~512^{2} \}$ & 0.723 & 0.607 & 0.749 & 0.845 & 0.856 & 0.658 & 0.781 & 0.566 & \textbf{0.044} \\
		& $\{ 64^{2}, ~128^{2}, ~256^{2}, ~512^{2} \}$ & 0.796 & 0.853 & 0.809 & 0.855 & 0.858 & 0.671 & 0.861 & 0.665 & 0.045 \\
		& $\{ 32^{2}, ~64^{2}, ~128^{2}, ~256^{2}, ~512^{2} \}$ & 0.779 & 0.859 & 0.805 & 0.854 & 0.847 & 0.657 & 0.863 & 0.57  & 0.046 \\
		& $\{ 64^{2}, ~128^{2}, ~256^{2} \}$ & 0.757 & 0.748 & 0.823 & 0.846 & 0.85  & 0.642 & 0.82  & 0.568 & 0.044 \\
		\midrule
		\multirow{4}[2]{*}{VGG-16} & $\{ 128^{2}, ~256^{2}, ~512^{2} \}$ & 0.757 & 0.599 & 0.772 & \textbf{0.874} & 0.794 & 0.708 & 0.874 & 0.679 & 0.102 \\
		& $\{ 64^{2}, ~128^{2}, ~256^{2}, ~512^{2} \}$ & 0.802 & 0.842 & 0.818 & 0.868 & 0.873 & 0.716 & 0.877 & 0.621 & 0.104 \\
		& $\{ 32^{2}, ~64^{2}, ~128^{2}, ~256^{2}, ~512^{2} \}$ & 0.795 & 0.854 & 0.825 & 0.857 & 0.851 & 0.724 & 0.876 & 0.576 & 0.104 \\
		& $\{ 64^{2}, ~128^{2}, ~256^{2} \}$ & 0.762 & 0.743 & 0.822 & 0.863 & 0.759 & 0.712 & 0.88  & 0.558 & 0.104 \\
		\midrule
		\multirow{3}[2]{*}{ResNet-50} & $\{ 128^{2}, ~256^{2}, ~512^{2} \}$ & 0.77  &  0.613 & 0.831 & 0.853 & 0.852 & 0.757 & 0.873 & 0.615 & 0.219 \\
		& $\{ 64^{2}, ~128^{2}, ~256^{2}, ~512^{2} \}$ & 0.784 & 0.761 & 0.824 & 0.86  & 0.822 & 0.768 & 0.859 & 0.595 & 0.219 \\
		& $\{ 32^{2}, ~64^{2}, ~128^{2}, ~256^{2}, ~512^{2} \}$ & 0.804 & 0.876 & 0.812 & 0.86  & 0.854 & 0.747 & 0.874 & 0.605 & 0.22 \\
		\midrule
		\multirow{3}[2]{*}{ResNet-101} & $\{ 128^{2}, ~256^{2}, ~512^{2} \}$ & 0.761 & 0.606 & 0.83  & 0.864 & 0.802 & 0.769 & 0.875 & 0.578 & 0.268 \\
		& $\{ 64^{2}, ~128^{2}, ~256^{2}, ~512^{2} \}$ & 0.773 & 0.841 & 0.814 & 0.848 & 0.852 & 0.749 & 0.863 & 0.446 & 0.267 \\
		& $\{ 32^{2}, ~64^{2}, ~128^{2}, ~256^{2}, ~512^{2} \}$ & 0.801 & 0.872 & 0.809 & 0.839 & 0.852 & \textbf{0.772} & 0.883 & 0.581 & 0.268 \\
		\midrule
		PVANet & $\{ 48^{2}, ~96^{2}, ~144^{2}, ~256^{2}, ~512^{2} \}$ & \textbf{0.841} & \textbf{0.884} & \textbf{0.843} & 0.871 & \textbf{0.877} & 0.765 & \textbf{0.890} & \textbf{0.760} & 0.072 \\
	\end{tabular}%
	\label{tab:table2}%
\end{table*}%
~\\
\textbf{Different networks.} In this part, we mainly use 4 networks pre-trained on ImageNet for classification to initialize our detection model: ZF net \cite{zeiler2014visualizing} proposed by Zeiler and Fergus is a small and fast convolutional version; VGG-16 net \cite{simonyan2014very}, a medium version, proposed by Simonyan and Zisserman has 16 shareable convolutional layers; the deeper ResNet \cite{he2016deep}, including ResNet-50 and ResNet-101, introduces a residual learning framework to ease the optimization of training stage; and the latest PVANet \cite{kim2016pvanet} has less channels and more layers. We fine tune all convolutional layers of ZF net and PVANet, the conv3\_1 and up of VGG-16 net and ResNet.

 From ZF, VGG-16 to ResNet-50, table \ref{tab:table2} roughly demonstrates that as nets go deeper, we get higher detection accuracy and more testing time per image. In anchor scales of \{ $64^{2}$, $128^{2}$, $256^{2}$, $512^{2}$ \}, the mAPs fluctuate, which mostly is a trade-off between stronger semantics and coarser resolutions. Moreover, from ResNet-50 to ResNet-101, mAPs are nearly unchanged. we due it to the complexity of our dataset. Just as paper \cite{kim2016pvanet} claims the deep but lightweight PVANet can achieve solid detection results while minimizing computation cost, we indeed obtain the best performance on our urine sediment recognition task.

~\\
\textbf{Different anchor scales.} Unlike generic objects in camera images, the particles of urinary sediment vary widely in their shapes, sizes and numbers. Moreover, some urinary micro-images include a lot of small objects (like erythrocyte and leukocyte), so as many anchors as possible should be covered in our experiment, especially small scales.

 In this part, we compare the detection results using different anchor scales. First, for networks of ZF, VGG-16 and ResNet we all choose the default settings (the anchor scales of \{ $128^{2}$, $256^{2}$, $512^{2}$ \} and the aspect ratios of \{1:1, 1:2, 2:1\}) as benchmarks. Then, keep aspect ratios unchanged and gradually increase anchors with smaller scales (i.e., $64^{2}$ and $32^{2}$). Overall, table \ref{tab:table2} shows us that more anchors yield higher mAP. In detail, increasing anchor scales \{ $64^{2}$, $128^{2}$, $256^{2}$, $512^{2}$ \} to \{ $32^{2}$, $64^{2}$, $128^{2}$, $256^{2}$, $512^{2}$ \} can not achieve better performance on both ZF net and VGG-16 net. It mainly due to the capacity of networks becoming saturated because we do get an accuracy boost when using ResNet-50 and ResNet-101. Further, we delete the scale of $512^{2}$ as comparison only using ZF and VGG-16 nets. On ZF net, the scales of $\{ 64^{2}, ~128^{2}, ~256^{2} \}$ has the same 9 anchors with \{ $128^{2}$, $256^{2}$, $512^{2}$ \}, but outperforms by 3.4\% mAP. Similarly, on VGG-16 net, increases by 0.5\% mAP. It indicates that most particles in our dataset are small objects and the small anchor scales are indispensable. In addition, we note that deeper networks take more test time, but anchor scales have little impact on detection cost. Finally, it's worth mentioning that the PVANet with best performance takes less test time despite deeper layers, partly because of more anchor scales (5 x 5) but thin structure.

~\\
\textbf{Data augmentation.} Commonly, adopting data augmentation in deep learning can expand training samples, avoid over-fitting and improve test accuracy, especially for small-scale training sets. Faster R-CNN also adopts a horizontal flip to augment training set. Empirically, we append a vertical flip to further expand training data. As comparison, we remove all data augmentations and only use original data in training-stage. Table \ref{tab:table3} shows us that adopting horizontal flip or vertical flip alone does increase mAPs. However, there is no benefit to further append vertical flip after a horizontal flip.

 \begin{table*}[t!]
 	\centering
 	\tiny
 	\caption{The effect of data augmentation on test precision. The network is ZF using a anchor scales of $\{ 32^{2}, ~64^{2}, ~128^{2}, ~256^{2}, ~512^{2} \}$ and a aspect ratios of \{1:1, 1:2, 2:1\}.}
 	\begin{tabular}{c|c|ccccccc}
 		\multirow{2}[1]{*}{flip types} & \multirow{2}[1]{*}{mAP} &  \multirow{2}[1]{*}{eryth} & \multirow{2}[1]{*}{leuko} & \multirow{2}[1]{*}{epith} & \multirow{2}[1]{*}{cryst} & \multirow{2}[1]{*}{cast} & \multirow{2}[1]{*}{mycete} & \multirow{2}[1]{*}{epithn} \\
 		&       &        &       &       &       &       &       &  \\
 		\midrule
 		no flip & 0.748 & \textbf{0.865} & 0.819 & 0.826 & 0.764 & 0.582 & 0.844 & 0.533 \\
 		\midrule
 		only horizontal flip & \textbf{0.779} & 0.859 & 0.805 & 0.854 & 0.847 & 0.657 & \textbf{0.863} & 0.57 \\
 		\midrule
 		only verticle flip & 0.767 & 0.853 & \textbf{0.827} & \textbf{0.855} & \textbf{0.879} & 0.647 & 0.858 & 0.448 \\
 		\midrule
 		horizontal and verticle flip & 0.742 & 0.756 & 0.795 & 0.836 & 0.771 & \textbf{0.677} & 0.763 & \textbf{0.599} \\
 	\end{tabular}%
 	\label{tab:table3}%
 \end{table*}%
\begin{table*}[t!]
	\centering
	\tiny
	\caption{Comparisons on ZF net using different anchor scales when adding a multi-scale structure from MS-FRCNN.}
	\begin{tabular}{c|c|c|ccccccc|c}
		\multirow{2}[1]{*}{method} & \multirow{2}[1]{*}{anchor scales} & \multirow{2}[1]{*}{mAP} & \multirow{2}[1]{*}{eryth} & \multirow{2}[1]{*}{leuko} & \multirow{2}[1]{*}{epith} & \multirow{2}[1]{*}{cryst} & \multirow{2}[1]{*}{cast} & \multirow{2}[1]{*}{mycete} & \multirow{2}[1]{*}{epithn} & test time \\
		&       &       &       &       &       &       &       &       &       &  (sec/img) \\
		\midrule
		FRCNN & $\{ 128^{2}, ~256^{2}, ~512^{2} \}$ & 0.723 & 0.607 & 0.749 & 0.845 & 0.856 & 0.658 & 0.781 & 0.566 & 0.044 \\
		MS-FRCNN & $\{ 128^{2}, ~256^{2}, ~512^{2} \}$ & 0.712 & 0.601 & 0.747 & 0.817 & 0.822 & 0.61  & 0.781 & 0.607 & 0.075 \\
		\midrule
		FRCNN & $\{ 64^{2}, ~128^{2}, ~256^{2}, ~512^{2} \}$ & 0.796 & 0.853 & 0.809 & 0.855 & 0.858 & 0.671 & 0.861 & 0.665 & 0.045 \\
		MS-FRCNN & $\{ 64^{2}, ~128^{2}, ~256^{2}, ~512^{2} \}$ & 0.756 & 0.845 & 0.811 & 0.824 & 0.835 & 0.639 & 0.815 & 0.512 & 0.08 \\
		\midrule
		FRCNN & $\{ 32^{2}, ~64^{2}, ~128^{2}, ~256^{2}, ~512^{2} \}$ & 0.779 & 0.859 & 0.805 & \textbf{0.854} & \textbf{0.847} & \textbf{0.657} & 0.863 & 0.57  & 0.046 \\
		MS-FRCNN & $\{ 32^{2}, ~64^{2}, ~128^{2}, ~256^{2}, ~512^{2} \}$ & 0.775 & \textbf{0.867} & \textbf{0.81} & 0.836 & 0.809 & 0.646 & \textbf{0.871} & \textbf{0.589} & 0.077 \\
	\end{tabular}%
	\label{tab:table4}%
\end{table*}%

~\\
\textbf{MS-FRCNN.} Generally, Faster R-CNN could obtain excellent performance on several natural image benchmarks \cite{everingham2010pascal,lin2014microsoft,russakovsky2015imagenet} that hold objects almost occupying the majority of an image. But as mentioned in the previous section, most objects in urine sediment micro-images are small and low-resolution. Faster R-CNN only uses one higher convolutional layer as feature map, which hardly detects some small objects because of bigger stride and larger receptive field size. Therefore, inspired by \cite{long2015fully} that combines semantic information from a deep, coarse layer with appearance information from a shallow, fine layer for accurate and detailed segmentations, there have been proposed several multi-scale approaches \cite{bell2016inside,hoang2016multiple,liu2016ssd,kong2016hypernet,cai2016unified,kim2016pvanet}, so that the size of receptive field could match various-size objects, especially small instances.

In order to validate the effectiveness of multi-scale methods for urine particles recognition, we conduct a series of experiments based on the MS-FRCNN architecture. The final results are shown in table \ref{tab:table4}. Overall, MS-FRCNN takes more test time per image and the mAPs are worse a bit than original Faster R-CNN (FRCNN). But we can get an interesting observation from the table, as the number of anchors increases, the final gap between precisions becomes smaller (a difference of 0.4\%). In addition, the accuracy of small objects (i.e., eryth, leuko, and epithn) is more superior than no multi-scale. It is a mention that the PVANet also contains a multi-scale structure \cite{kong2016hypernet}, but different from MS-FRCNN used in this part. We argue that the excellent performance of PVNet partly benefits from it.

~\\
\textbf{OHEM-FRCNN.}  We choose Faster R-CNN as a base object detector and embed the novel bootstrapping technique, online hard example mining (OHEM). As reported in table \ref{tab:table5}, OHEM improves the mAP of Faster R-CNN (FRCNN) from 79.5\% to 81\% while taking approximately the same test time. Specifically, all categories except leukocyte yield better APs, where erythrocyte, cast and epithelial nuclei benefit more. In addition, the gains from OHEM can be increased by enlarging and complicating training set. 

\subsubsection{Urine particles recognition based on SSD}
When training SSD, we fine-tune a pre-trained model with SGD for 120k mini-batch iterations, with a mini-batch size of 32 on 1 GPU (a mini-batch size of 16 on 2 GPU during SSD500 training), a momentum of 0.9 and a weight decay of 0.0005. By default, we adopt the multistep learning rate policy with a base learning rate of 0.001 (0.01 when use batch normalization for all newly added layers), a stepvalue of [80,000, 10,000, 120,000] and a gamma of 0.1.

\begin{table*}[t!]
	\centering
	\tiny
	\caption{Comparisons between FRCNN and OHEM-FRCNN on VGG-16 net using the same anchor scales of $\{ 32^{2}, ~64^{2}, ~128^{2}, ~256^{2}, ~512^{2} \}$.}
	\begin{tabular}{c|c|ccccccc|c}
		\multirow{2}[1]{*}{method} & \multirow{2}[1]{*}{mAP} & \multirow{2}[1]{*}{eryth} & \multirow{2}[1]{*}{leuko} & \multirow{2}[1]{*}{epith} & \multirow{2}[1]{*}{cryst} & \multirow{2}[1]{*}{cast} & \multirow{2}[1]{*}{mycete} & \multirow{2}[1]{*}{epithn} & test time \\
		&       &       &       &       &       &       &       &       &  (sec/img) \\
		\midrule
		FRCNN & 0.795 & 0.854 & 0.825 & 0.857 & 0.851 & 0.724 & 0.876 & 0.576 & 0.104 \\
		OHEM-FRCNN & 0.810 & 0.871 & 0.807 & 0.866 & 0.859 & 0.755 & 0.877 & 0.633 & 0.115 \\
	\end{tabular}%
	\label{tab:table5}%
\end{table*}%
\begin{table*}[t!]
	\centering
	\tiny
	\caption{Detection results using SSD model, where SSD300 has an input size of 300 x 300, SSD500 increases it to 500 x 500, and the penultimate row, SSD300$^{*}$, represents a Trimmed SSD removing conv7, conv8, and conv9 layers.}
	\begin{tabular}{c|c|c|c|ccccccc|c}
		\multirow{2}[1]{*}{SSD model} & \multirow{2}[1]{*}{$S_{min}$} & \multirow{2}[1]{*}{$S_{max}$} & \multirow{2}[1]{*}{mAP} & \multirow{2}[1]{*}{eryth} & \multirow{2}[1]{*}{leuko} & \multirow{2}[1]{*}{epith} & \multirow{2}[1]{*}{cryst} & \multirow{2}[1]{*}{cast} & \multirow{2}[1]{*}{mycete} & \multirow{2}[1]{*}{epithn} & test time  \\
		&       &       &       &       &       &       &       &       &       &       & (sec/img) \\
		\midrule
		\multirow{2}[2]{*}{SSD300} & 0.2   & 0.9   & 0.732 & 0.841 & \textbf{0.764} & 0.828 & 0.745 & 0.559 & 0.797 & 0.587 & 0.021 \\
		& 0.1   & 0.9   & 0.752 & 0.766 & 0.741 & \textbf{0.838} & \textbf{0.782} & 0.7   & 0.839 & 0.596 & 0.021 \\
		\midrule
		SSD300$^{*}$ & 0.2   & 0.9   & \textbf{0.773} & \textbf{0.846} & 0.748 & 0.837 & 0.772 & \textbf{0.721} & \textbf{0.85} & \textbf{0.638} & 0.021 \\
		\midrule
		SSD500 & 0.1   & 0.9   & 0.658 & 0.557 & 0.609 & 0.834 & 0.632 & 0.669 & 0.792 & 0.512 & 0.047 \\
	\end{tabular}%
	\label{tab:table6}%
\end{table*}%

~\\
\textbf{Different scales of the default boxes.} We have known that SSD discretizes the output space of bounding boxes into a set of default boxes over different aspect ratios and scales at each feature map cell. In order to relate these default boxes from different feature maps to corresponding receptive fields, the authors in paper \cite{liu2016ssd} designed a scale strategy that regularly but roughly responses specific boxes to specific areas of the image, where the lowest feature map has a minimum scale of $ S_{min} $  and the highest feature map has a maximum scale of $ S_{max} $, and all other feature maps in between are regularly scattered (more details, please refer to the paper \cite{liu2016ssd}).

Considering lots of small particles in urine sediment images, we adjust empirically the scales of default boxes when training SSD300. From experimental results in table \ref{tab:table6}, we can see that decreasing the minimum scale of 0.2 to 0.1 (the maximum scale of 0.9 remains unchanged.) increases mAP by 2\% in which the AP of cast increases by 14.1\%.

~\\
\textbf{Trimmed SSD.} In order to reduce the complexity of the original SSD model and avert over-fitting on the small-scale urinalysis database, we take a pruning strategy, Trimmed SSD. Specifically, in this experiment we remove conv7, conv8, and conv9 layers of SSD300.
The penultimate row in table \ref{tab:table6} shows us that deleting the conv7, conv8, and conv9 layers does yield a better mAP (a boost of 4.1\%), but no better speed.
                 
~\\
\textbf{Different input sizes.} Generally, increasing the size of input images can improve detection accuracy, especially to small objects. Imitating the paper \cite{liu2016ssd}, we also increase the input size from 300 x 300 to 500 x 500. Here we train SSD500 only once, with a minimum scale of 0.1 and a maximum scale of 0.9. Unfortunately, we just obtain a poor 65.8\% mAP (the last row in table \ref{tab:table6}), because of decreasing the batch size setting from 32 to 16 to run this model in limited GPU resources. We argue that better results can be achieved if increase the setting of batch size.

~\\
So far, there is still much room for SSD improvement. Given the boost of Trimmed SSD, we have glimpsed the structural deficiencies of original SSD model, but lack more sophisticated analysis for further research. Also, we can design a more suitable default boxes distribution over different scales and aspect ratios to fit the urinalysis database. They are interesting and open questions for us to future study.

~\\
The above experimental results clearly make out that Faster R-CNN outperforms SSD in precision (Figure \ref{fig:detections} also shows this in appendix), but the latter is faster. In view of the importance of accuracy in medical applications, Faster R-CNN will be an priority for automated urinalysis system.

\section{Adding bells \& whistles}
As mentioned, Faster R-CNN exceeds SSD in detection accuracy, especially on PVANet with fast detection speed by a significant margin. In structure, Faster R-CNN mainly contains a region proposal network (RPN) than SSD for ROIs generation. In this section, we study in detail the impact of several factors to region proposal generation while combining above experiments. Further, we also compare PVANet against VGG-16 on specific detection performances more intuitively.

\subsection{Analysis for region proposal generation}

\begin{figure*}[t!]
      	\centering
      	\subfigure[]{\label{fig:anchors}
      		\includegraphics[height=5cm,width=6cm]{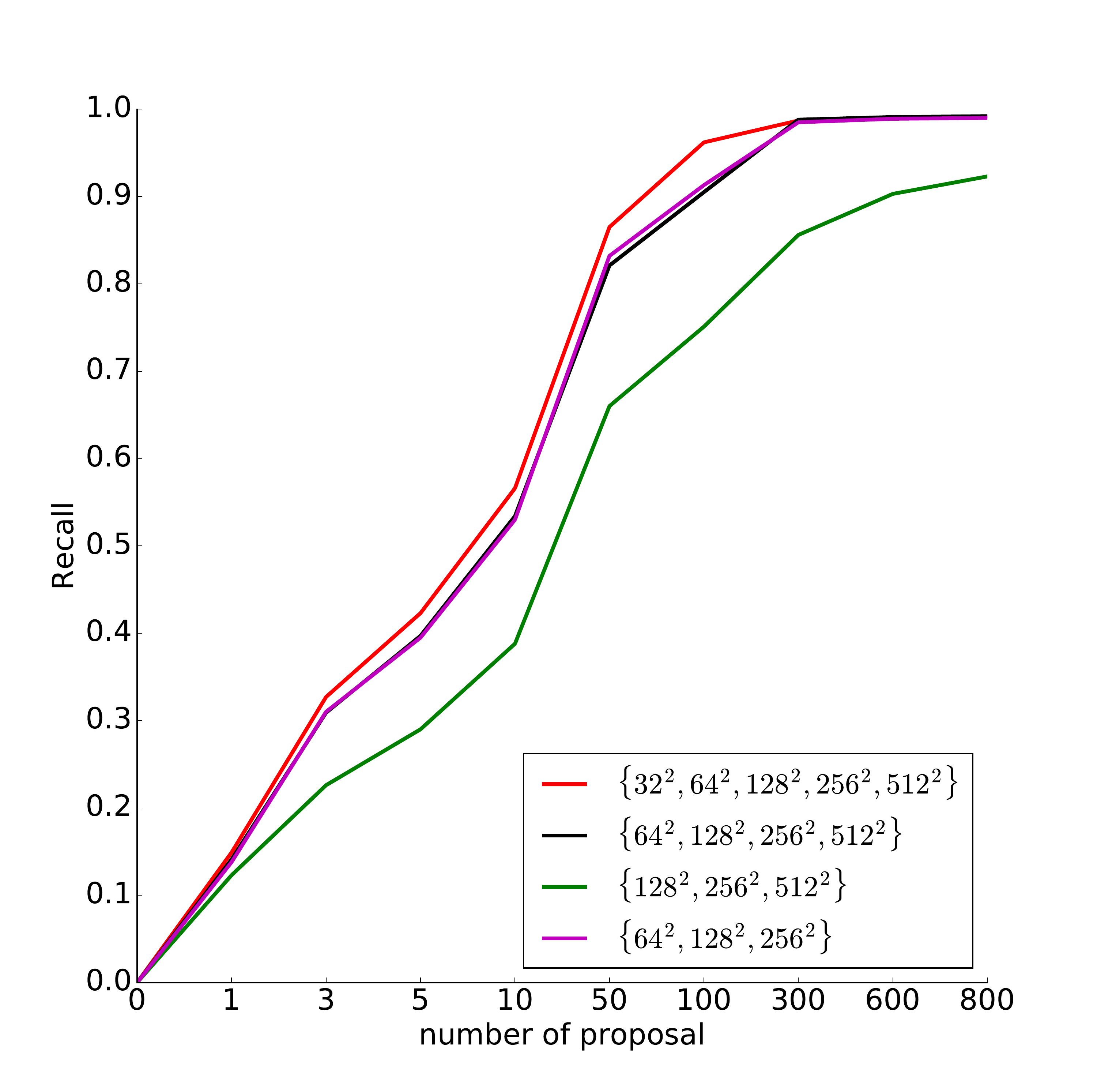}} 
      	\hspace{-5ex}    	
      	\subfigure[]{\label{fig:nets}
      		\includegraphics[height=5cm,width=6cm]{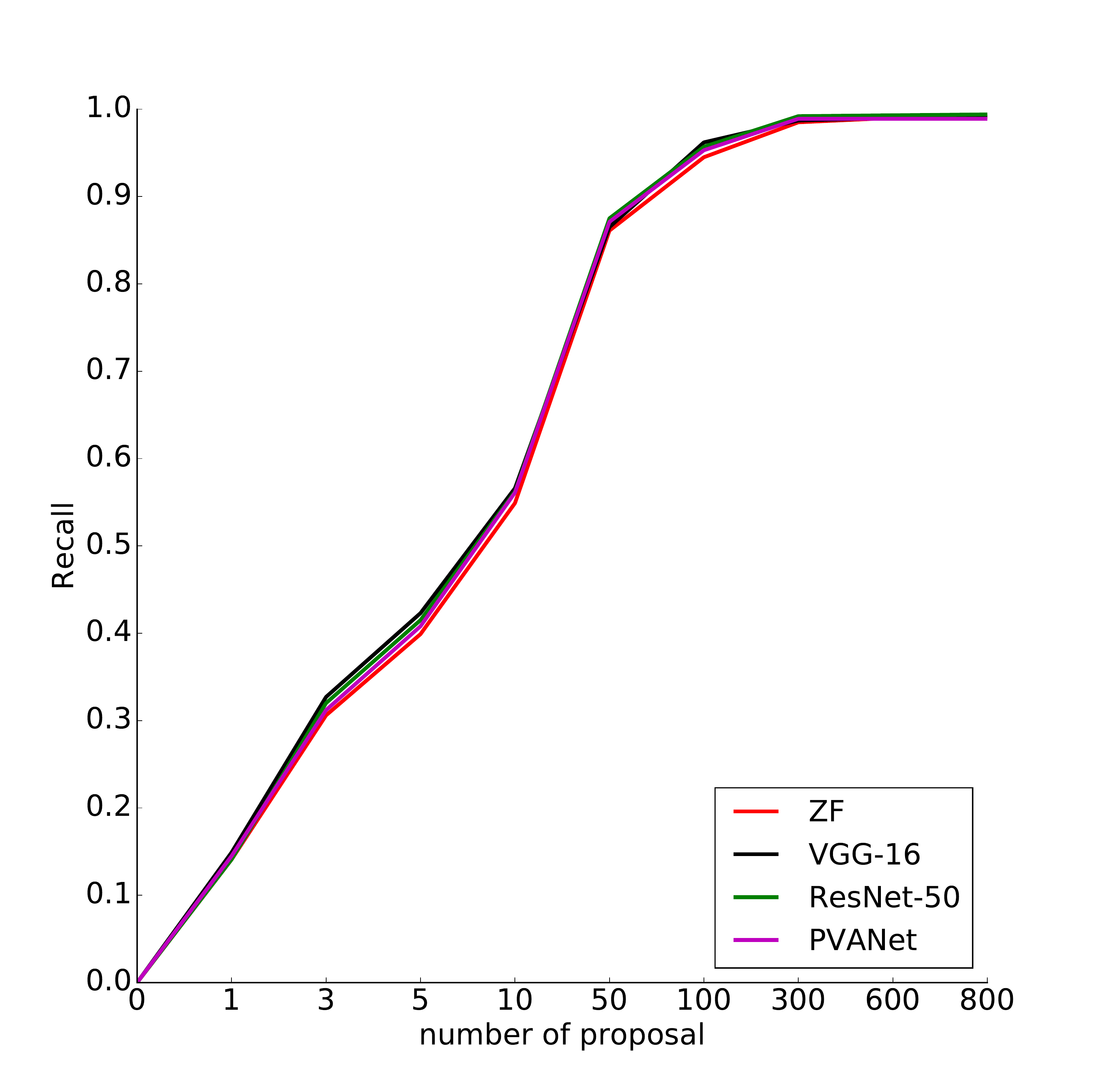}}
      	\hspace{-5ex}        	
      	\subfigure[]{\label{fig:iou}
      		\includegraphics[height=5cm,width=6cm]{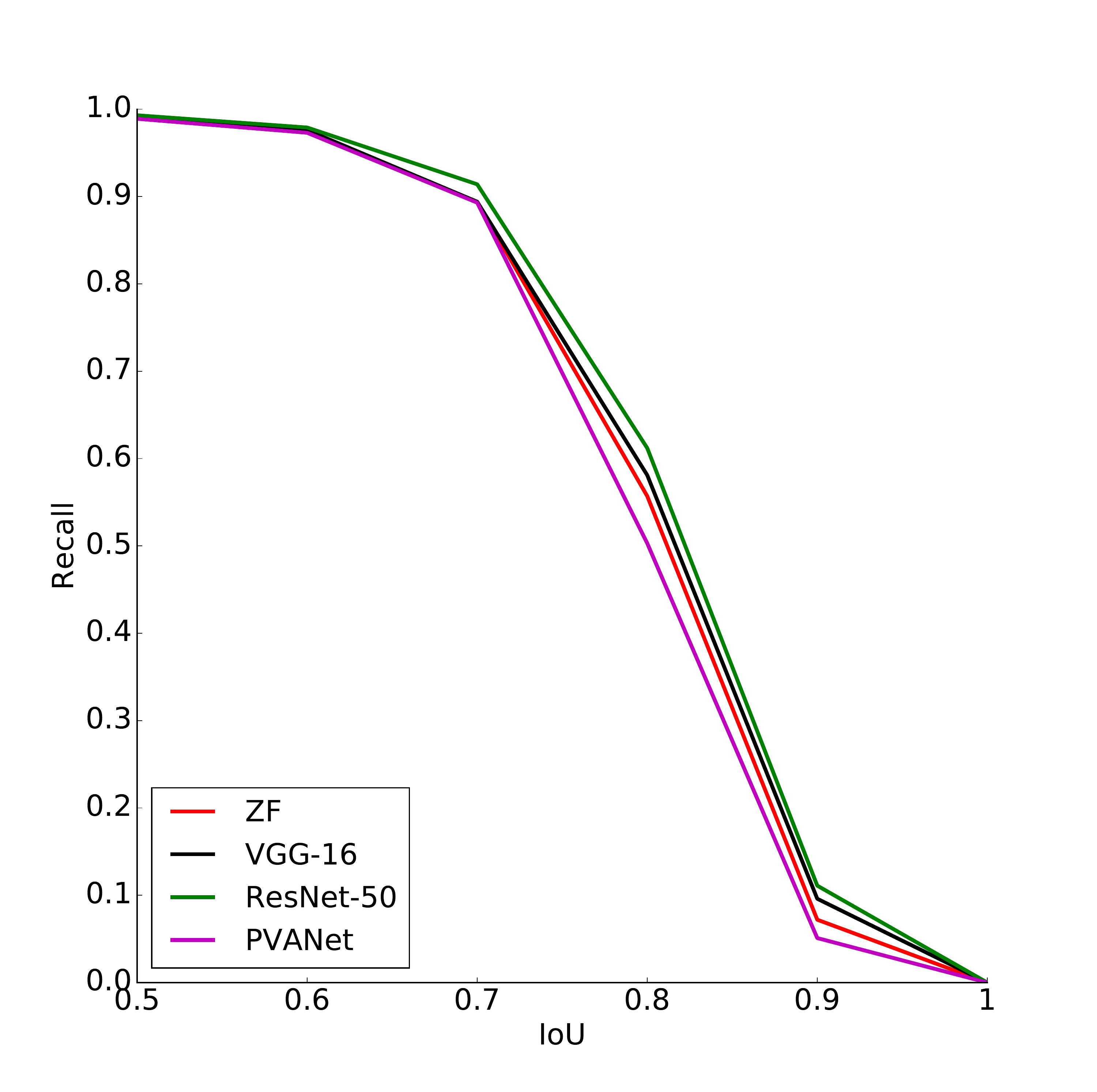}}
      	\hspace{-5ex}        	   	
      	\caption{Analysis for region proposal generation on urinalysis database. (a) Recall versus number of proposal for different anchor scales using VGG-16 with a fixed IoU of 0.5. (b) Recall versus number of proposal for different networks with a fixed IoU of 0.5. (c) Recall versus IoU threshold for different networks with a fixed number of proposal (600).}
      	\label{fig：recall}
      \end{figure*}

\textbf{Anchor scales.} In this part, we provide analysis of anchor scales affection to object proposals on VGG-16 net. The curve of recall for anchor scales at different proposal numbers is plotted in figure \ref{fig:anchors}. Correspondingly, the related detection performances are shown in table \ref{tab:table2} (the VGG-16 module). Figure \ref{fig:anchors} displays, from the default anchor scales \{ $128^{2}$, $256^{2}$, $512^{2}$ \}, the proposals recall increases gradually while adding smaller scales (i.e., $64^{2}$ and $32^{2}$), and the scales of \{ $64^{2}$, $128^{2}$, $256^{2}$, $512^{2}$ \} outperforms the scales of \{ $128^{2}$, $256^{2}$, $512^{2}$ \} by a significant margin, closed to the scales of \{ $64^{2}$, $128^{2}$, $256^{2}$, $512^{2}$ \}. The results are consistent with the detection accuracy with respect to mAP \& APs, and indicates two keys about anchor scales: (1) the more the better. (2) the smaller the superior. Reasonable design of anchor scales benefits ROIs generation and final detection. 

\textbf{Networks.} Table \ref{tab:table2} has fully demonstrated the impact of different networks to detection performances with respect to accuracy and speed. Treating RPN as a class-agnostic object detector, we further investigate different networks in terms of proposals quality. Figure \ref{fig:nets}, plotting recall versus number of proposals with a loose IoU of 0.5, shows little differences between several networks when adopting the best anchor scales of \{ $32^{2}$, $64^{2}$, $128^{2}$, $256^{2}$, $512^{2}$ \}. For higher IoU thresholds, shown in figure \ref{fig:iou}, the recall of PVANet drops faster than other networks. Given its superior detection performances, PVANet must obtain more compensation from the latter stage. 

\subsection{PVANet versus VGG-16}
Although PVANet displays under-performances for region proposal generation, eventually it achieves very prominent detection results, a mAP of 84.1\% shown in table \ref{tab:table2}. By inference, we owe it to the latter Fast R-CNN detector. In this part, we compare PVANet against VGG-16 to verify the inference.

\begin{figure*}[t!]
      	\centering
      	\subfigure[VGG-16]{\label{fig:vgg16}
      		\includegraphics[height=6cm,width=6.5cm]{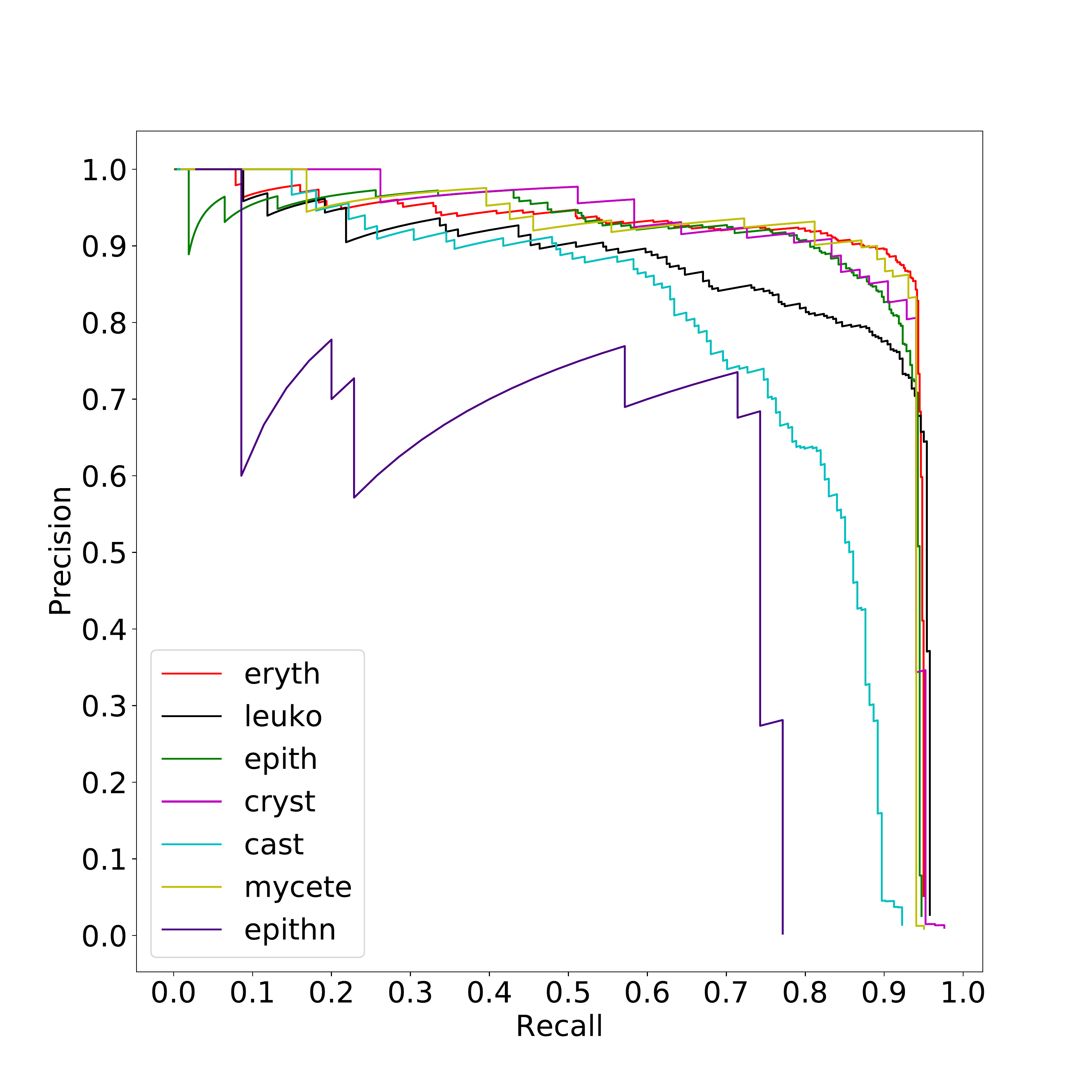}}     	
      	\subfigure[PVANet]{\label{fig:pva}
      		\includegraphics[height=6cm,width=6.5cm]{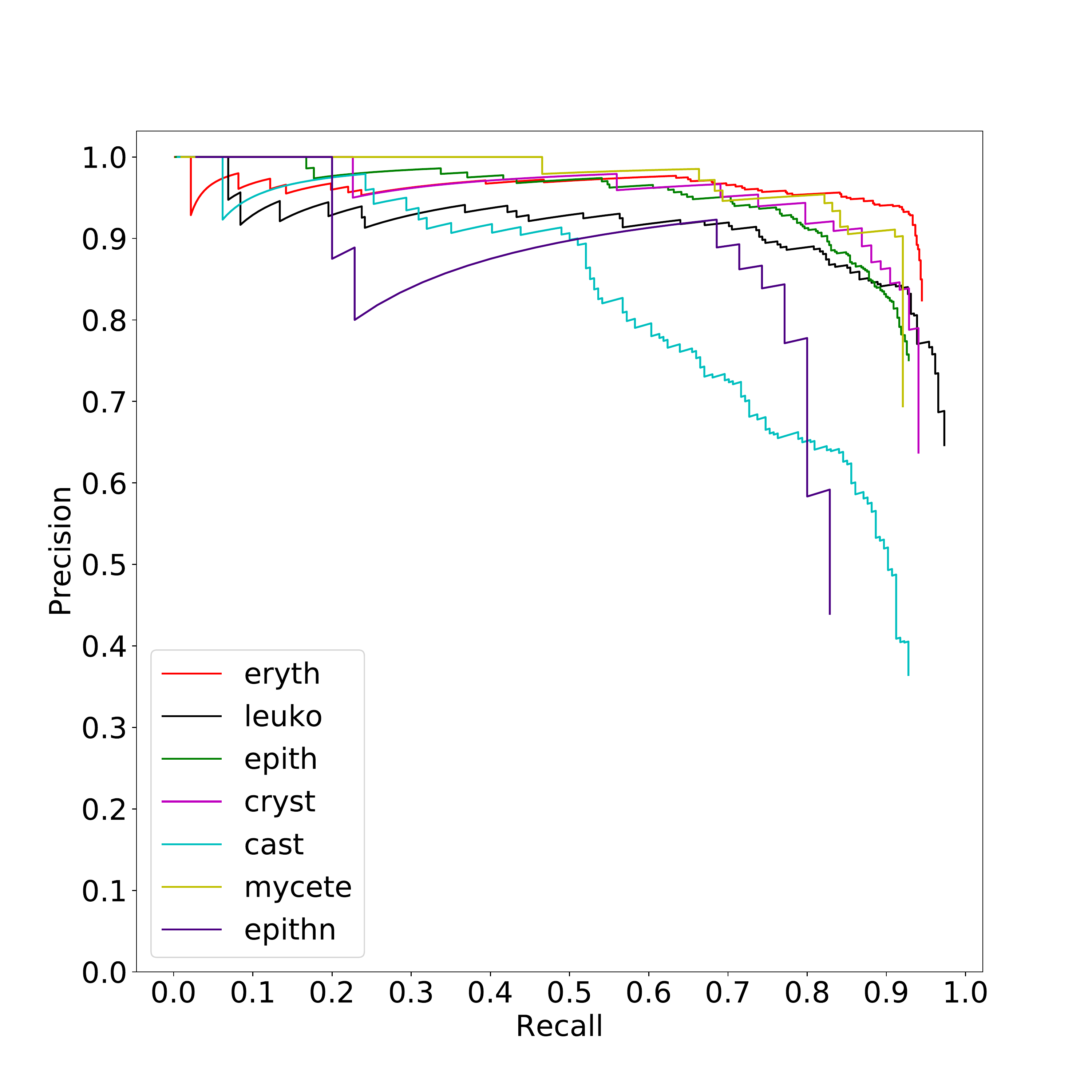}}     	  	   	
      	\caption{Precision versus recall. (a) VGG-16 net with a anchor scales of  \{ $32^{2}$, $64^{2}$, $128^{2}$, $256^{2}$, $512^{2}$ \} and a mAP of 79.5\%. (b) PVANet with a mAP of 84.1\%.}
      	\label{fig：p-r}
      \end{figure*}

Figure \ref{fig：p-r} shows curves of precision-to-recall separately on PVANet and VGG-16 networks. In contrast, PVANet maintains higher precisions stably, as recalls increases. The precisions of VGG-16 net drop sharply in the end. For detections of cast and epithelial nuclei, PVANet also performs better than VGG-16 net.

\section{Conclusions}
In this paper, we treat the urine particles recognition as object detection and select two well-known CNN-based approaches, Faster R-CNN and SSD, as base detection framework. They are segmentation free and can learn task-specific features in an end-to-end manner. When applying Faster R-CNN, SSD, and their variants to urine particles recognition, we effectively adopt the mechanism of deep transfer learning. Moreover, we conduct extensive experimental analysis to demonstrate the impact of various factors, including training strategies, network structures, anchor scales, and so on. After a variety of experiments, we obtain a best mAP of 84.1\% with a test time of 70 ms per image while using Faster R-CNN on PVANet. We believe the result is very meaningful for automatic urine sediment examination and the experimental analysis is instructive for other researchers. Of course, for urine particles recognition, there is still a lot of room to further improve and it will be an interesting future direction for us.

\section*{Acknowledgements}
This research was partially supported by the Natural Science Foundation of Hunan Province, China (No. 14JJ2008) and the National Natural Science Foundation of China under Grant No. 61602522, No. 61573380, No. 61672542.

\section*{Compliance with ethical standards}
          
Funding: It has been stated in the ``Acknowledgments`` section.
~\\
Conflict of Interest: The authors declare that they have no conflict of interest.
~\\
Ethical approval: This article does not contain any studies with human participants or animals performed by any of the authors.

\bibliographystyle{splncs}
\bibliography{usa}


\onecolumn 
\begin{appendices} 
	\section*{Appendix}   
   \begin{figure*}[h!]
   	\centering  	       	
   	\begin{minipage}{0.3\linewidth}
   		\centering
   		\setlength{\abovecaptionskip}{0cm}
   		\setlength{\belowcaptionskip}{0cm}            	
   		\includegraphics[height=3cm,width=4.5cm]{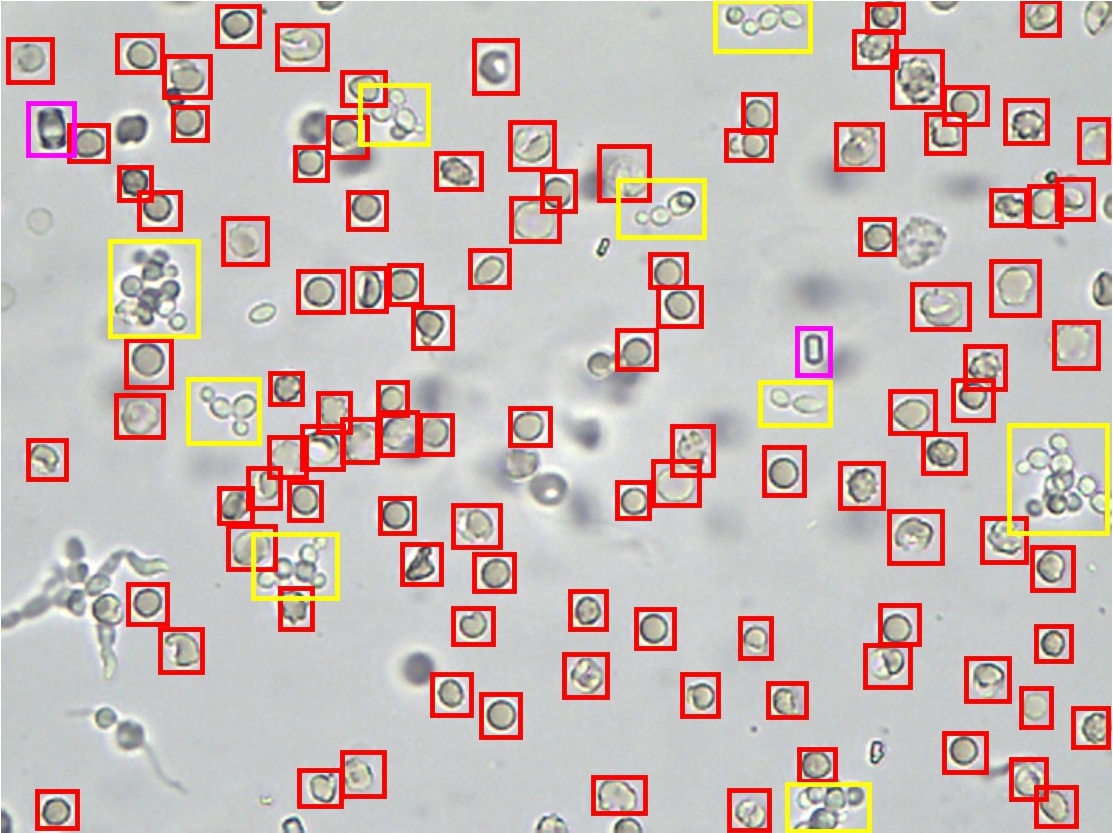}
   		\caption*{(a) annotations}
   	\end{minipage}
   	\hspace{-5ex}
   	\vspace{-0ex}           	
   	\begin{minipage}{0.3\linewidth}
   		\centering              	
   		\setlength{\abovecaptionskip}{0cm}
   		\setlength{\belowcaptionskip}{0cm}
   		\includegraphics[height=3cm,width=4.5cm]{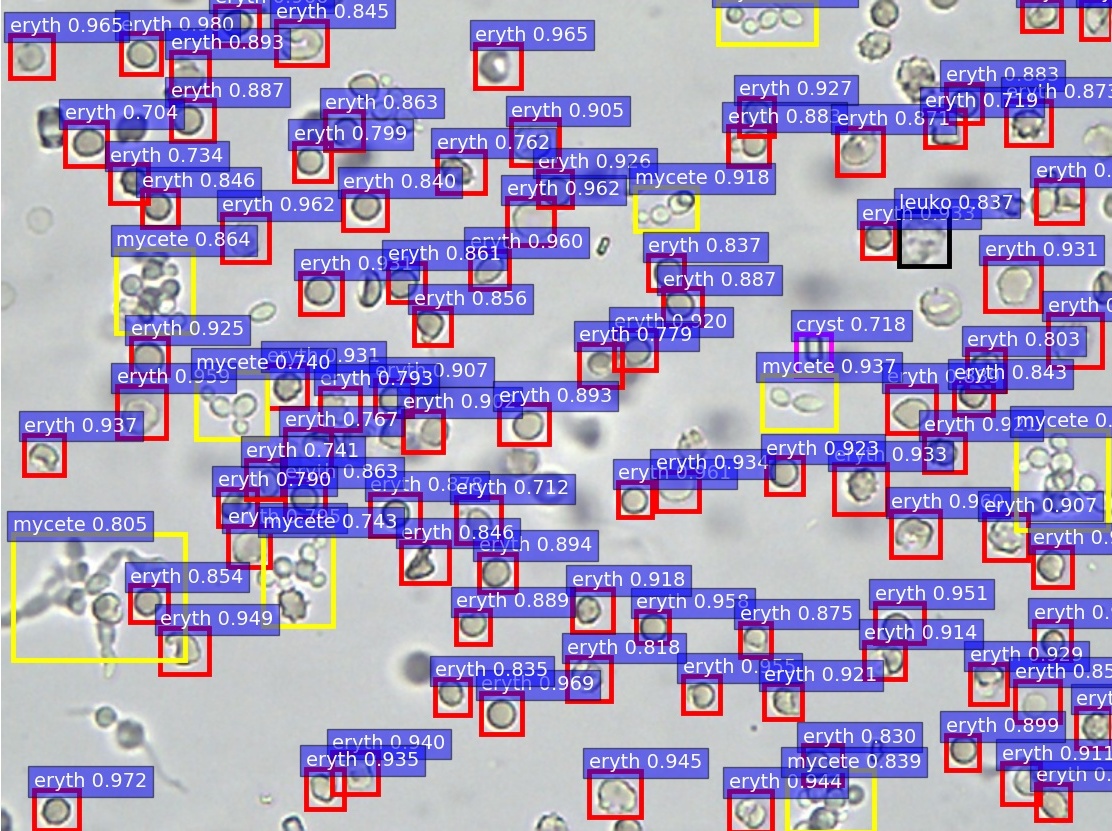}
   		\caption*{(b) ZF}
   	\end{minipage}
   	\hspace{-5ex}
   	\vspace{-0ex}          	
   	\begin{minipage}{0.3\linewidth}
   		\centering           		
   		\setlength{\abovecaptionskip}{0cm}
   		\setlength{\belowcaptionskip}{0cm}
   		\includegraphics[height=3cm,width=4.5cm]{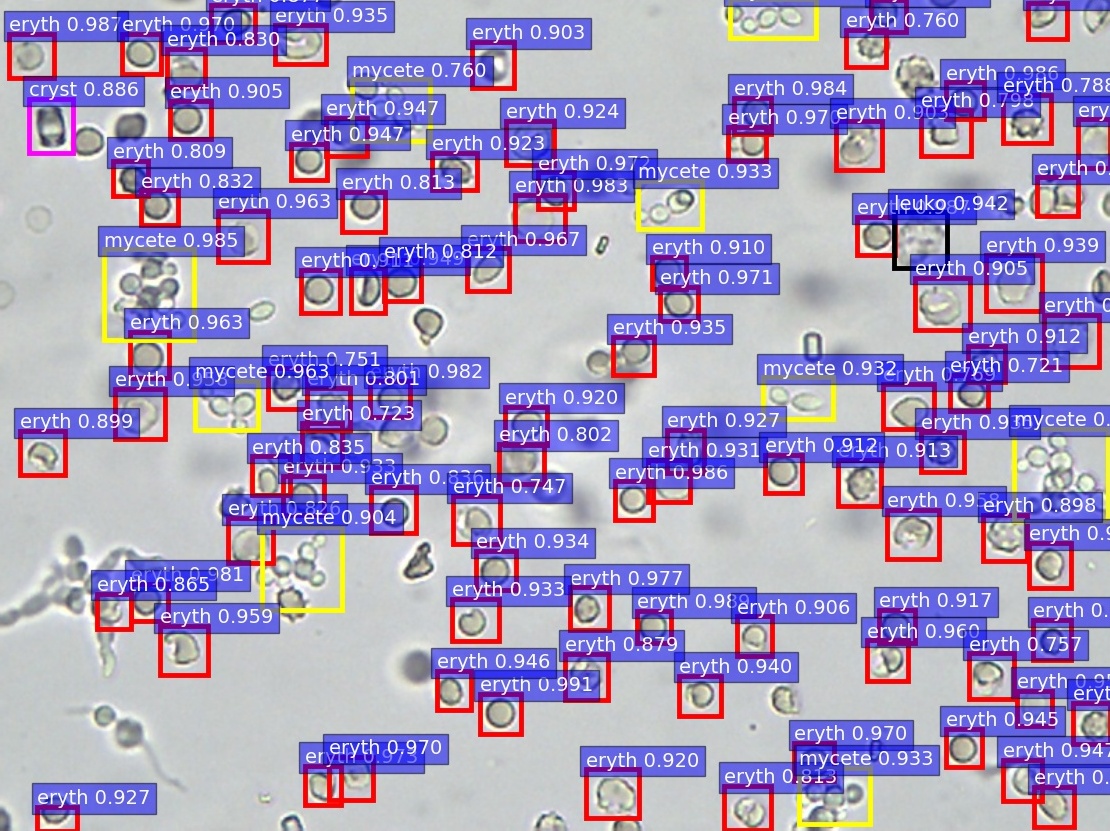}
   		\caption*{(c) VGG-16}
   	\end{minipage}
   	\hspace{-5ex}
   	\vspace{-0ex}          	
   	
   	\begin{minipage}{0.3\linewidth}
   		\centering              	
   		\setlength{\abovecaptionskip}{0cm}
   		\setlength{\belowcaptionskip}{0cm}
   		\includegraphics[height=3cm,width=4.5cm]{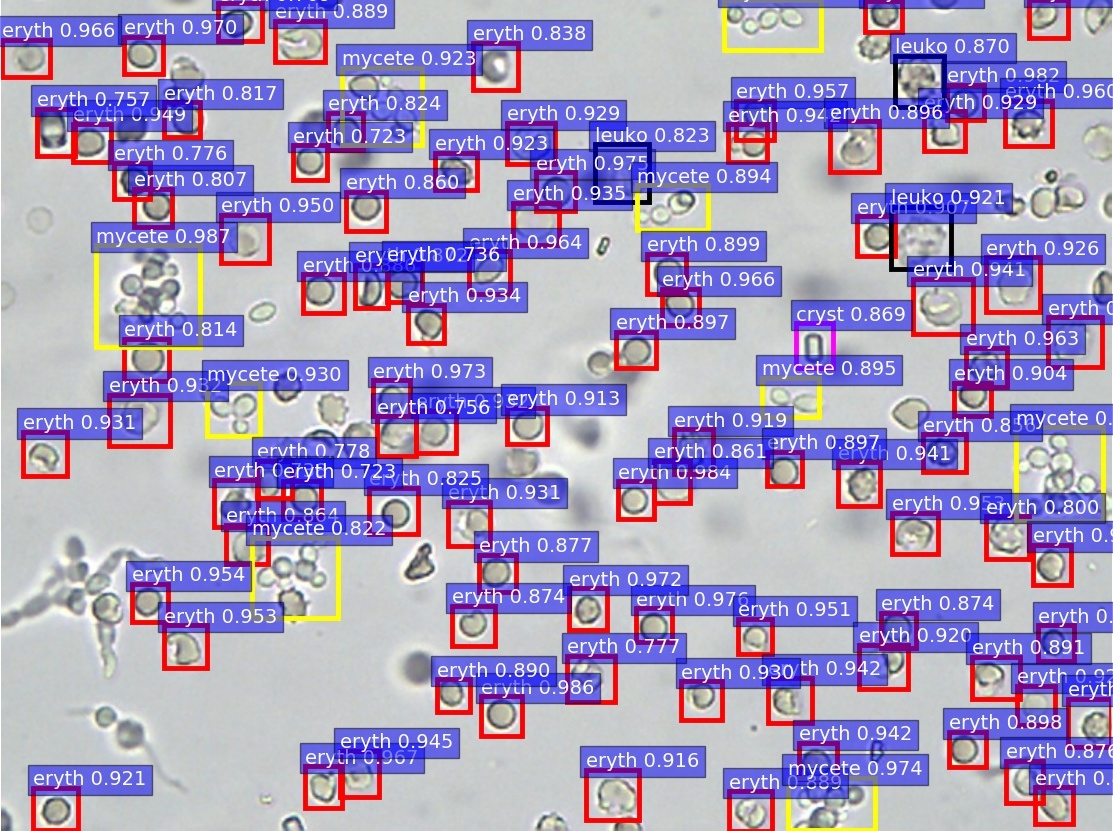}
   		\caption*{(d) ResNet-50}
   	\end{minipage}
   	\hspace{-5ex}
   	\vspace{-0ex}          	            	
   	\begin{minipage}{0.3\linewidth}
   		\centering
   		\setlength{\abovecaptionskip}{0cm}
   		\setlength{\belowcaptionskip}{0cm}
   		\includegraphics[height=3cm,width=4.5cm]{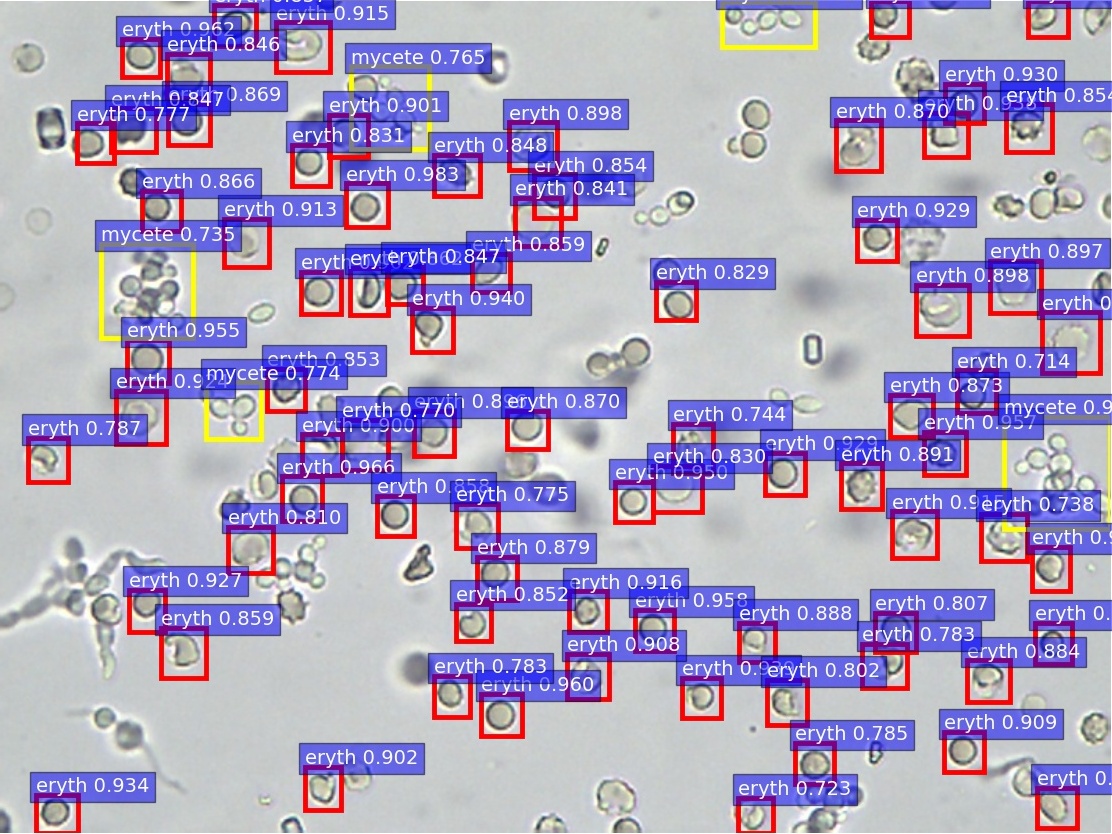}
   		\caption*{(e) PVANet}
   	\end{minipage}
   	\hspace{-5ex}
   	\vspace{-0ex}          	
   	\begin{minipage}{0.3\linewidth}
   		\centering
   		\setlength{\abovecaptionskip}{0cm}
   		\setlength{\belowcaptionskip}{0cm}
   		\includegraphics[height=3cm,width=4.5cm]{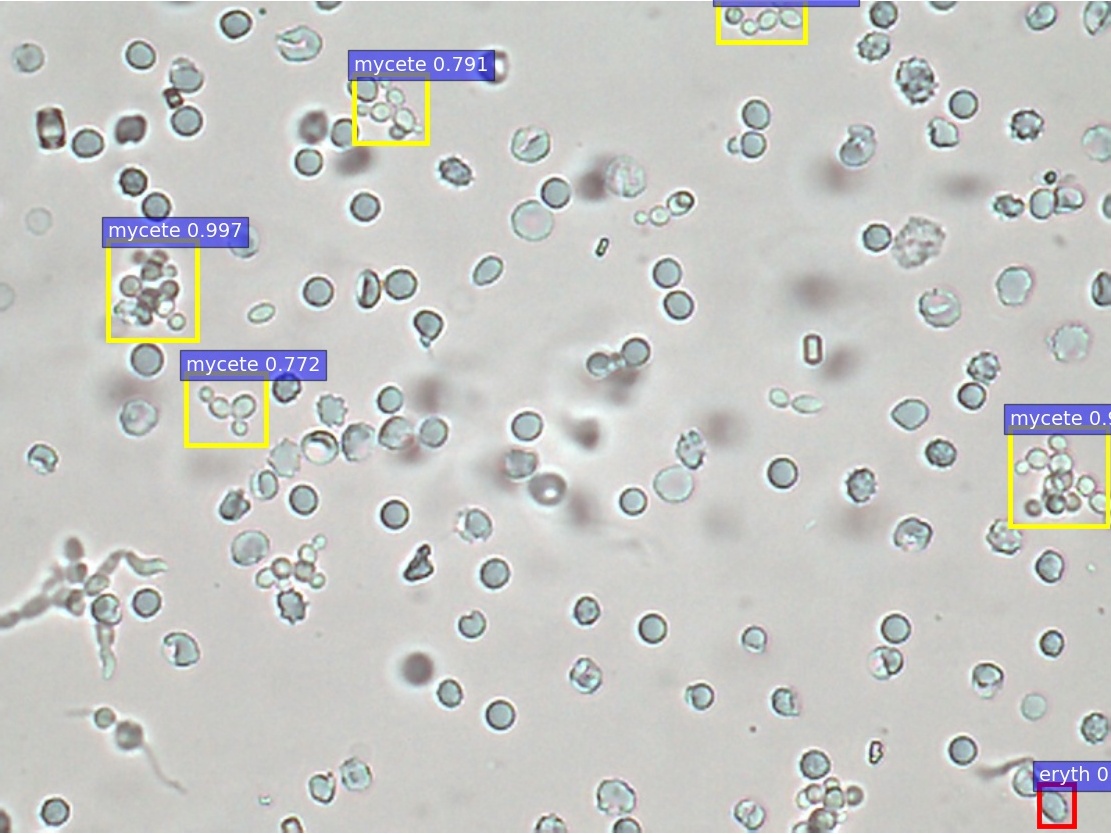}
   		\caption*{(f) SSD300$^{*}$}
   	\end{minipage}
   	\hspace{-5ex}
   	\vspace{-0ex}          	
   	\setlength{\abovecaptionskip}{0cm}
   	\setlength{\belowcaptionskip}{0cm}                     	
   	\caption*{\Romannum{1} : detection results of erythrocyte}
   	\label{fig:eryth}         	
   \end{figure*}
   
   \begin{figure*}[h!]
   	\centering              	
   	\begin{minipage}{0.3\linewidth}
   		\centering
   		\setlength{\abovecaptionskip}{0cm}
   		\setlength{\belowcaptionskip}{0cm}            	
   		\includegraphics[height=3cm,width=4.5cm]{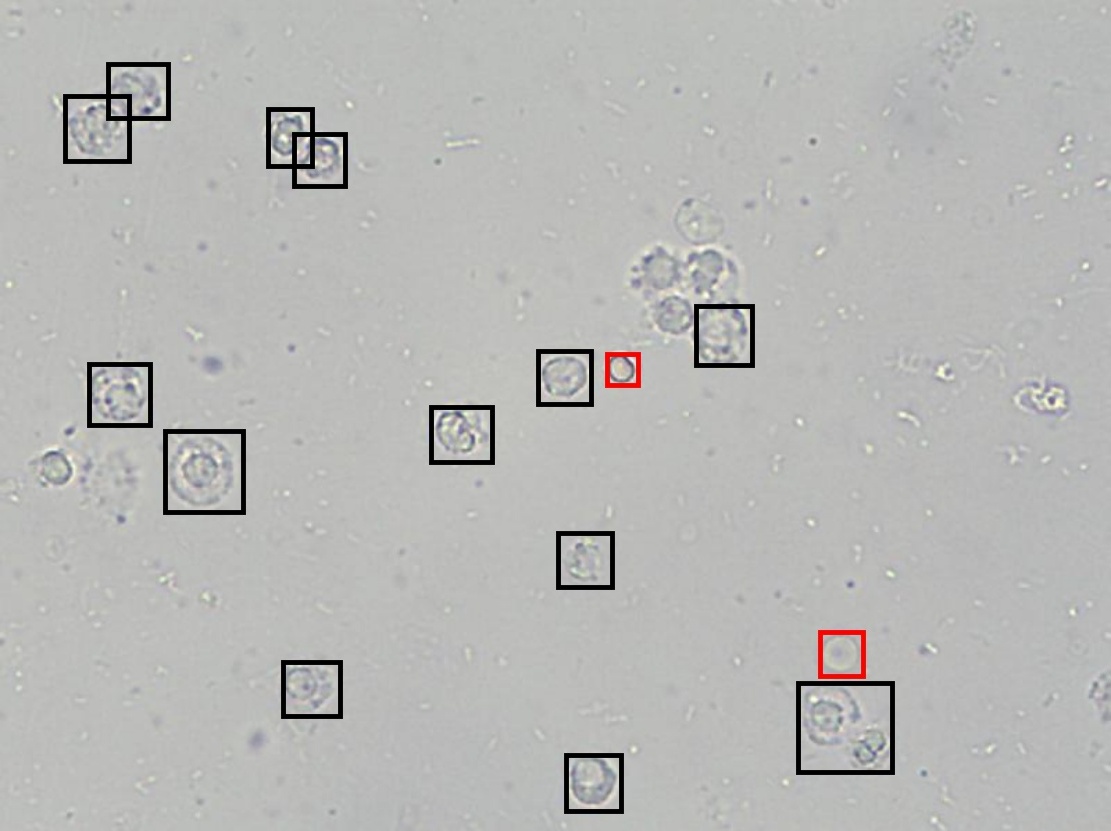}
   		\caption*{(a) annotations}
   	\end{minipage}
   	\hspace{-5ex}
   	\vspace{-0ex}          	
   	\begin{minipage}{0.3\linewidth}
   		\centering              	
   		\setlength{\abovecaptionskip}{0cm}
   		\setlength{\belowcaptionskip}{0cm}
   		\includegraphics[height=3cm,width=4.5cm]{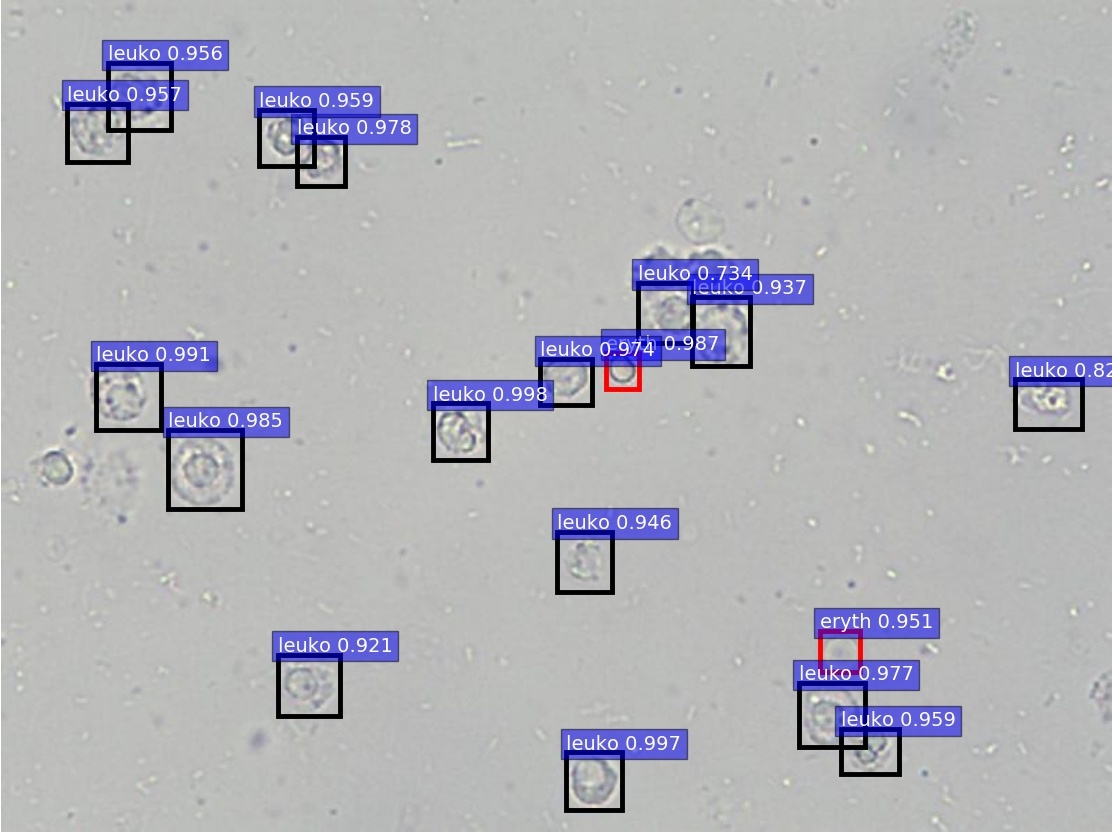}
   		\caption*{(b) ZF}
   	\end{minipage}
   	\hspace{-5ex}
   	\vspace{-0ex}          	
   	\begin{minipage}{0.3\linewidth}
   		\centering           		
   		\setlength{\abovecaptionskip}{0cm}
   		\setlength{\belowcaptionskip}{0cm}
   		\includegraphics[height=3cm,width=4.5cm]{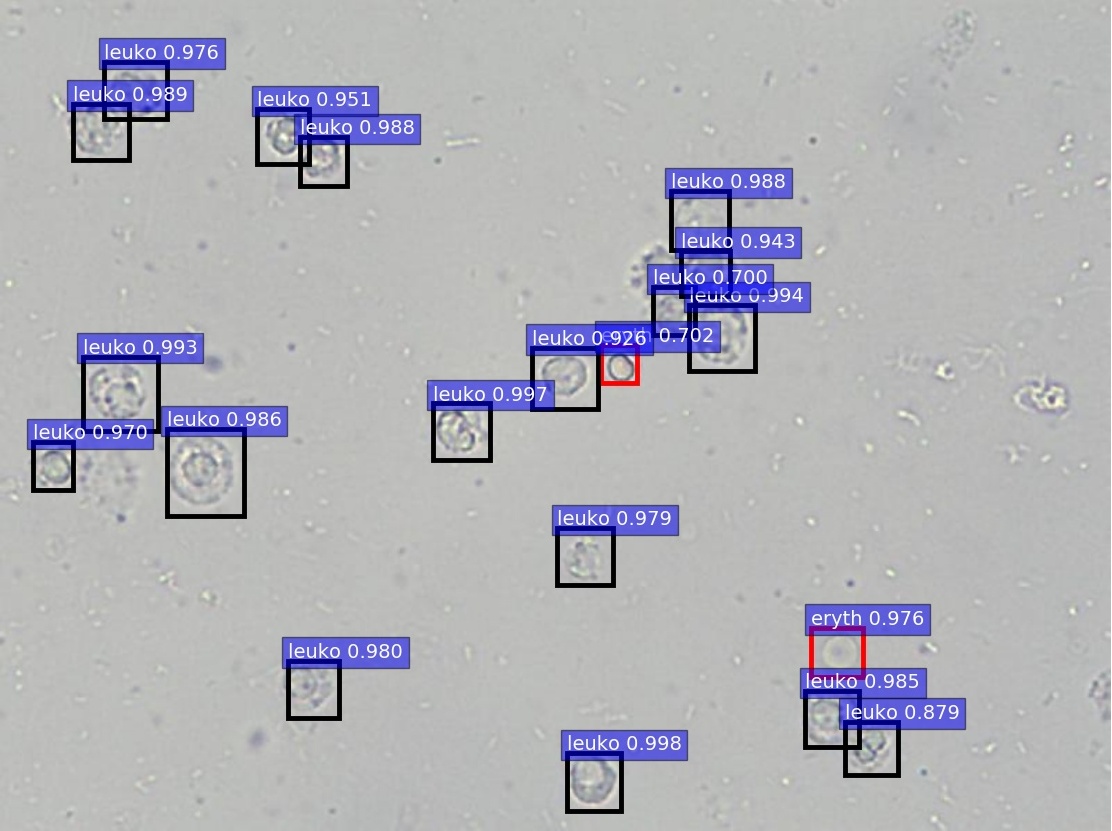}
   		\caption*{(c) VGG-16}
   	\end{minipage}
   	\hspace{-5ex}
   	\vspace{-0ex}          	
   	
   	\begin{minipage}{0.3\linewidth}
   		\centering              	
   		\setlength{\abovecaptionskip}{0cm}
   		\setlength{\belowcaptionskip}{0cm}
   		\includegraphics[height=3cm,width=4.5cm]{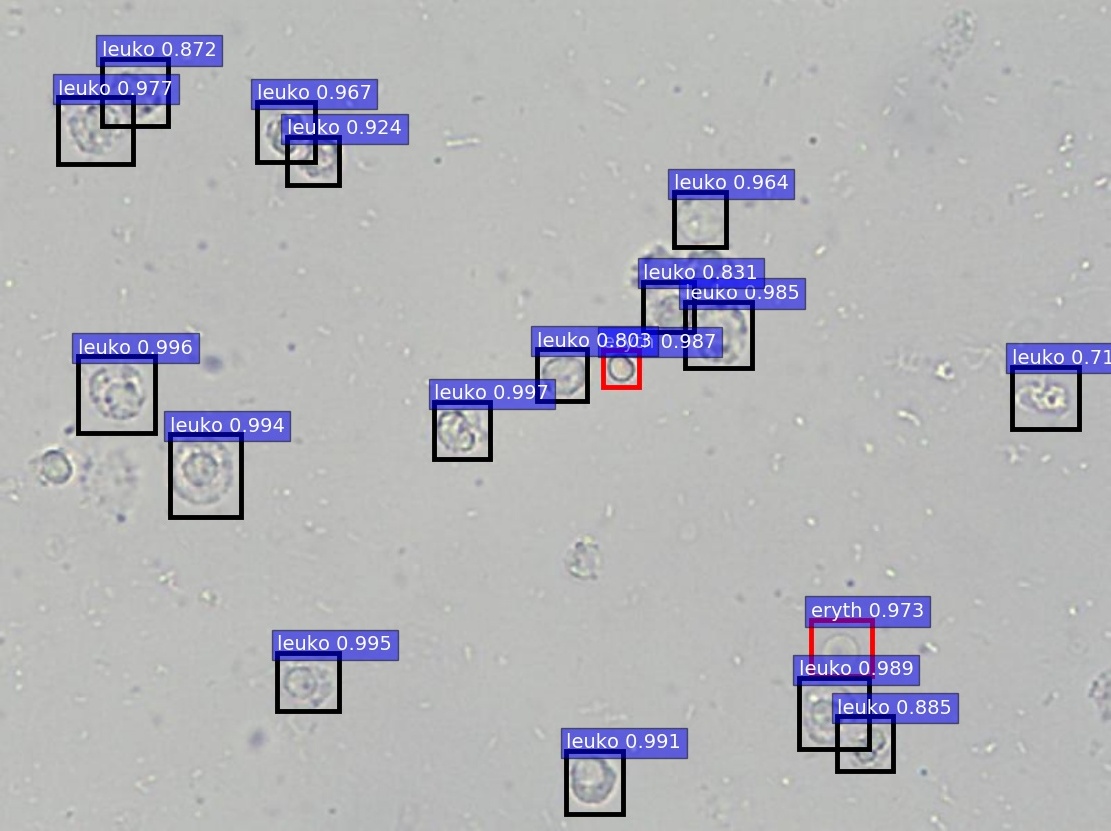}
   		\caption*{(d) ResNet-50}
   	\end{minipage}            	
   	\hspace{-5ex}
   	\vspace{-0ex}
   	\begin{minipage}{0.3\linewidth}
   		\centering
   		\setlength{\abovecaptionskip}{0cm}
   		\setlength{\belowcaptionskip}{0cm}
   		\includegraphics[height=3cm,width=4.5cm]{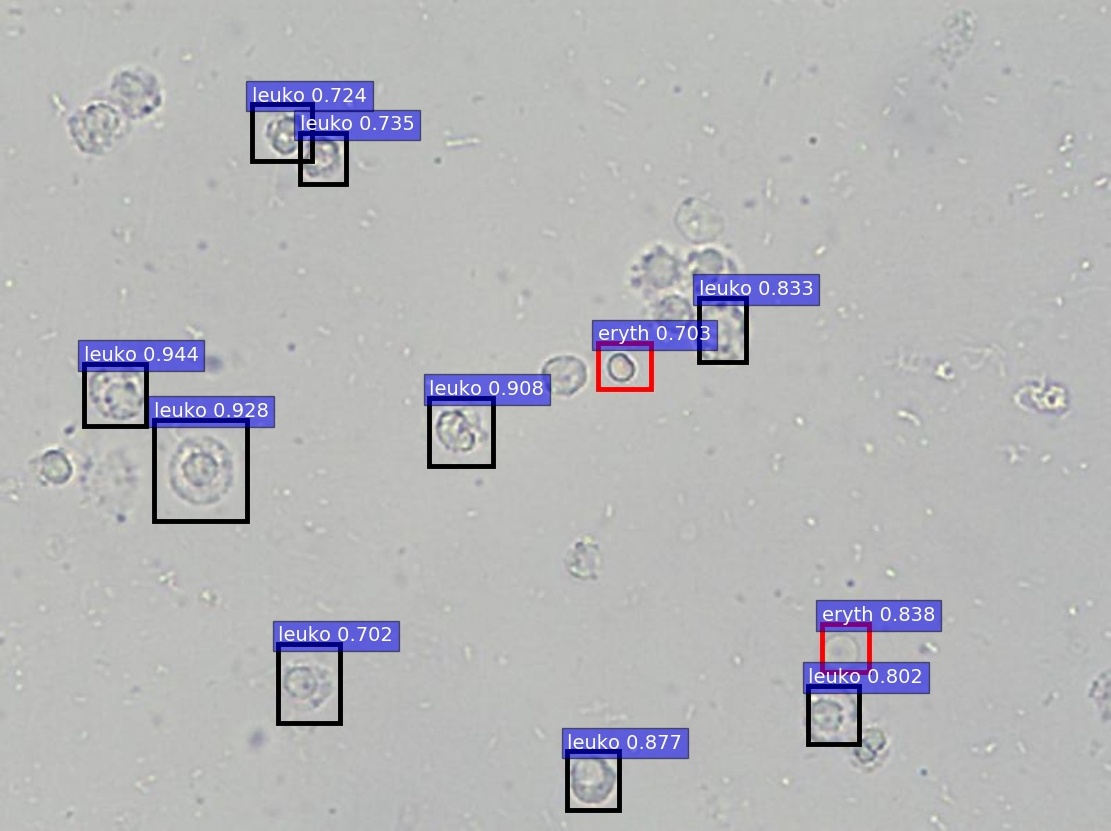}
   		\caption*{(e) PVANet}
   	\end{minipage}
   	\hspace{-5ex}
   	\vspace{-0ex}          	
   	\begin{minipage}{0.3\linewidth}
   		\centering
   		\setlength{\abovecaptionskip}{0cm}
   		\setlength{\belowcaptionskip}{0cm}
   		\includegraphics[height=3cm,width=4.5cm]{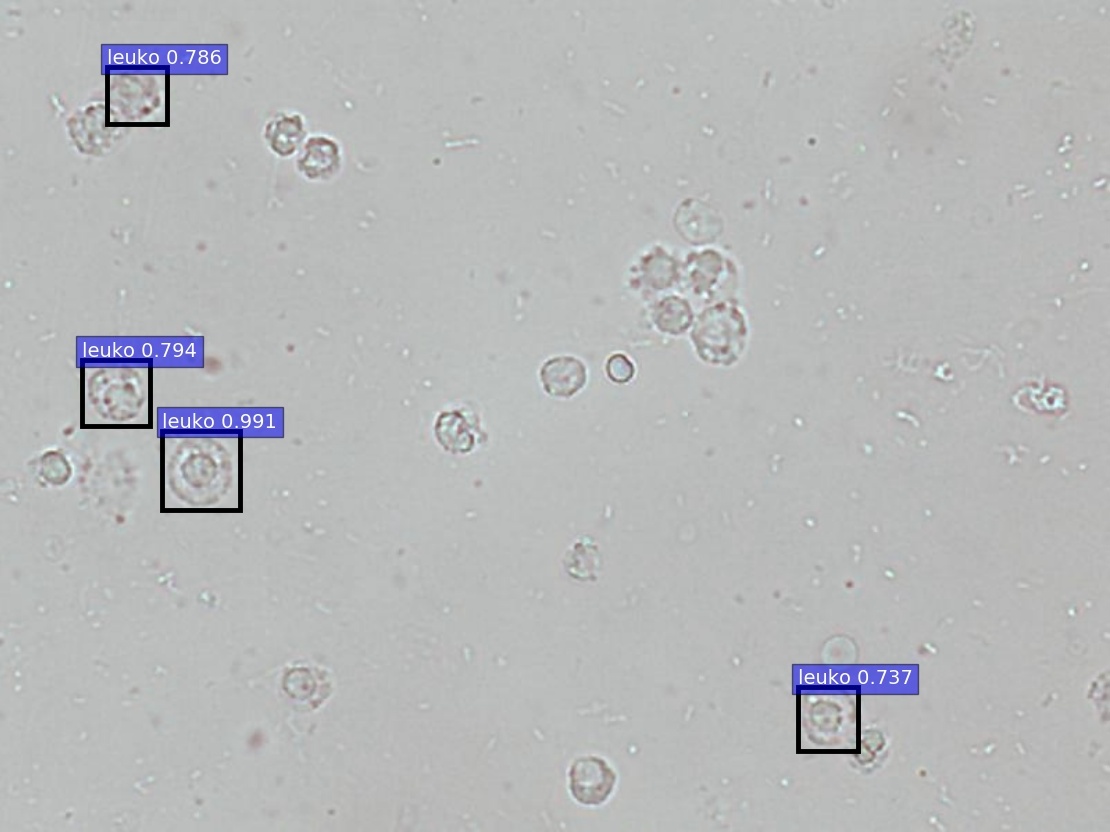}
   		\caption*{(f) SSD300$^{*}$}
   	\end{minipage}
   	\hspace{-5ex}
   	\vspace{-0ex}          	
   	\setlength{\abovecaptionskip}{0cm}
   	\setlength{\belowcaptionskip}{0cm}                     	
   	\caption*{\Romannum{2} : detection results of leukocyte}
   	\label{fig:leuko}         	
   \end{figure*}
   
   \begin{figure*}[t!]
   	\centering
   	\begin{minipage}{0.3\linewidth}
   		\centering
   		\setlength{\abovecaptionskip}{0cm}
   		\setlength{\belowcaptionskip}{0cm}            	
   		\includegraphics[height=3cm,width=4.5cm]{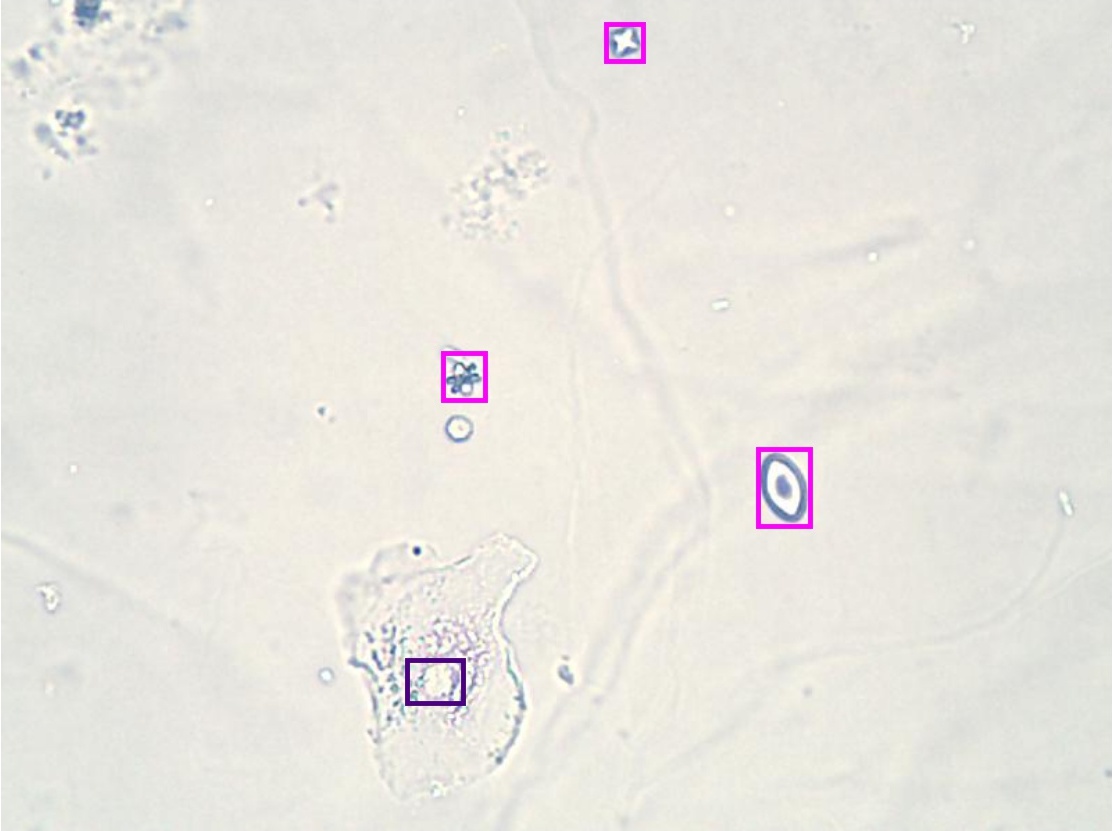}
   		\caption*{(a) annotations}
   	\end{minipage}
   	\hspace{-5ex}
   	\vspace{-0ex}
   	\begin{minipage}{0.3\linewidth}
   		\centering              	
   		\setlength{\abovecaptionskip}{0cm}
   		\setlength{\belowcaptionskip}{0cm}
   		\includegraphics[height=3cm,width=4.5cm]{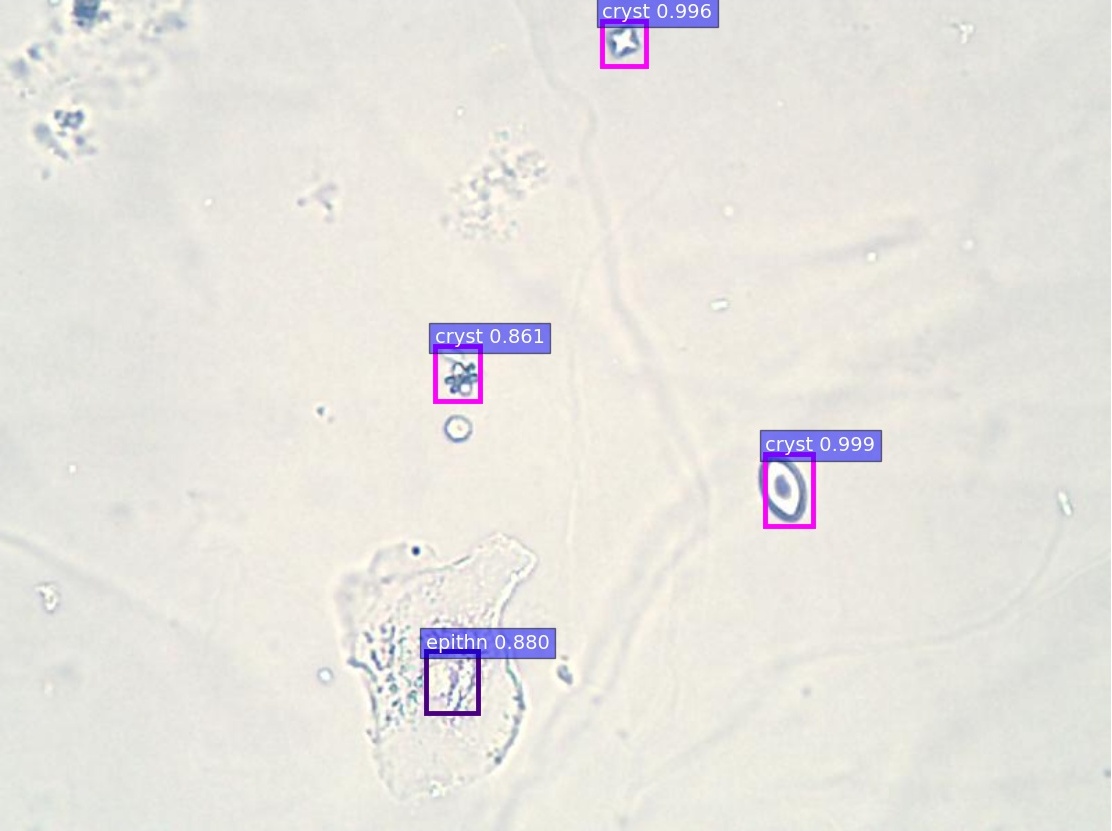}
   		\caption*{(b) ZF}
   	\end{minipage}
   	\hspace{-5ex}
   	\vspace{-0ex}
   	\begin{minipage}{0.3\linewidth}
   		\centering           		
   		\setlength{\abovecaptionskip}{0cm}
   		\setlength{\belowcaptionskip}{0cm}
   		\includegraphics[height=3cm,width=4.5cm]{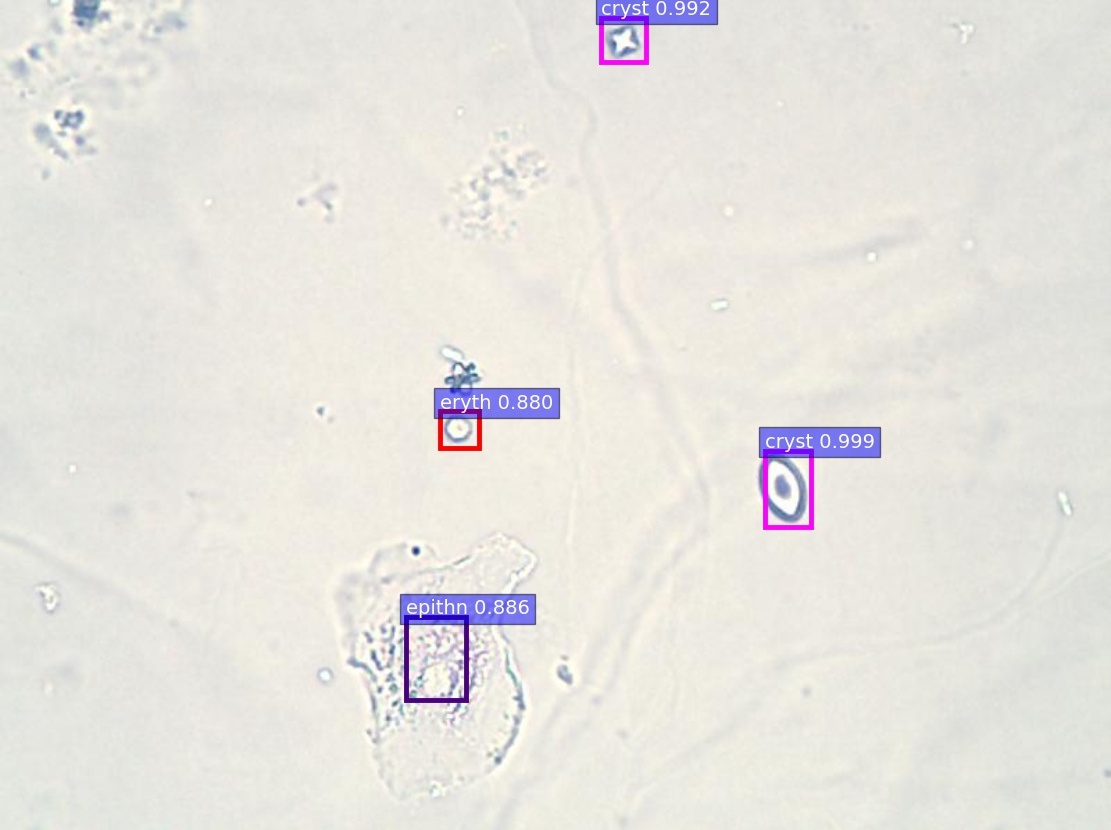}
   		\caption*{(c) VGG-16}
   	\end{minipage}
   	\hspace{-5ex}
   	\vspace{-0ex}
   	
   	\begin{minipage}{0.3\linewidth}
   		\centering              	
   		\setlength{\abovecaptionskip}{0cm}
   		\setlength{\belowcaptionskip}{0cm}
   		\includegraphics[height=3cm,width=4.5cm]{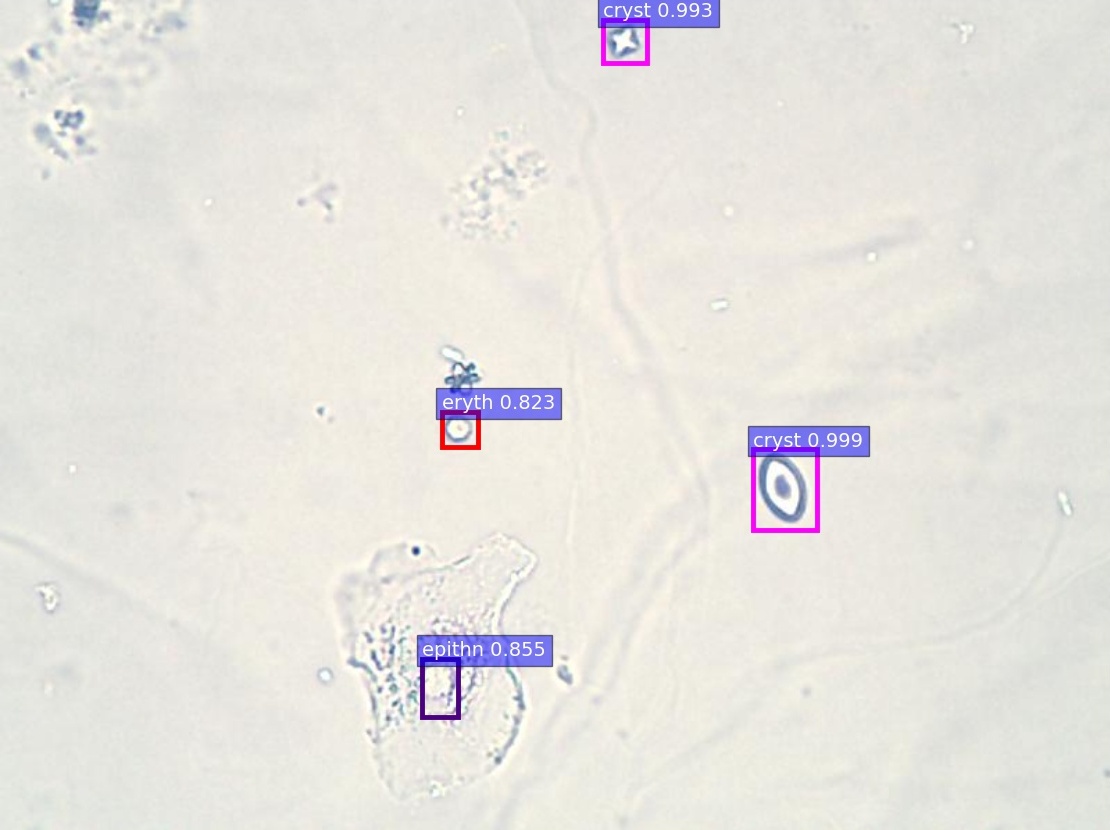}
   		\caption*{(d) ResNet-50}
   	\end{minipage}            	
   	\hspace{-5ex}
   	\vspace{-0ex}
   	\begin{minipage}{0.3\linewidth}
   		\centering
   		\setlength{\abovecaptionskip}{0cm}
   		\setlength{\belowcaptionskip}{0cm}
   		\includegraphics[height=3cm,width=4.5cm]{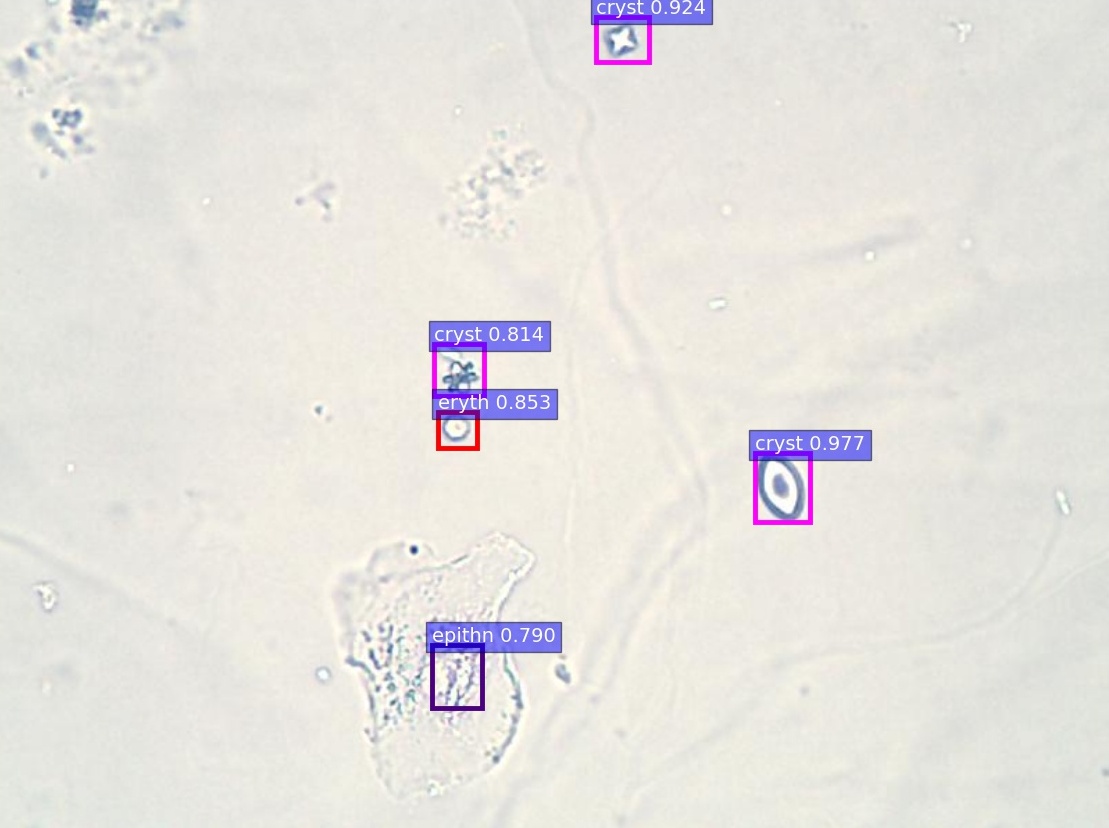}
   		\caption*{(e) PVANet}
   	\end{minipage}
   	\hspace{-5ex}
   	\vspace{-0ex}
   	\begin{minipage}{0.3\linewidth}
   		\centering
   		\setlength{\abovecaptionskip}{0cm}
   		\setlength{\belowcaptionskip}{0cm}
   		\includegraphics[height=3cm,width=4.5cm]{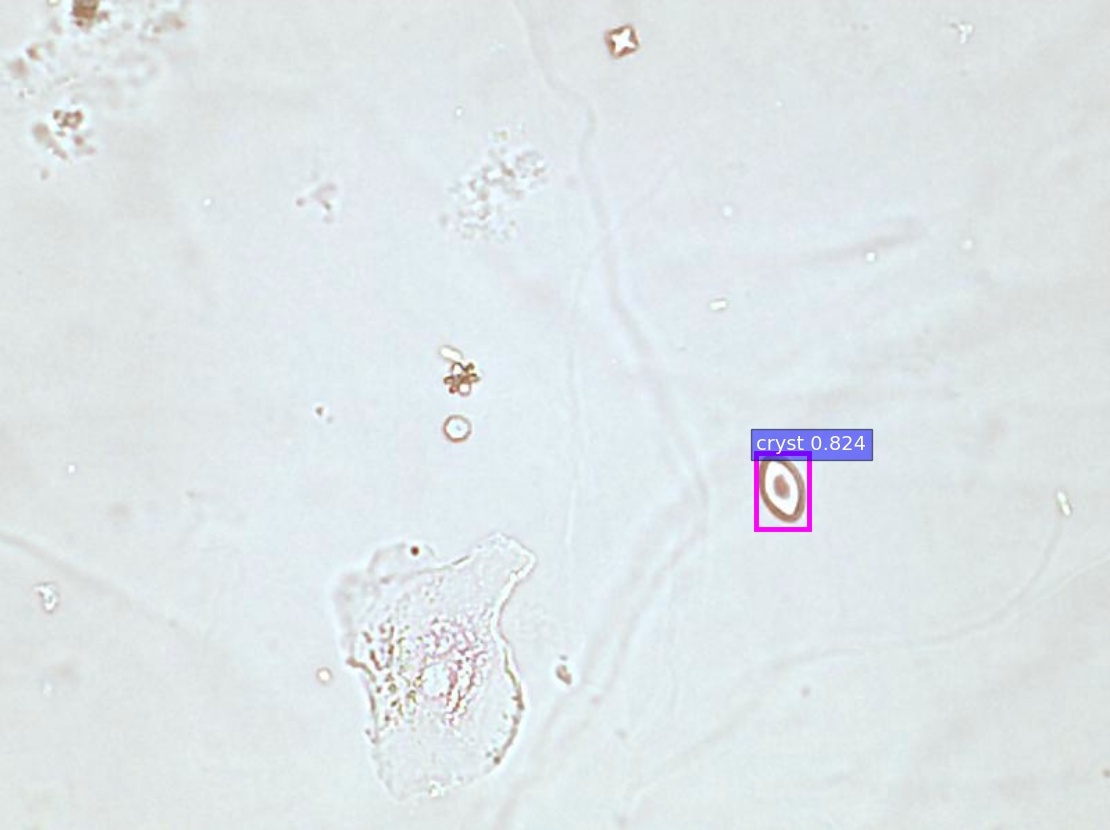}
   		\caption*{(f) SSD300$^{*}$}
   	\end{minipage}
   	\hspace{-5ex}
   	\vspace{-0ex}
   	\setlength{\abovecaptionskip}{0cm}
   	\setlength{\belowcaptionskip}{0cm}                     	
   	\caption*{\Romannum{3} : detection results of crystal}
   	\label{fig:crystal}         	
   \end{figure*}
   
   \begin{figure*}[h!]
   	\centering
   	\begin{minipage}{0.3\linewidth}
   		\centering
   		\setlength{\abovecaptionskip}{0cm}
   		\setlength{\belowcaptionskip}{0cm}            	
   		\includegraphics[height=3cm,width=4.5cm]{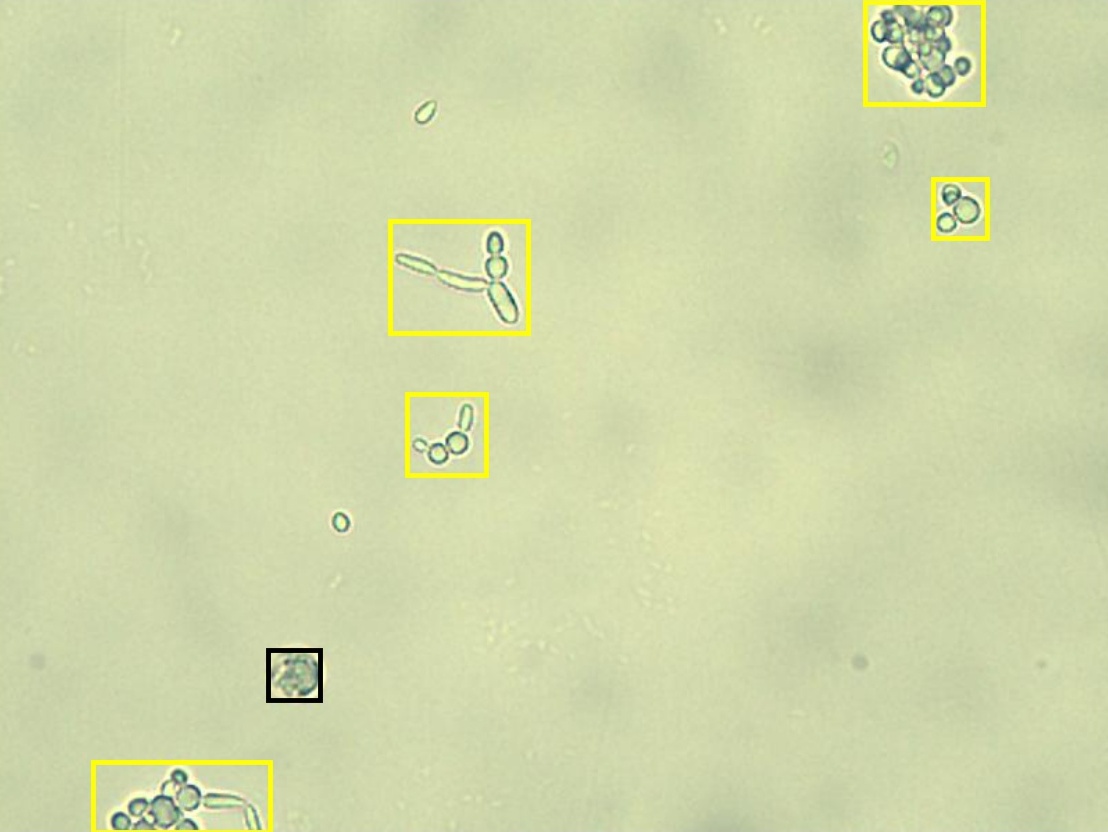}
   		\caption*{(a) annotations}
   	\end{minipage}
   	\hspace{-5ex}
   	\vspace{-0ex}           	
   	\begin{minipage}{0.3\linewidth}
   		\centering              	
   		\setlength{\abovecaptionskip}{0cm}
   		\setlength{\belowcaptionskip}{0cm}
   		\includegraphics[height=3cm,width=4.5cm]{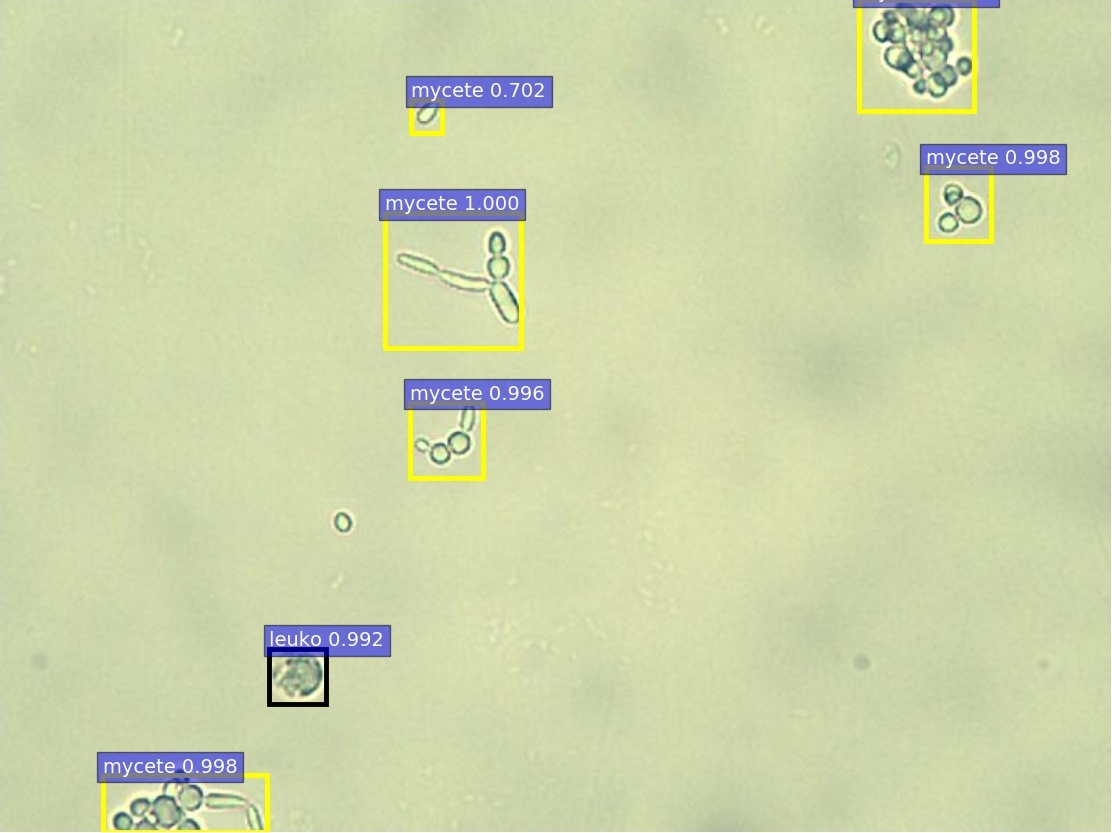}
   		\caption*{(b) ZF}
   	\end{minipage}
   	\hspace{-5ex}
   	\vspace{-0ex}
   	\begin{minipage}{0.3\linewidth}
   		\centering           		
   		\setlength{\abovecaptionskip}{0cm}
   		\setlength{\belowcaptionskip}{0cm}
   		\includegraphics[height=3cm,width=4.5cm]{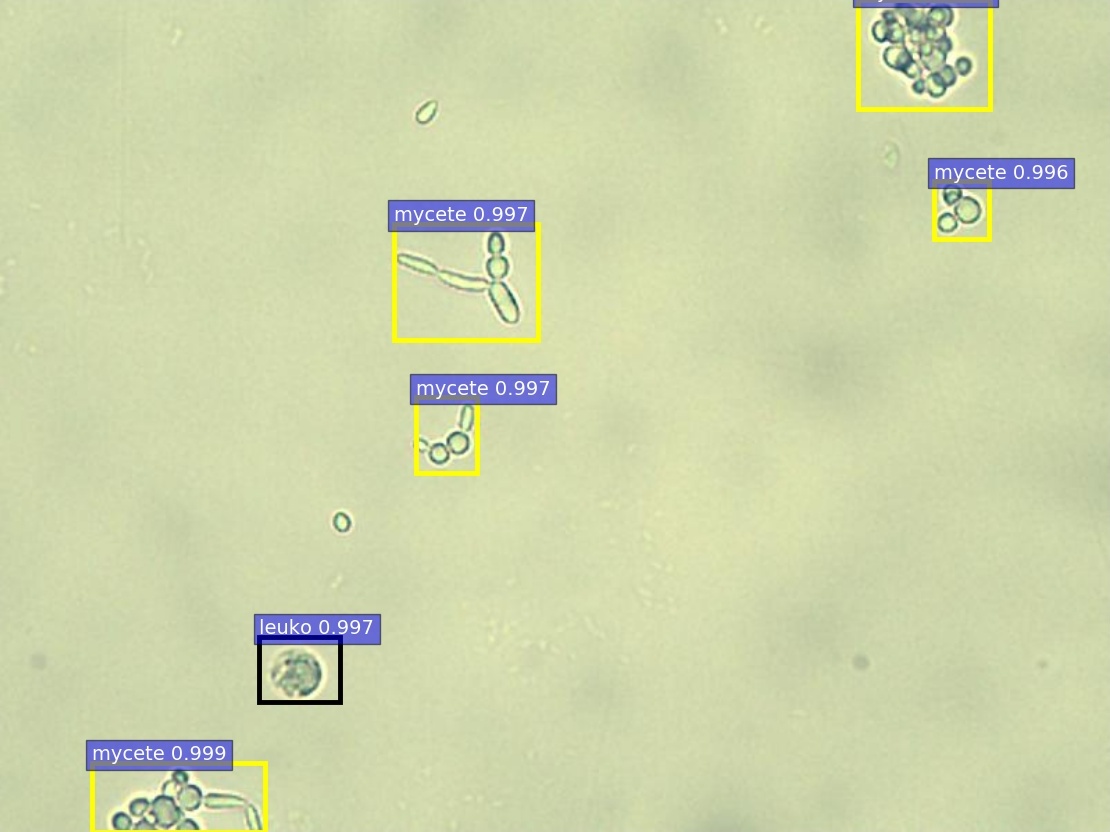}
   		\caption*{(c) VGG-16}
   	\end{minipage}
   	\hspace{-5ex}
   	\vspace{-0ex}
   	
   	\begin{minipage}{0.3\linewidth}
   		\centering              	
   		\setlength{\abovecaptionskip}{0cm}
   		\setlength{\belowcaptionskip}{0cm}
   		\includegraphics[height=3cm,width=4.5cm]{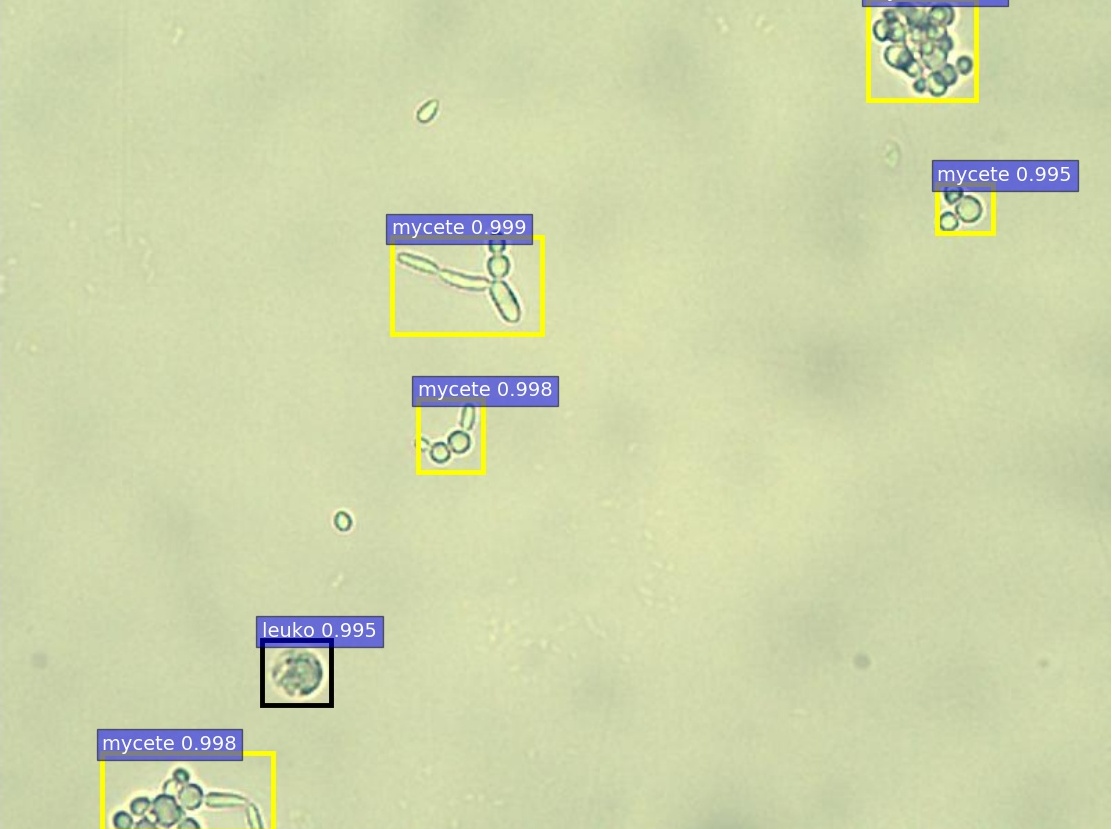}
   		\caption*{(d) ResNet-50}
   	\end{minipage}            	
   	\hspace{-5ex}
   	\vspace{-0ex}
   	\begin{minipage}{0.3\linewidth}
   		\centering
   		\setlength{\abovecaptionskip}{0cm}
   		\setlength{\belowcaptionskip}{0cm}
   		\includegraphics[height=3cm,width=4.5cm]{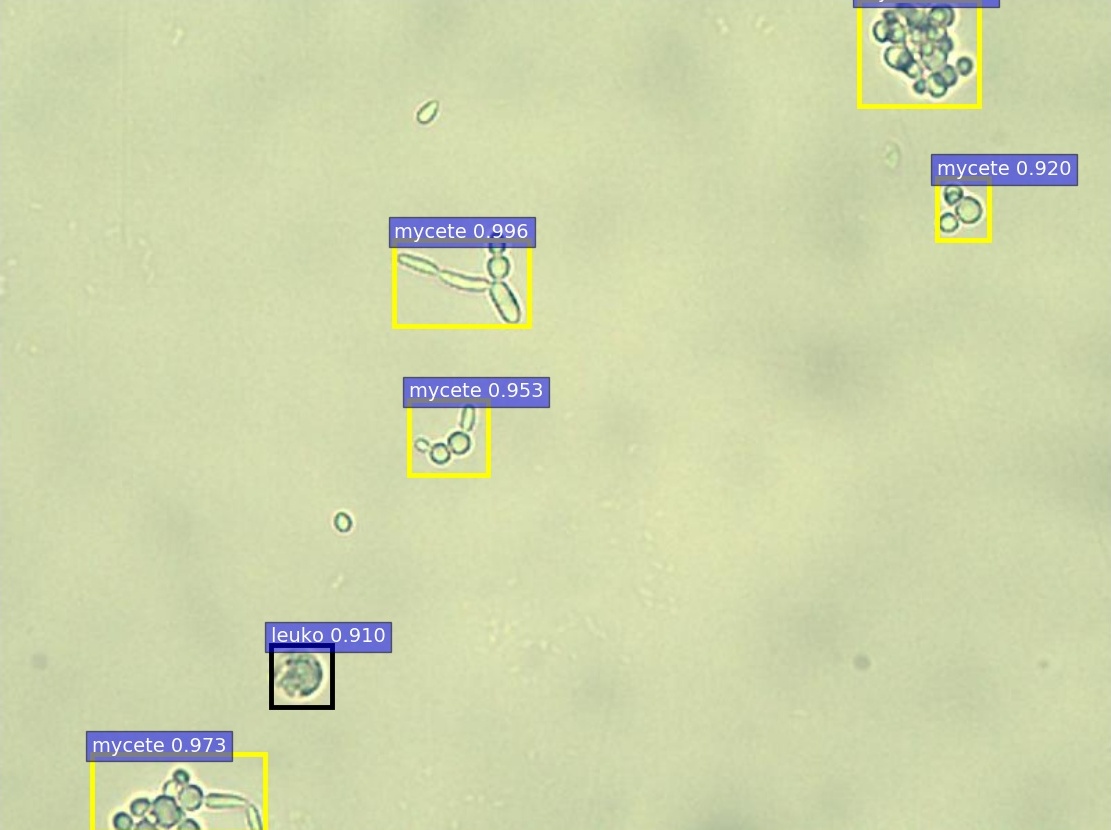}
   		\caption*{(e) PVANet}
   	\end{minipage}
   	\hspace{-5ex}
   	\vspace{-0ex}
   	\begin{minipage}{0.3\linewidth}
   		\centering
   		\setlength{\abovecaptionskip}{0cm}
   		\setlength{\belowcaptionskip}{0cm}
   		\includegraphics[height=3cm,width=4.5cm]{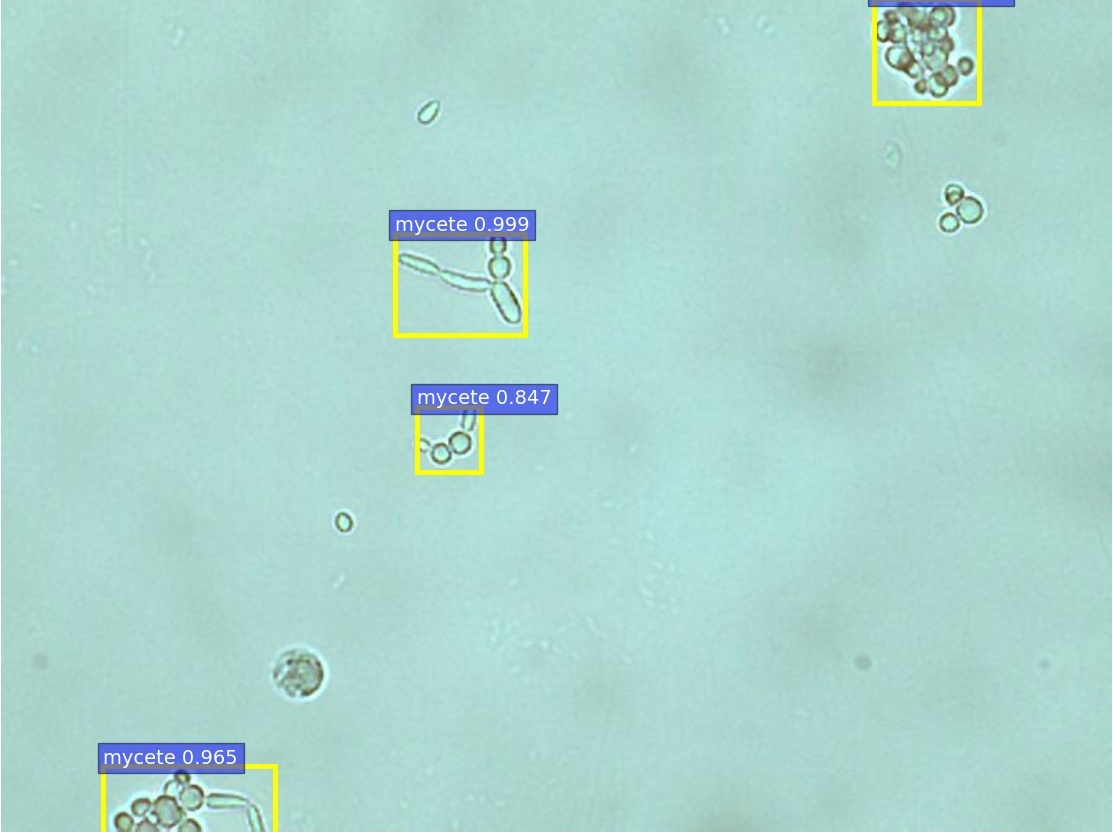}
   		\caption*{(f) SSD300$^{*}$}
   	\end{minipage}
   	\hspace{-5ex}
   	\vspace{-0ex}
   	\setlength{\abovecaptionskip}{0cm}
   	\setlength{\belowcaptionskip}{0cm}                     	
   	\caption*{\Romannum{4} : detection results of mycete}
   	\label{fig:mycete}         	
   \end{figure*}
   
   \begin{figure*}[t!]
   	\centering             	
   	\begin{minipage}{0.3\linewidth}
   		\centering
   		\setlength{\abovecaptionskip}{0cm}
   		\setlength{\belowcaptionskip}{0cm}            	
   		\includegraphics[height=3cm,width=4.5cm]{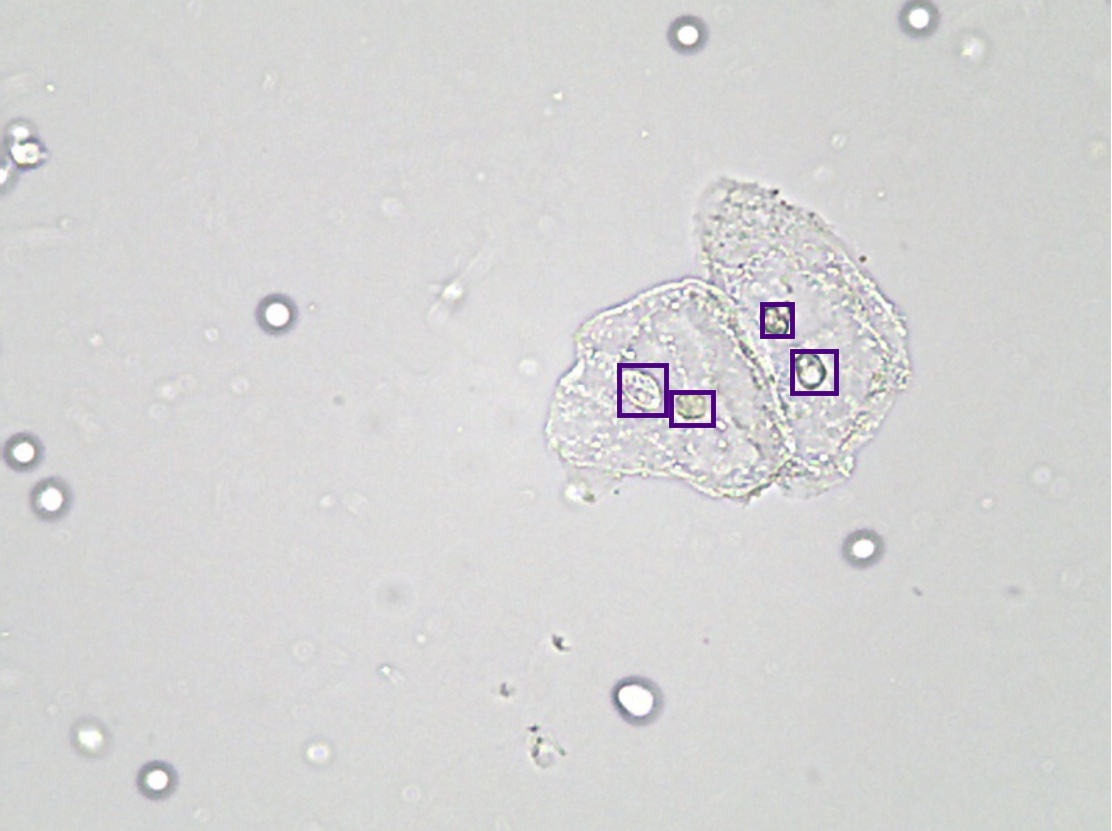}
   		\caption*{(a) annotations}
   	\end{minipage}
   	\hspace{-5ex}
   	\vspace{-0ex}
   	\begin{minipage}{0.3\linewidth}
   		\centering              	
   		\setlength{\abovecaptionskip}{0cm}
   		\setlength{\belowcaptionskip}{0cm}
   		\includegraphics[height=3cm,width=4.5cm]{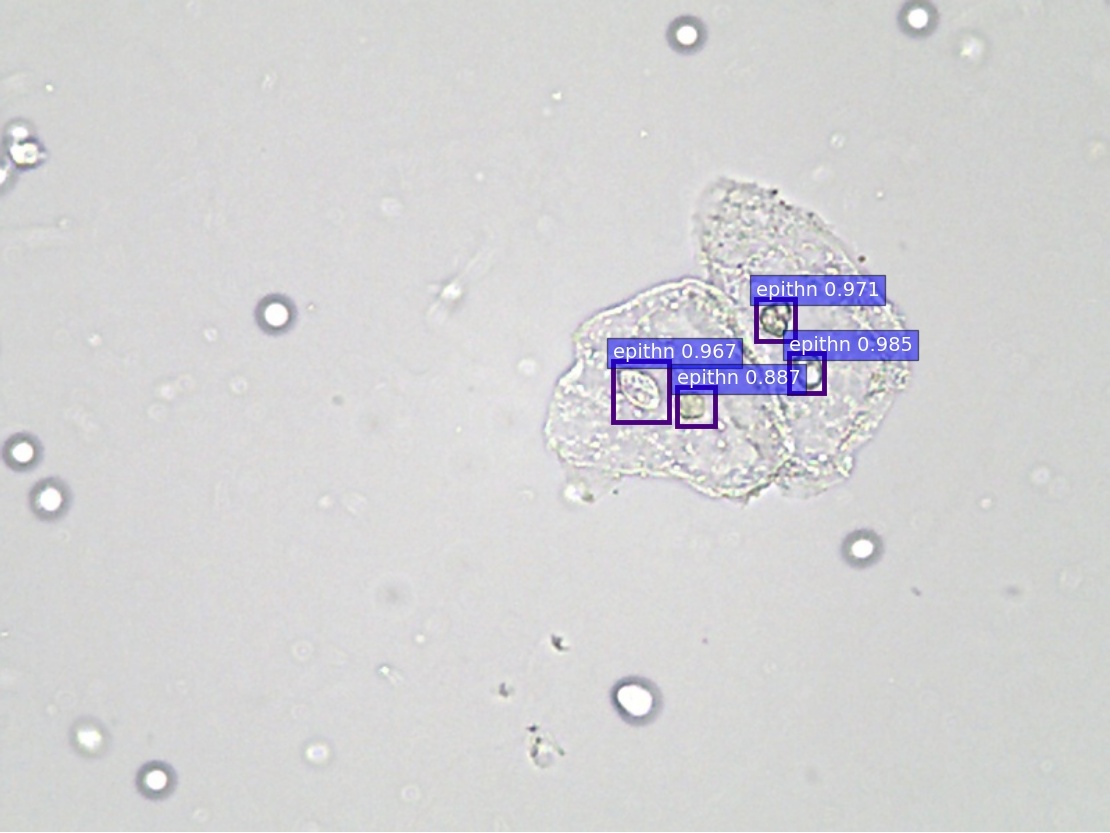}
   		\caption*{(b) ZF}
   	\end{minipage}
   	\hspace{-5ex}
   	\vspace{-0ex}
   	\begin{minipage}{0.3\linewidth}
   		\centering           		
   		\setlength{\abovecaptionskip}{0cm}
   		\setlength{\belowcaptionskip}{0cm}
   		\includegraphics[height=3cm,width=4.5cm]{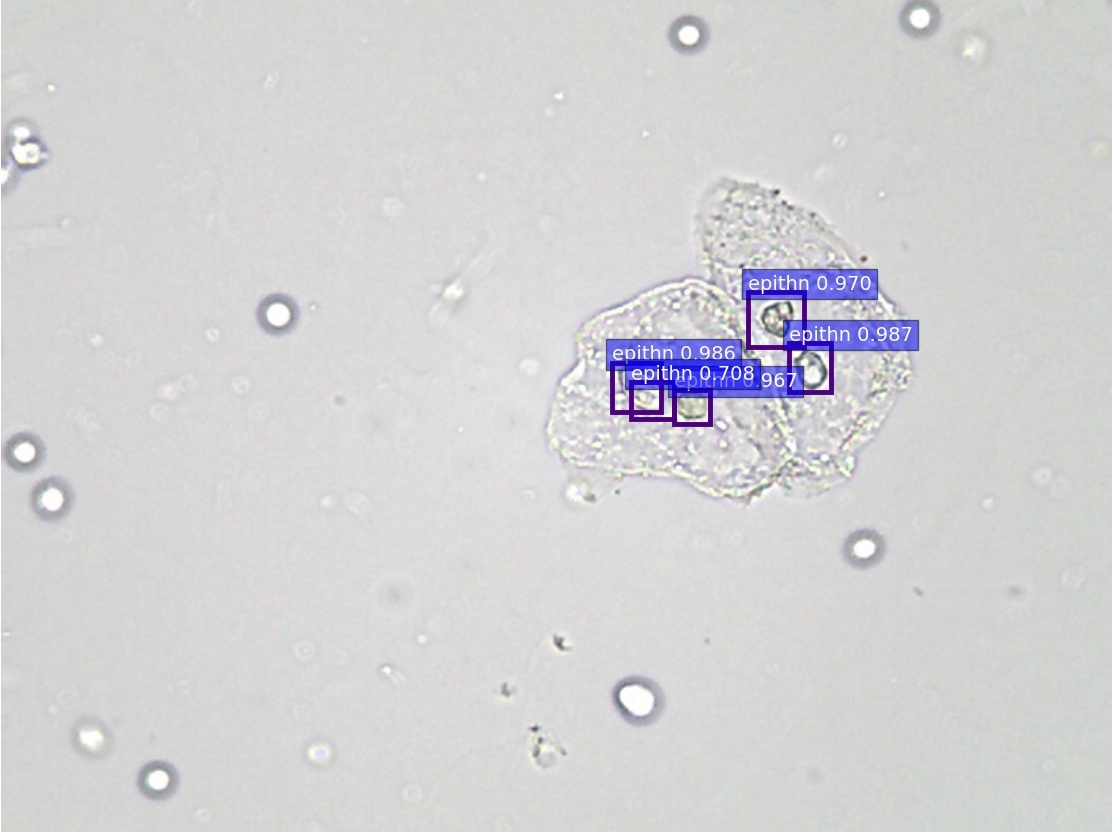}
   		\caption*{(c) VGG-16}
   	\end{minipage}
   	\hspace{-5ex}
   	\vspace{-0ex}
   	
   	\begin{minipage}{0.3\linewidth}
   		\centering              	
   		\setlength{\abovecaptionskip}{0cm}
   		\setlength{\belowcaptionskip}{0cm}
   		\includegraphics[height=3cm,width=4.5cm]{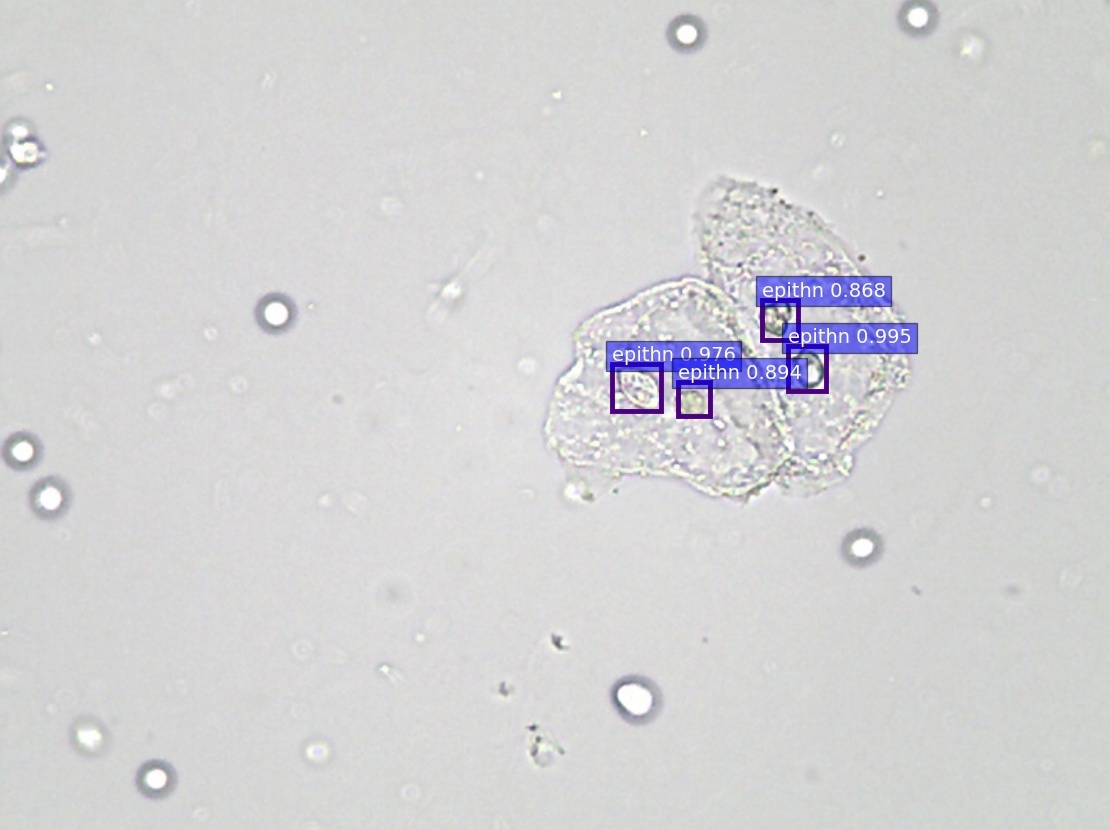}
   		\caption*{(d) ResNet-50}
   	\end{minipage}            	
   	\hspace{-5ex}
   	\vspace{-0ex}
   	\begin{minipage}{0.3\linewidth}
   		\centering
   		\setlength{\abovecaptionskip}{0cm}
   		\setlength{\belowcaptionskip}{0cm}
   		\includegraphics[height=3cm,width=4.5cm]{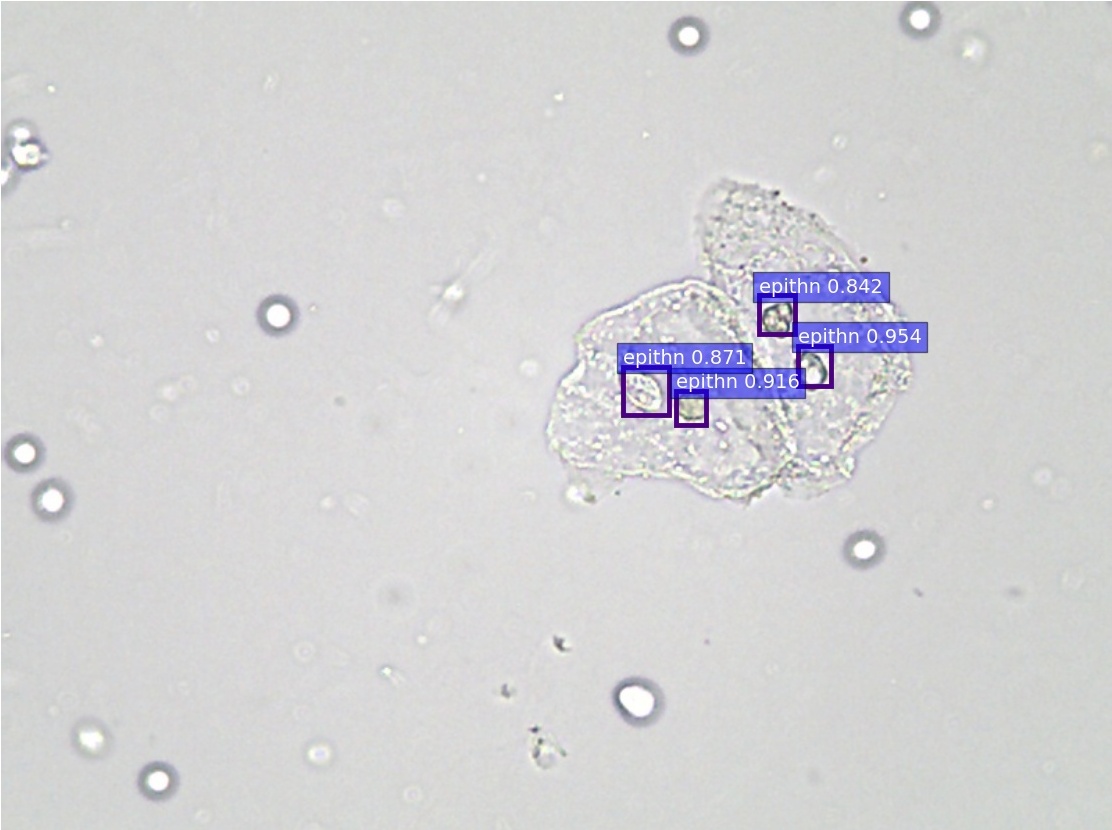}
   		\caption*{(e) PVANet}
   	\end{minipage}
   	\hspace{-5ex}
   	\vspace{-0ex}
   	\begin{minipage}{0.3\linewidth}
   		\centering
   		\setlength{\abovecaptionskip}{0cm}
   		\setlength{\belowcaptionskip}{0cm}
   		\includegraphics[height=3cm,width=4.5cm]{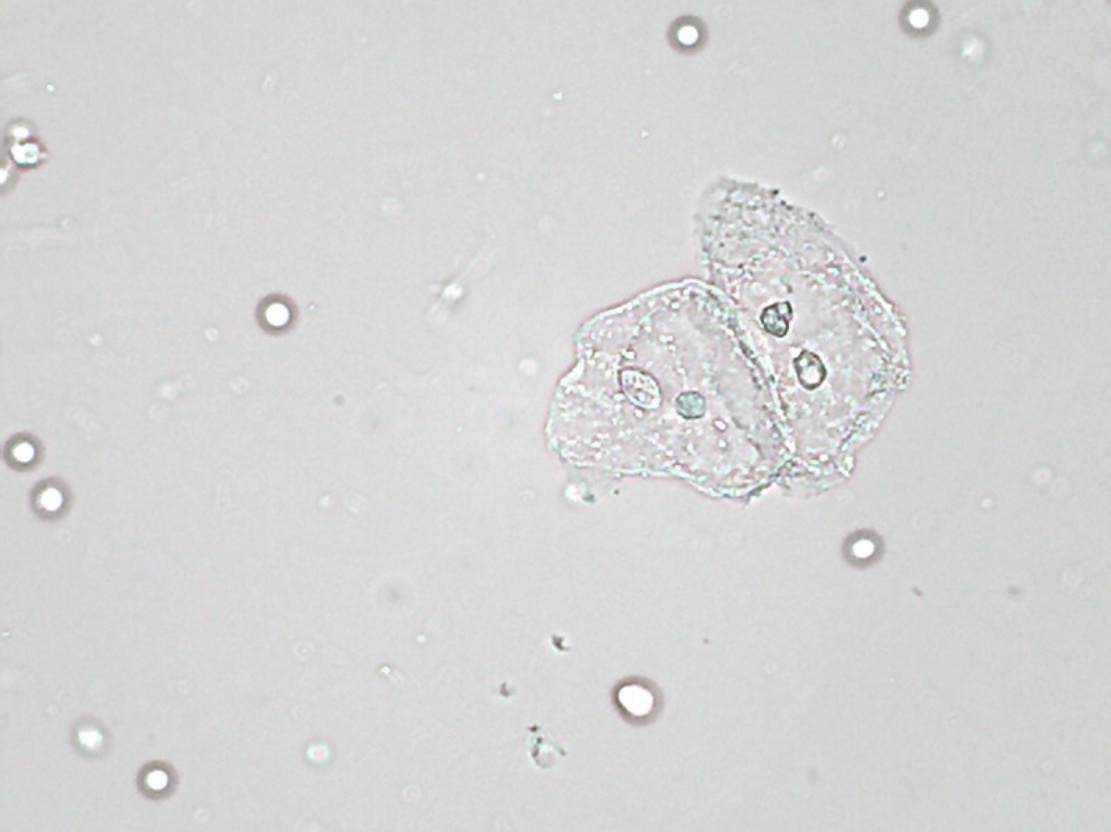}
   		\caption*{(f) SSD300$^{*}$}
   	\end{minipage}
   	\hspace{-5ex}
   	\vspace{-0ex}
   	\setlength{\abovecaptionskip}{0cm}
   	\setlength{\belowcaptionskip}{0cm}                     	
   	\caption*{\Romannum{5} : detection results of epithelial nuclei}
   	\label{fig:epithn}         	
   \end{figure*}
   
   \begin{figure*}[t!]
   	\centering
   	\begin{minipage}{0.3\linewidth}
   		\centering
   		\setlength{\abovecaptionskip}{0cm}
   		\setlength{\belowcaptionskip}{0cm}            	
   		\includegraphics[height=3cm,width=4.5cm]{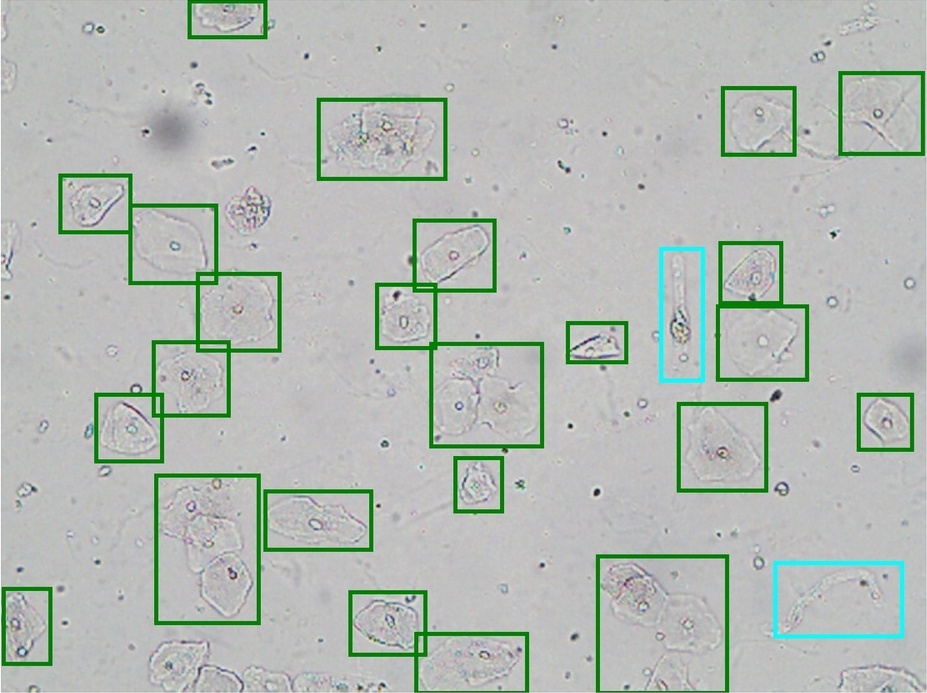}
   		\caption*{(a) annotations}
   	\end{minipage}
   	\hspace{-5ex}
   	\vspace{-0ex}           	
   	\begin{minipage}{0.3\linewidth}
   		\centering              	
   		\setlength{\abovecaptionskip}{0cm}
   		\setlength{\belowcaptionskip}{0cm}
   		\includegraphics[height=3cm,width=4.5cm]{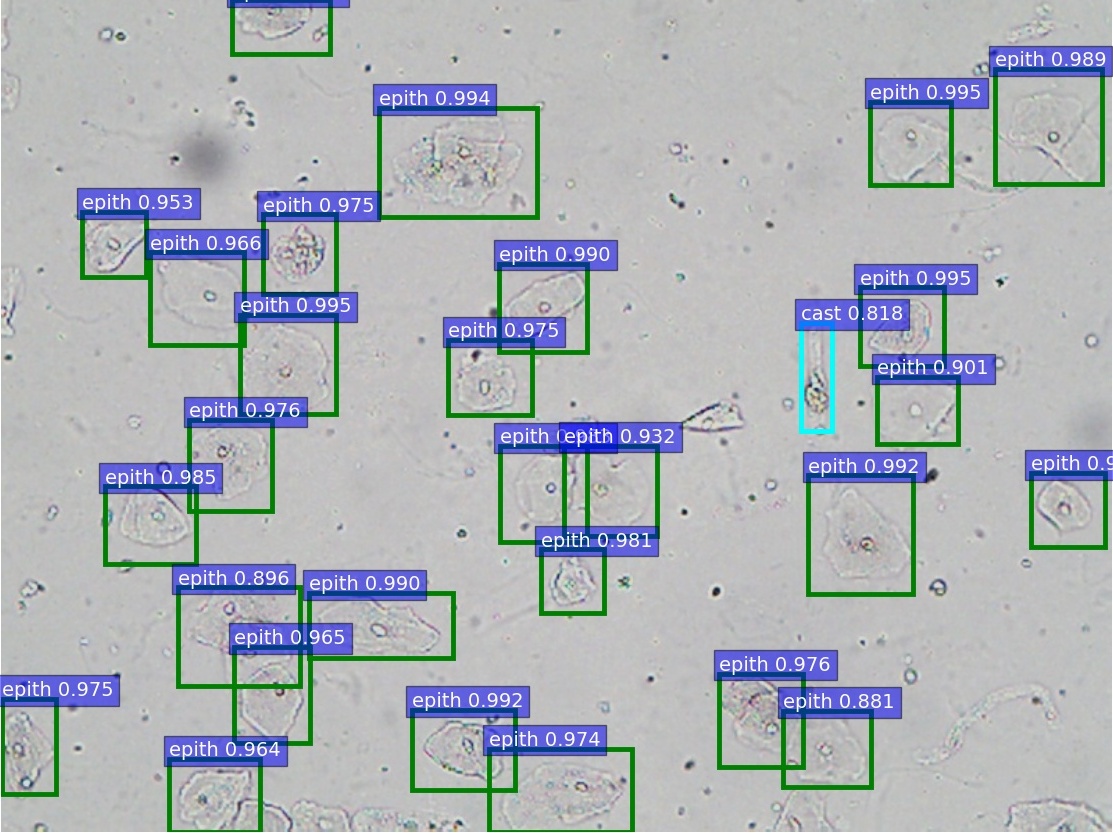}
   		\caption*{(b) ZF}
   	\end{minipage}
   	\hspace{-5ex}
   	\vspace{-0ex}
   	\begin{minipage}{0.3\linewidth}
   		\centering           		
   		\setlength{\abovecaptionskip}{0cm}
   		\setlength{\belowcaptionskip}{0cm}
   		\includegraphics[height=3cm,width=4.5cm]{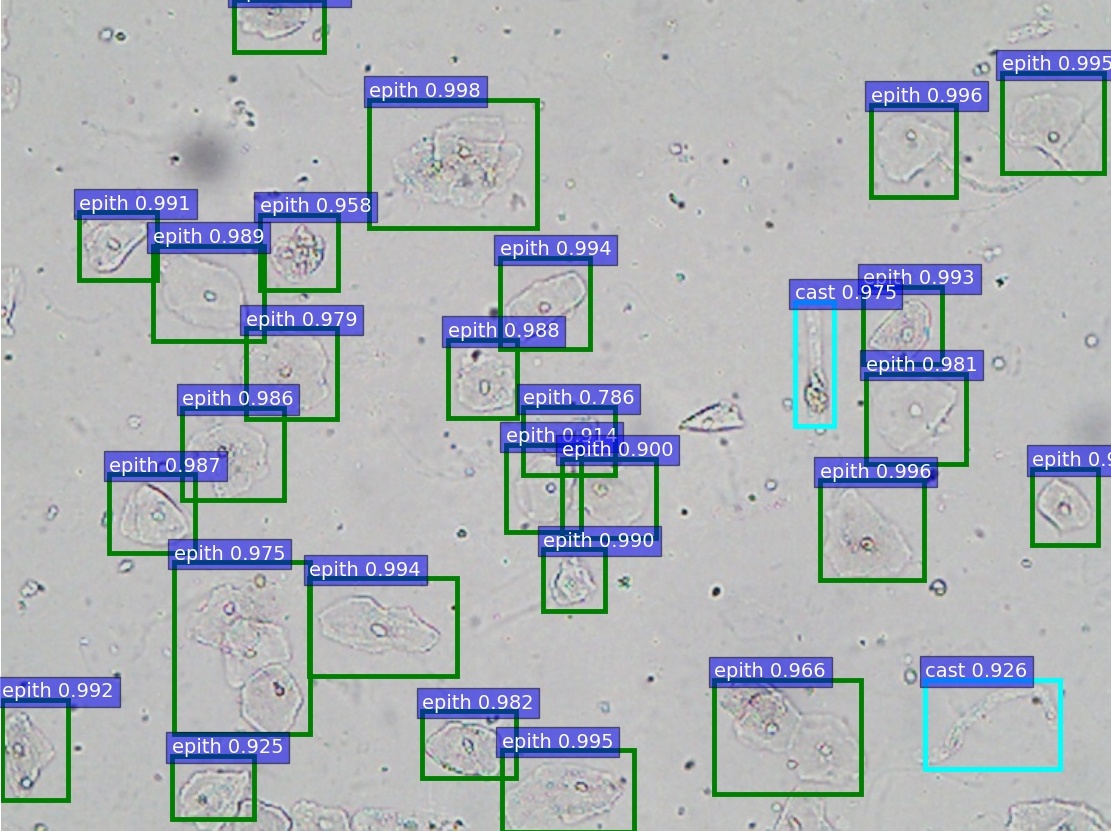}
   		\caption*{(c) VGG-16}
   	\end{minipage}
   	\hspace{-5ex}
   	\vspace{-0ex}
   	
   	\begin{minipage}{0.3\linewidth}
   		\centering              	
   		\setlength{\abovecaptionskip}{0cm}
   		\setlength{\belowcaptionskip}{0cm}
   		\includegraphics[height=3cm,width=4.5cm]{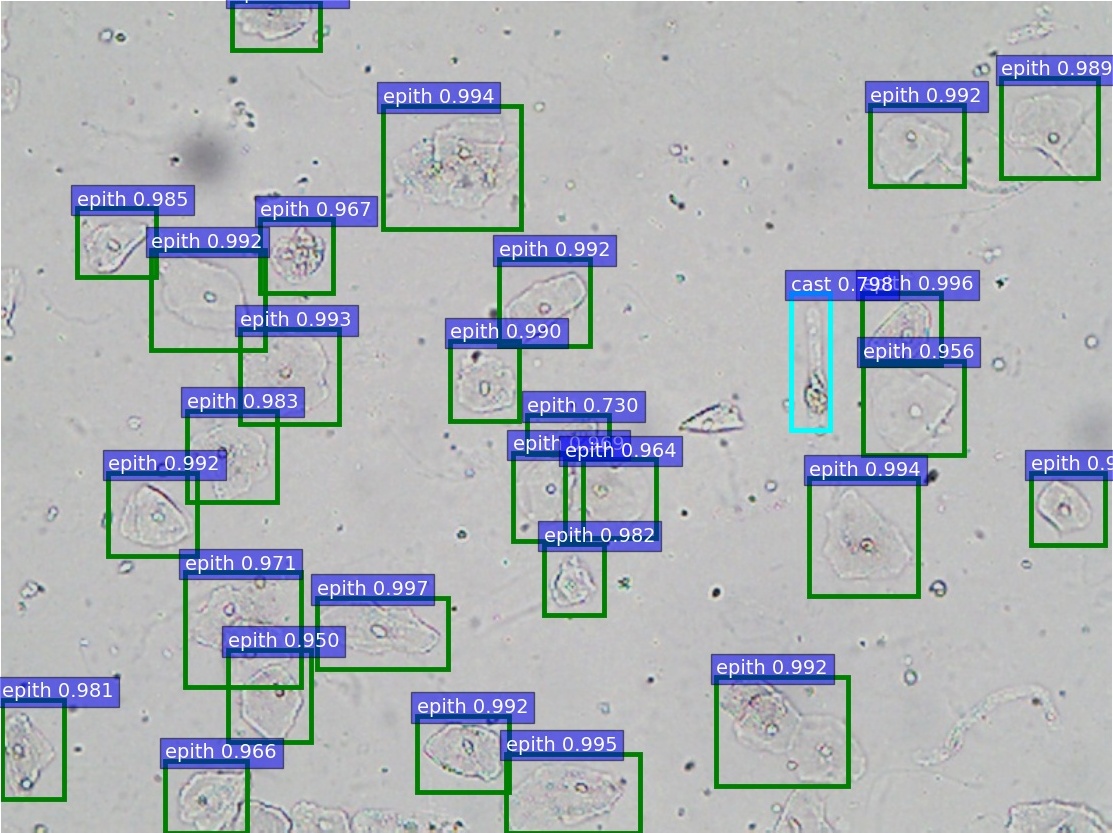}
   		\caption*{(d) ResNet-50}
   	\end{minipage}            	
   	\hspace{-5ex}
   	\vspace{-0ex}
   	\begin{minipage}{0.3\linewidth}
   		\centering
   		\setlength{\abovecaptionskip}{0cm}
   		\setlength{\belowcaptionskip}{0cm}
   		\includegraphics[height=3cm,width=4.5cm]{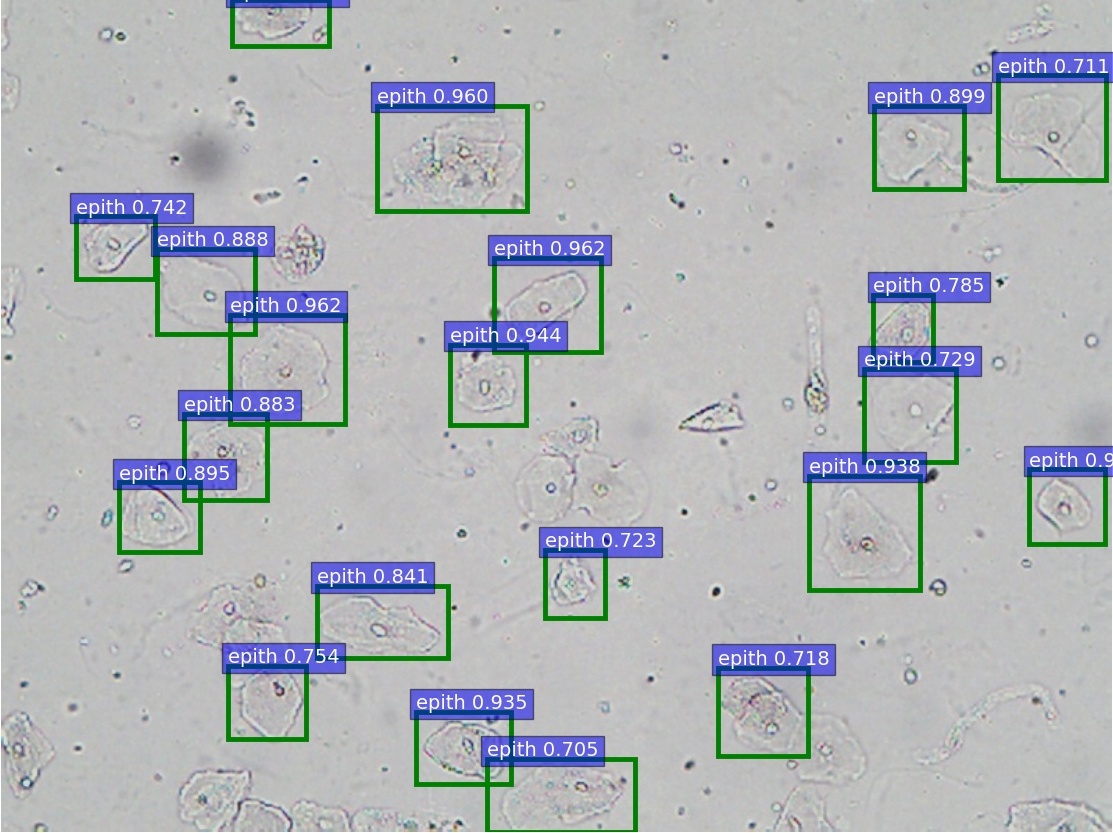}
   		\caption*{(e) PVANet}
   	\end{minipage}
   	\hspace{-5ex}
   	\vspace{-0ex}
   	\begin{minipage}{0.3\linewidth}
   		\centering
   		\setlength{\abovecaptionskip}{0cm}
   		\setlength{\belowcaptionskip}{0cm}
   		\includegraphics[height=3cm,width=4.5cm]{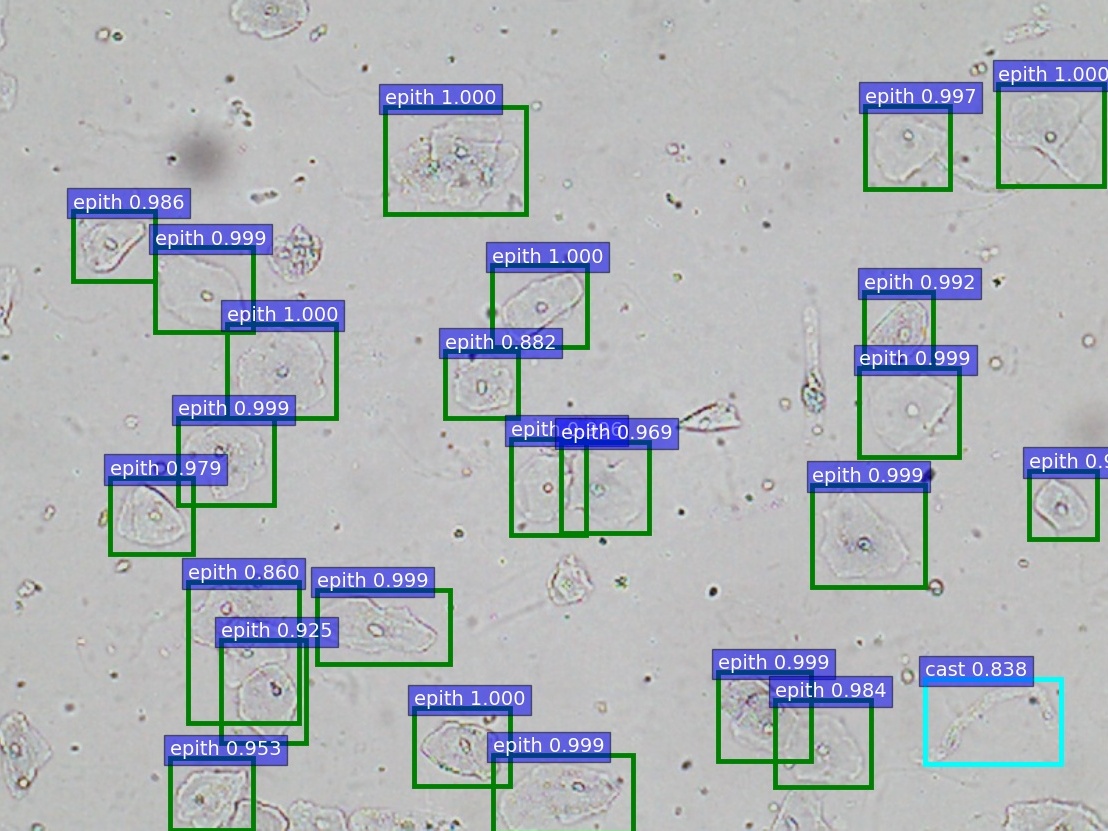}
   		\caption*{(f) SSD300$^{*}$}
   	\end{minipage}
   	\hspace{-5ex}
   	\vspace{-0ex}
   	\setlength{\abovecaptionskip}{0cm}
   	\setlength{\belowcaptionskip}{0cm}                     	
   	\caption*{\Romannum{6} : detection results of epithelial cell}
   	\label{fig:epit}         	
   \end{figure*}
   
   \begin{figure*}[t!]
   	\centering
   	\begin{minipage}{0.3\linewidth}
   		\centering
   		\setlength{\abovecaptionskip}{0cm}
   		\setlength{\belowcaptionskip}{0cm}            	
   		\includegraphics[height=3cm,width=4.5cm]{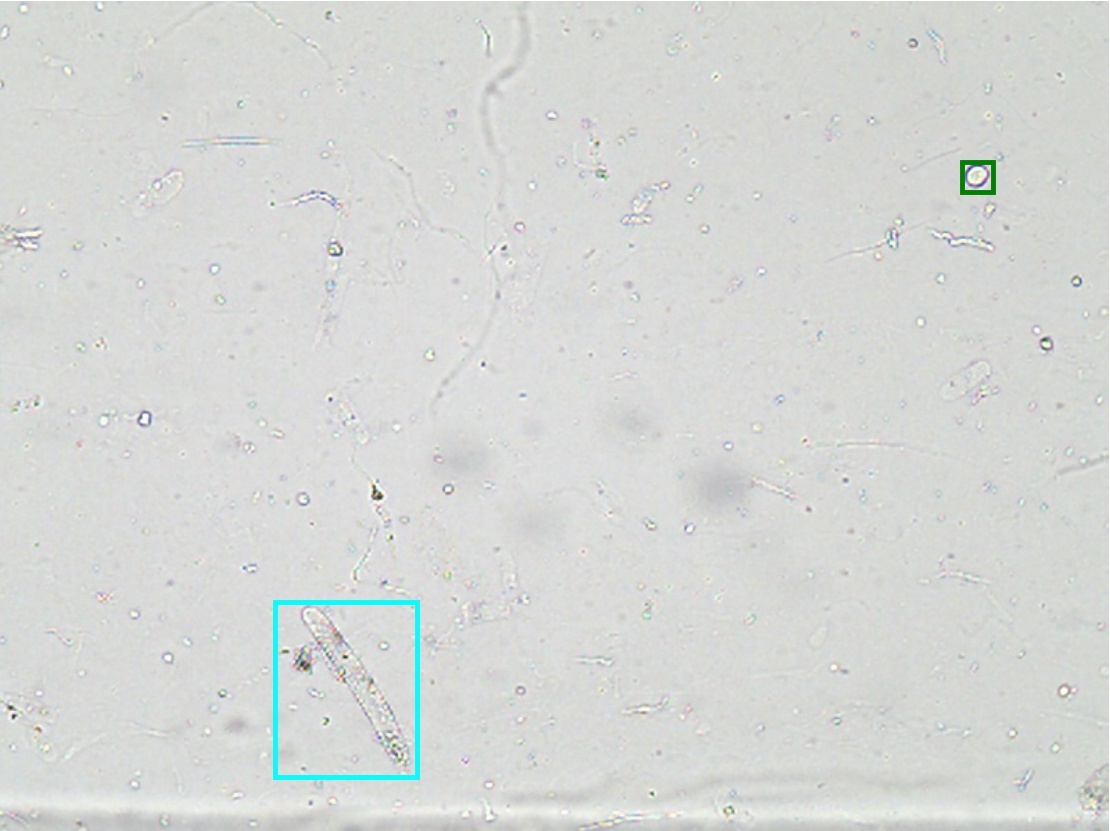}
   		\caption*{(a) annotations}
   	\end{minipage}
   	\hspace{-5ex}
   	\vspace{-0ex}           	
   	\begin{minipage}{0.3\linewidth}
   		\centering              	
   		\setlength{\abovecaptionskip}{0cm}
   		\setlength{\belowcaptionskip}{0cm}
   		\includegraphics[height=3cm,width=4.5cm]{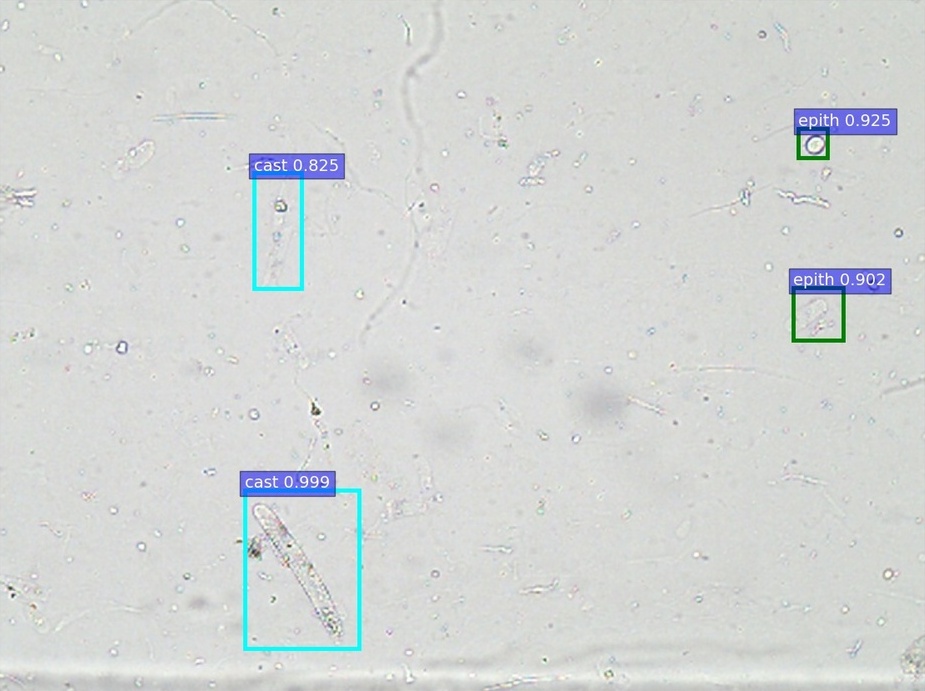}
   		\caption*{(b) ZF}
   	\end{minipage}
   	\hspace{-5ex}
   	\vspace{-0ex}
   	\begin{minipage}{0.3\linewidth}
   		\centering           		
   		\setlength{\abovecaptionskip}{0cm}
   		\setlength{\belowcaptionskip}{0cm}
   		\includegraphics[height=3cm,width=4.5cm]{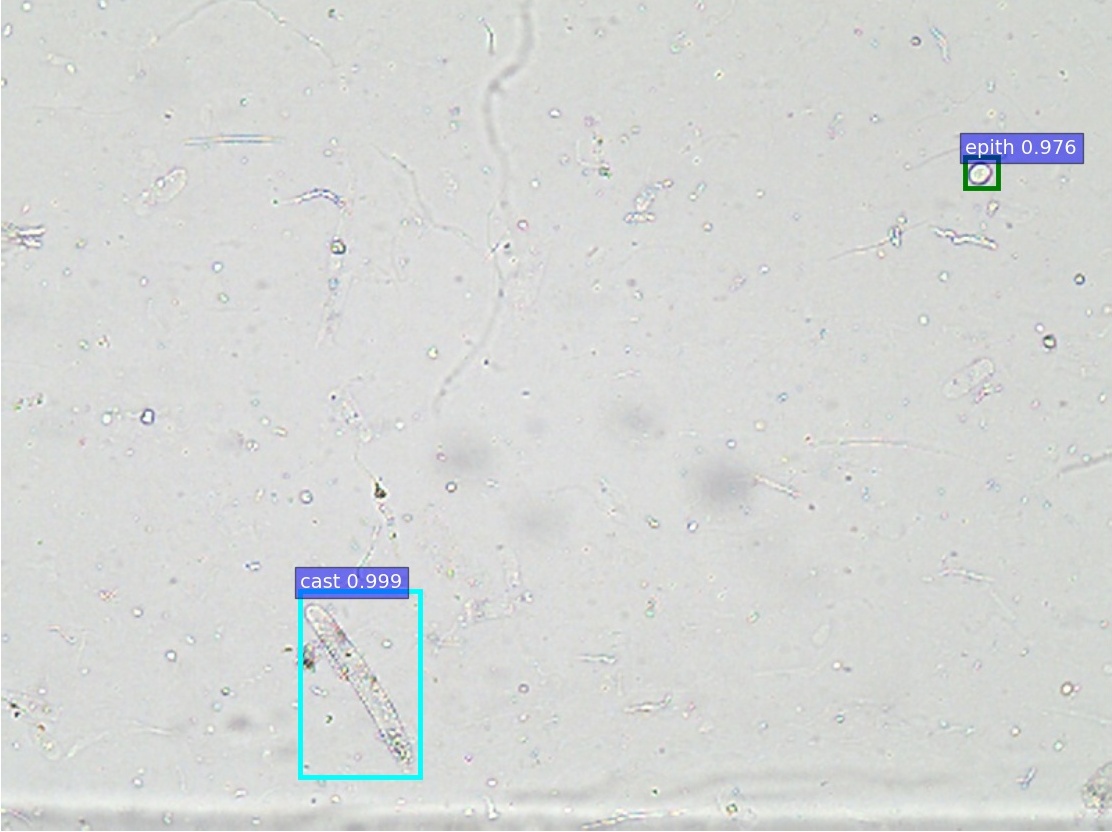}
   		\caption*{(c) VGG-16}
   	\end{minipage}
   	\hspace{-5ex}
   	\vspace{-0ex}
   	
   	\begin{minipage}{0.3\linewidth}
   		\centering              	
   		\setlength{\abovecaptionskip}{0cm}
   		\setlength{\belowcaptionskip}{0cm}
   		\includegraphics[height=3cm,width=4.5cm]{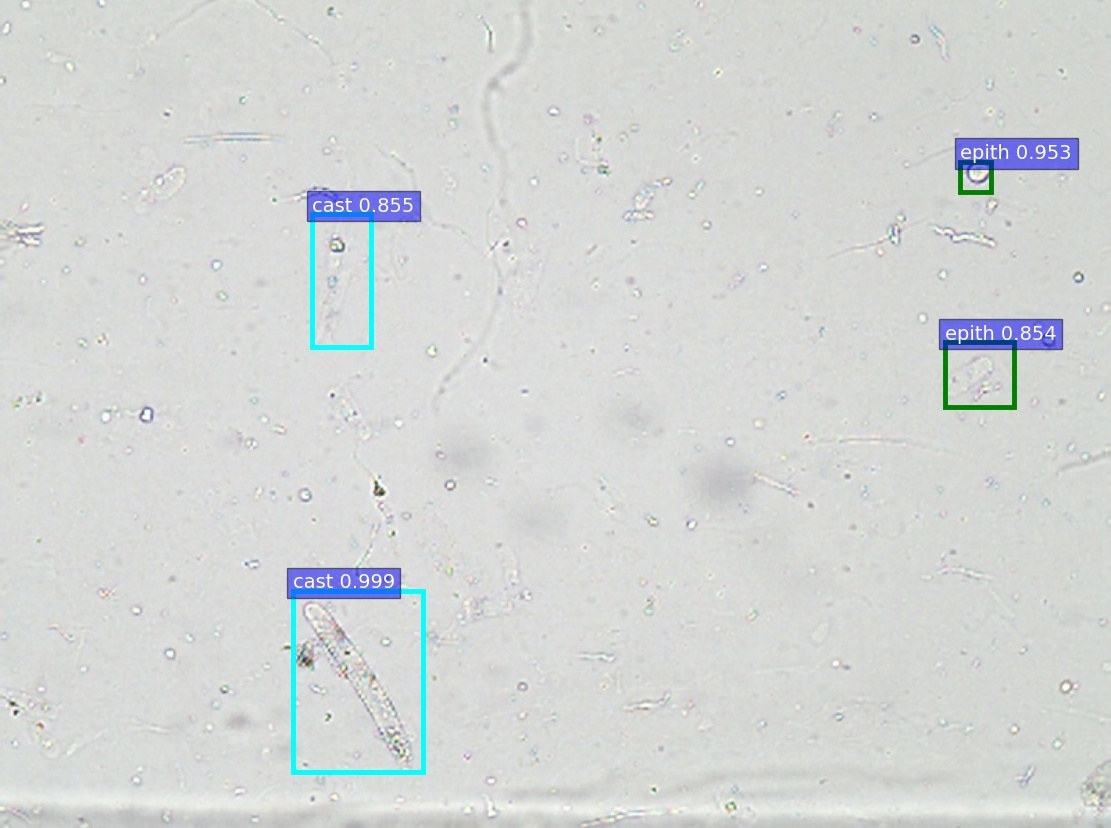}
   		\caption*{(d) ResNet-50}
   	\end{minipage}            	
   	\hspace{-5ex}
   	\vspace{-0ex}
   	\begin{minipage}{0.3\linewidth}
   		\centering
   		\setlength{\abovecaptionskip}{0cm}
   		\setlength{\belowcaptionskip}{0cm}
   		\includegraphics[height=3cm,width=4.5cm]{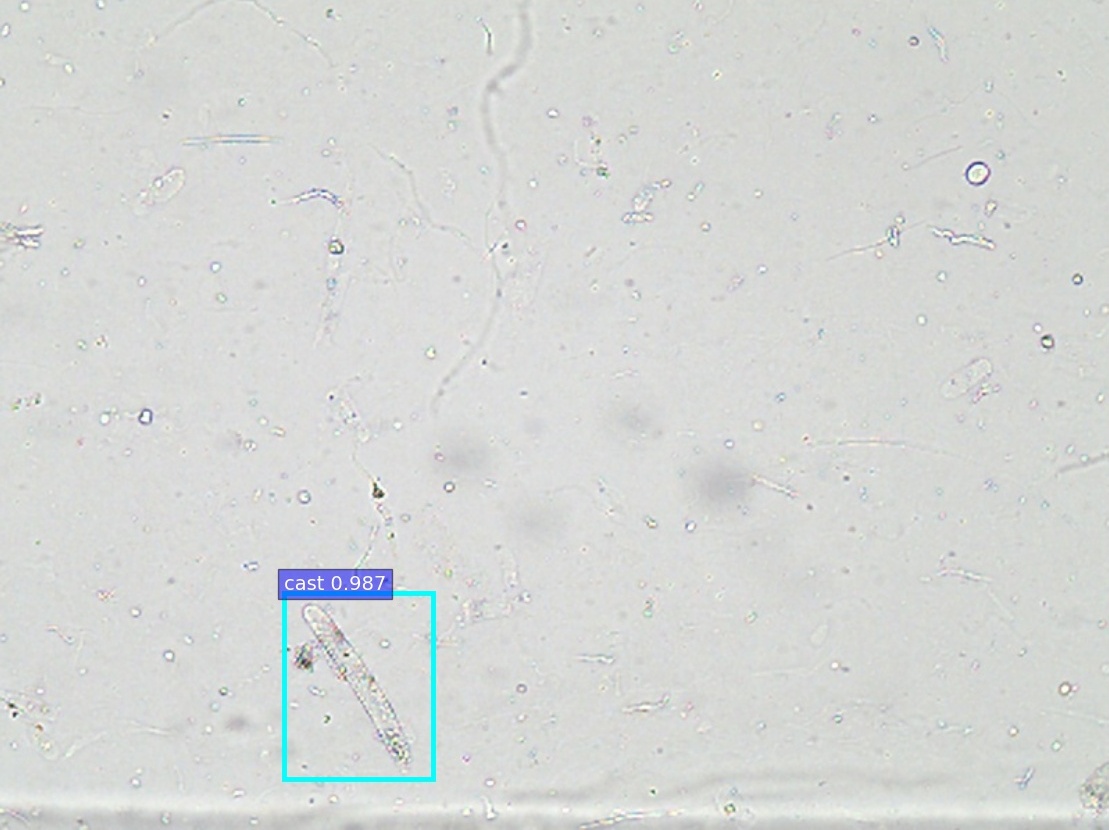}
   		\caption*{(e) PVANet}
   	\end{minipage}
   	\hspace{-5ex}
   	\vspace{-0ex}           	
   	\begin{minipage}{0.3\linewidth}
   		\centering
   		\setlength{\abovecaptionskip}{0cm}
   		\setlength{\belowcaptionskip}{0cm}
   		\includegraphics[height=3cm,width=4.5cm]{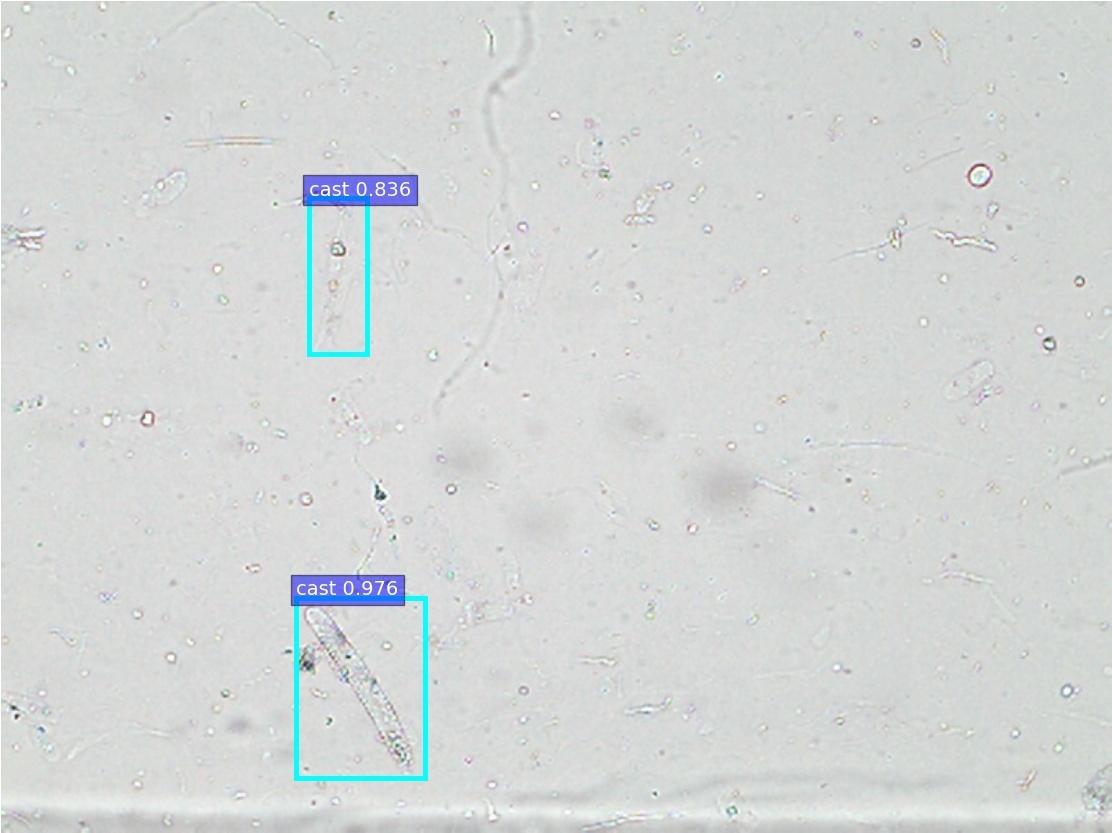}
   		\caption*{(f) SSD300$^{*}$}
   	\end{minipage}
   	\hspace{-5ex}
   	\vspace{-0ex}
   	\setlength{\abovecaptionskip}{0cm}
   	\setlength{\belowcaptionskip}{0cm}                     	
   	\caption*{\Romannum{7} : detection results of cast}
   	\label{fig:cast}         	
   \end{figure*}
   
   \begin{figure*}[t!]
   	\centering
   	\setlength{\abovecaptionskip}{0cm}
   	\setlength{\belowcaptionskip}{0cm}        	    	
   	\caption{Selected detection examples of urine particles on urinalysis test set. We show detections with scores higher than 0.7. All examples are divided into 7 groups, where 5 groups are at high-power field (i.e., erythrocyte, leukocyte, crystal, mycete, epithelial nuclei ) and the other 2 groups at low-power field (i.e., epithelial cell, cast ). In each group: (a) shows original image with ground truth boxes; (b-d) are Faster R-CNN detections separately on ZF, VGG-16 and ResNet-50 networks with a anchor scales of $\{ 32^{2}, ~64^{2}, ~128^{2}, ~256^{2}, ~512^{2} \}$; (e) shows detection results on PVANet; (f) shows detection results on SSD300$^{*}$ model. For the ground truths and detection boxes, different categories use only different colors: eryth (red), leuko (black), epith (green), crystal (magenta), cast (cyan), mycete (yellow). As shown in this figure, the performance of SSD is inferior to Faster R-CNN, and it misses a lot of small objects.}
   	\label{fig:detections}
   \end{figure*}
\end{appendices} 
  
\end{document}